\documentclass[11pt, a4paper,logo,  copyright, nonumbering]{paddleocr}
\usepackage[numbers]{natbib}
\usepackage{dblfloatfix}
\usepackage{ulem}
\usepackage{float}
\usepackage{tocloft}
\usepackage{caption}
\usepackage{dramatist}
\usepackage{xspace}
\usepackage{array}
\usepackage{pifont} %
\usepackage{multirow}
\usepackage{makecell}
\usepackage{tcolorbox}
\usepackage{xltabular}
\usepackage{titlesec} 
\usepackage{longtable}
\usepackage[hang, flushmargin]{footmisc}
\usepackage{booktabs} 
\usepackage{xcolor}   
\usepackage[table]{xcolor}
\definecolor{lightpink}{RGB}{255, 182, 193}
\definecolor{iccvblue}{rgb}{0.21,0.49,0.74}

\hypersetup{
    colorlinks=true,
    linkcolor=blue,
    citecolor=blue,
    filecolor=blue,
    urlcolor=blue,
    pdfpagemode=UseNone,
    pdfstartview=FitH
}
\usepackage{threeparttable} 
\usepackage{soul}
\usepackage{amsfonts}
\usepackage{amsmath}
\usepackage{amssymb}
\usepackage{lineno}
\usepackage{multirow}
\usepackage{adjustbox}

\usepackage{CJKutf8}
\usepackage{subcaption}
\usepackage{setspace}

\usepackage{dsfont}
\usepackage{array} %
\usepackage{tabularx} %
\usepackage{xcolor} %
\usepackage{tabularx}
\usepackage{booktabs}

\usepackage{lipsum}  %
\usepackage{multicol} %

\usepackage{listings}
\usepackage{xcolor}

\usepackage{tablefootnote}
\usepackage{hyperref}

\lstdefinelanguage{json}{
    basicstyle=\ttfamily\footnotesize,
    numbers=left,
    numberstyle=\tiny\color{gray},
    stepnumber=1,
    numbersep=8pt,
    showstringspaces=false,
    breaklines=true,
    frame=lines,
    backgroundcolor=\color{gray!10},
    morestring=[b]",
    literate=
     *{0}{{{\color{black}0}}}{1}
      {1}{{{\color{black}1}}}{1}
      {2}{{{\color{black}2}}}{1}
      {3}{{{\color{black}3}}}{1}
      {4}{{{\color{black}4}}}{1}
      {5}{{{\color{black}5}}}{1}
      {6}{{{\color{black}6}}}{1}
      {7}{{{\color{black}7}}}{1}
      {8}{{{\color{black}8}}}{1}
      {9}{{{\color{black}9}}}{1}
}

\makeatletter
\def\@BTrule[#1]{%
  \ifx\longtable\undefined
    \let\@BTswitch\@BTnormal
  \else\ifx\hline\LT@hline
    \nobreak
    \let\@BTswitch\@BLTrule
  \else
     \let\@BTswitch\@BTnormal
  \fi\fi
  \global\@thisrulewidth=#1\relax
  \ifnum\@thisruleclass=\tw@\vskip\@aboverulesep\else
  \ifnum\@lastruleclass=\z@\vskip\@aboverulesep\else
  \ifnum\@lastruleclass=\@ne\vskip\doublerulesep\fi\fi\fi
  \@BTswitch}
\makeatother

\addto\extrasenglish{
}

 {\begin{list}{}%
         {\setlength{\leftmargin}{#1}}%
         \item[]%
 }
 {\end{list}}

\reportnumber{001} %

\renewcommand{\today}{}

\title{\centering PaddleOCR-VL: Boosting Multilingual Document Parsing \\ via a 0.9B Ultra-Compact Vision-Language Model }

\author[*]{
\small
Cheng Cui, Ting Sun, Suyin Liang, Tingquan Gao, Zelun Zhang, Jiaxuan Liu,
\vspace{-0.4cm}
\\
\small
 Xueqing Wang, Changda Zhou, Hongen Liu, Manhui Lin, Yue Zhang, Yubo Zhang, 
\\
\small
Handong Zheng, Jing Zhang, Jun Zhang, Yi Liu, Dianhai Yu, Yanjun Ma
\vspace{0.2cm}
\\
\small
\textbf{PaddlePaddle Team, Baidu Inc.}
\\
\small
\texttt{paddleocr@baidu.com}
\vspace{0.2cm}
  \\
  {\small
  \raggedright{  
  \small
  \hspace{6.55em}  
  \includegraphics[height=0.9em]{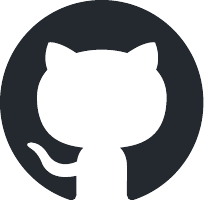} \textbf{Source Code}: \url{https://github.com/PaddlePaddle/PaddleOCR} \\
  \hspace{-1.2em}  
  \small
  \hspace{1.9em}  
  \includegraphics[height=1.0em]{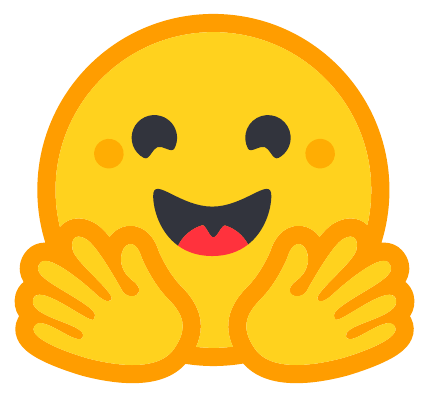} \textbf{Models \& Online Demo}: \url{https://huggingface.co/PaddlePaddle} \\
  \small
  }

  }
}

\renewcommand{\phi}{\varphi}

\renewcommand{\epsilon}{\varepsilon}
\renewcommand{\imath}{\mathrm{i}}

\newlength{\restsubwidth}
\newlength{\restsubheight}
\newlength{\restsubmoreheight}
\newcommand{\rest}[2]{%
        \settowidth{\restsubwidth}{\ensuremath{#2}}
        \settoheight{\restsubheight}{\ensuremath{{}_{#2}}}
        \ensuremath{{#1\hskip 0.5pt}_{\vrule\kern2pt\parbox[b][%
        4pt][b]{\the\restsubwidth}{%
                        \ensuremath{{}_{#2}}}}}
        }

\begin{abstract}

\vspace{-0.5cm} 

In this report, we propose PaddleOCR-VL, a SOTA and resource-efficient model tailored for document parsing. Its core component is PaddleOCR-VL-0.9B, a compact yet powerful vision-language model (VLM) that integrates a NaViT-style dynamic resolution visual encoder with the ERNIE-4.5-0.3B language model to enable accurate element recognition. This innovative model efficiently supports 109 languages and excels in recognizing complex elements (e.g., text, tables, formulas, and charts), while maintaining minimal resource consumption. Through comprehensive evaluations on widely used public benchmarks and in-house benchmarks, PaddleOCR-VL achieves SOTA performance in both page-level document parsing and element-level recognition. It significantly outperforms existing solutions, exhibits strong competitiveness against top-tier VLMs, and delivers fast inference speeds. These strengths make it highly suitable for practical deployment in real-world scenarios.

\end{abstract}

\begin{document}

\maketitle
\vspace{0cm} 
\begin{figure}[h]
\centering

\includegraphics[width=1\textwidth]{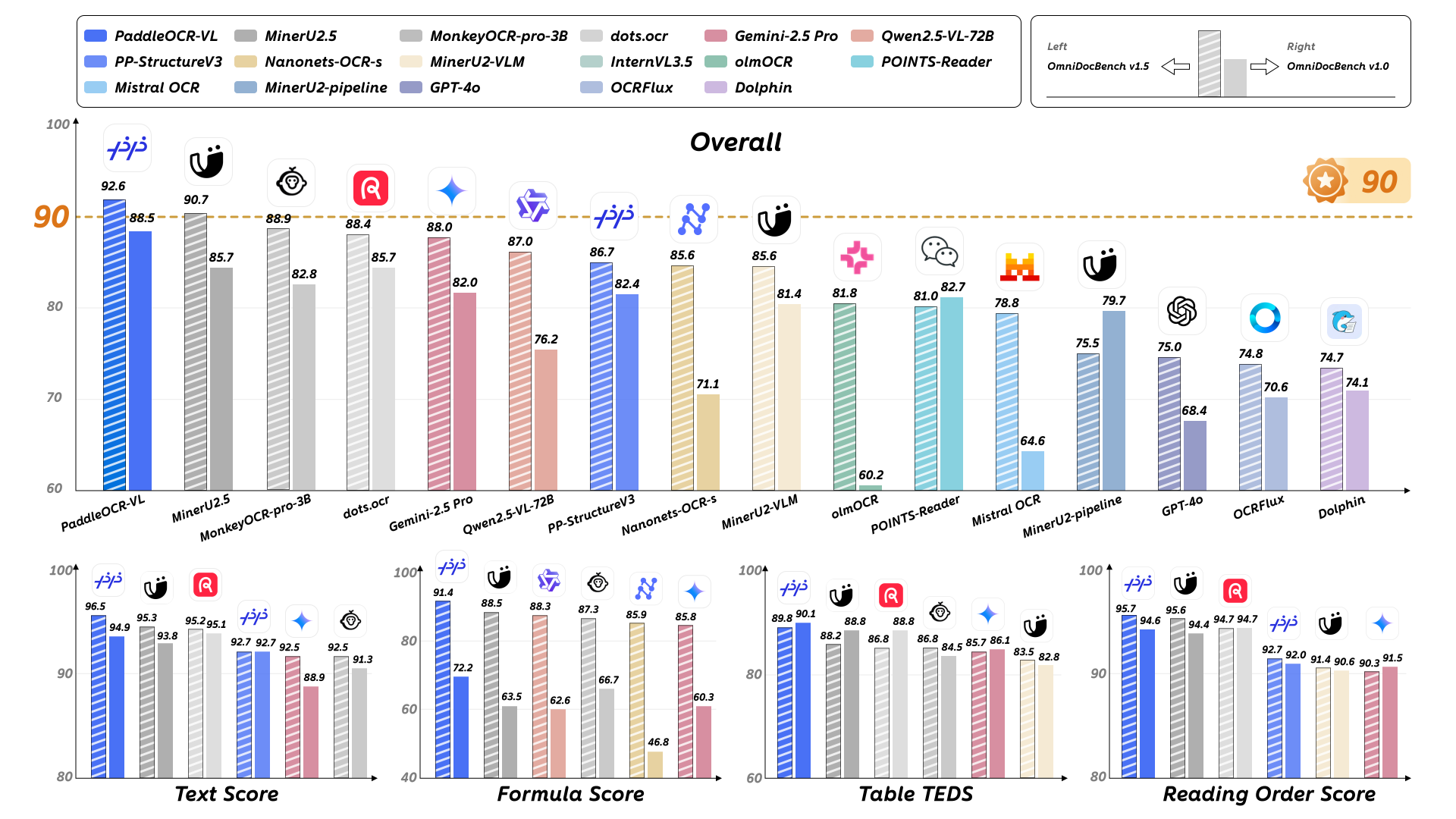} %

\caption{
    Performance of PaddleOCR-VL on OmniDocBench v1.0 and v1.5. 
}
\label{fig:dataset}
\end{figure}

\newpage
\setlength{\cftbeforesecskip}{6pt}   
\setlength{\cftbeforesubsecskip}{4pt} 
\setcounter{tocdepth}{2}
\tableofcontents

\newpage

\section{Introduction}

Documents serve as core information carriers, with their complexity and volume growing at an exponential rate, making document parsing an indispensable key technology. The primary goal of document parsing~\cite{li2025monkeyocr, niu2025mineru2, feng2025dolphin, liu2025points} is to enable deep structural and semantic understanding of a document’s layout. Specifically, it involves recognizing distinct text blocks and columns, distinguishing formulas, tables, charts, and images, determining the correct reading order, and detecting key elements (e.g., footnotes and image captions); these capabilities collectively lay a solid foundation for efficient information retrieval and data management. Furthermore, advanced document parsing enables large language models (LLMs)~\cite{ernie2025technicalreport, yang2025qwen3,achiam2023gpt}, especially when combined with Retrieval-Augmented Generation (RAG)~\cite{lewis2020retrieval}, to access high-quality knowledge and enhance their practical applications.

The inherent complexity of modern documents presents unique challenges: they often combine dense text, complex tables or chart, mathematical expressions, multiple languages and handwritten texts, with diverse layout structures. Recent research~\cite{li2025monkeyocr,wang2024mineru,cui2025paddleocr,Docling_Team_Docling,poznanski2025olmocr} in the field of document parsing primarily following two technological  approaches. The first approach~\cite{wang2024mineru,cui2025paddleocr} employs pipeline methodologies based on specialized, modular expert models. Although these methods offer strong performance, they are increasingly hindered by integration complexity, cumulative error propagation, and inherent limitations when handling highly complex documents. Secondly, end-to-end approaches~\cite{poznanski2025olmocr, nassar2025smoldocling, MinerU2} leveraging multimodal models aim to simplify the workflow and enable joint optimization. However, these methods often struggle with correct text order and can even generate hallucinations when faced with lengthy or complex layouts, while also incurring substantial computational overhead for long sequence outputs, thereby restricting their practical deployment.

To address these advancements and challenges, we present PaddleOCR-VL, a high-performance, resource-efficient document parsing solution based on a vision-language model. This innovation paves the way for the widespread application of multimodal document parsing, particularly in resource-constrained environments. 
PaddleOCR-VL combines a robust layout analysis model with a compact yet powerful vision-language model, PaddleOCR-VL-0.9B. 

Firstly, PaddleOCR-VL performs layout detection and reading order prediction to obtain the positional coordinates and reading order of elements (text blocks, tables, formulas, and charts). Compared to multimodal methods that rely on grounding and sequence output (e.g., MinerU2.5~\cite{niu2025mineru2}, Dolphin~\cite{feng2025dolphin}), our method offers faster inference speeds, lower training costs, and easier extensibility for new layout categories. Subsequently, the elements are segmented based on their positions and fed into PaddleOCR-VL-0.9B for recognition. This vision-language model is specifically designed for resource-efficient inference and excels at element recognition within document parsing. By integrating a NaViT-style~\cite{dehghani2023patch} dynamic high-resolution visual encoder with the lightweight ERNIE-4.5-0.3B~\cite{ernie2025technicalreport} language model, we have significantly enhanced the model’s dense text recognition capabilities and decoding efficiency. 

To train a powerful multimodal model, we have developed a high-quality training data construction pipeline. We collected over 30 million training samples through public data acquisition and data synthesis. We meticulously designed prompt engineering to guide the automatic labeling by general large models, based on the recognition results of expert models. Simultaneously, We performed data cleaning to remove low-quality or inconsistent annotations, such as those caused by model hallucinations. We designed an evaluation engine, which is an assessment collection that categorizes each element into more detailed categories. Through this automated evaluation, we can analyze the current model's training performance across different types. This allows us to conduct targeted hard sample mining based on element types and to construct similar challenging examples through data synthesis. Finally, we incorporated manual annotation for a small number of corner cases to complete the construction of the training data.

Comprehensive benchmarking on the public benchmarks, including OmniDocBench v1.0, v1.5~\cite{ouyang2025omnidocbench} and olmOCR-Bench~\cite{poznanski2025olmocr}, and in-house ones
demonstrate that PaddleOCR-VL achieves SOTA performance in document parsing task, significantly outperforming existing pipeline-based solutions and exhibiting strong competitiveness against leading vision-language models (VLMs). Moreover, PaddleOCR-VL is optimized for efficiency, delivering substantially lower latency and higher throughput than competing approaches.

PaddleOCR-VL actively addresses current challenges in document processing with a high-performance, resource-efficient multimodal document parsing solution. Its key contributions include:

\begin{itemize}

    \item \textbf{Compact yet Powerful VLM Architecture:} We present a novel vision-language model that is specifically designed for resource-efficient inference, achieving outstanding performance in element recognition. By integrating a NaViT-style dynamic high-resolution visual encoder with the lightweight ERNIE-4.5-0.3B language model, we significantly enhance the model’s recognition capabilities and decoding efficiency. This integration maintains high accuracy while reducing computational demands, making it well-suited for efficient and practical document processing applications.

    \item \textbf{High-quality Data Construction Methodology:} We propose a systematic and comprehensive methodology for constructing high-quality datasets, providing a solid train data foundation for efficient and robust document parsing. This methodology not only enables us to construct high-quality data on demand, but also provides a new perspective on the automated generation of high-quality data.

    \item \textbf{SOTA Performance Document Parsing:} PaddleOCR-VL achieves state-of-the-art performance in document parsing task. It excels in recognizing complex document elements, such as \textbf{text, tables, formulas, and charts}, making it suitable for a wide range of challenging content types, including handwritten text and historical documents. Supporting \textbf{109 languages}, including major global languages and those with diverse scripts like Russian, Arabic, and Hindi, PaddleOCR-VL is highly applicable to multilingual and globalized document processing scenarios.

\end{itemize}

\section{PaddleOCR-VL}

\subsection{Architecture}
\label{Architecture PaddleOCR-VL}

\begin{figure}[h]
\centering
\includegraphics[width=\linewidth]{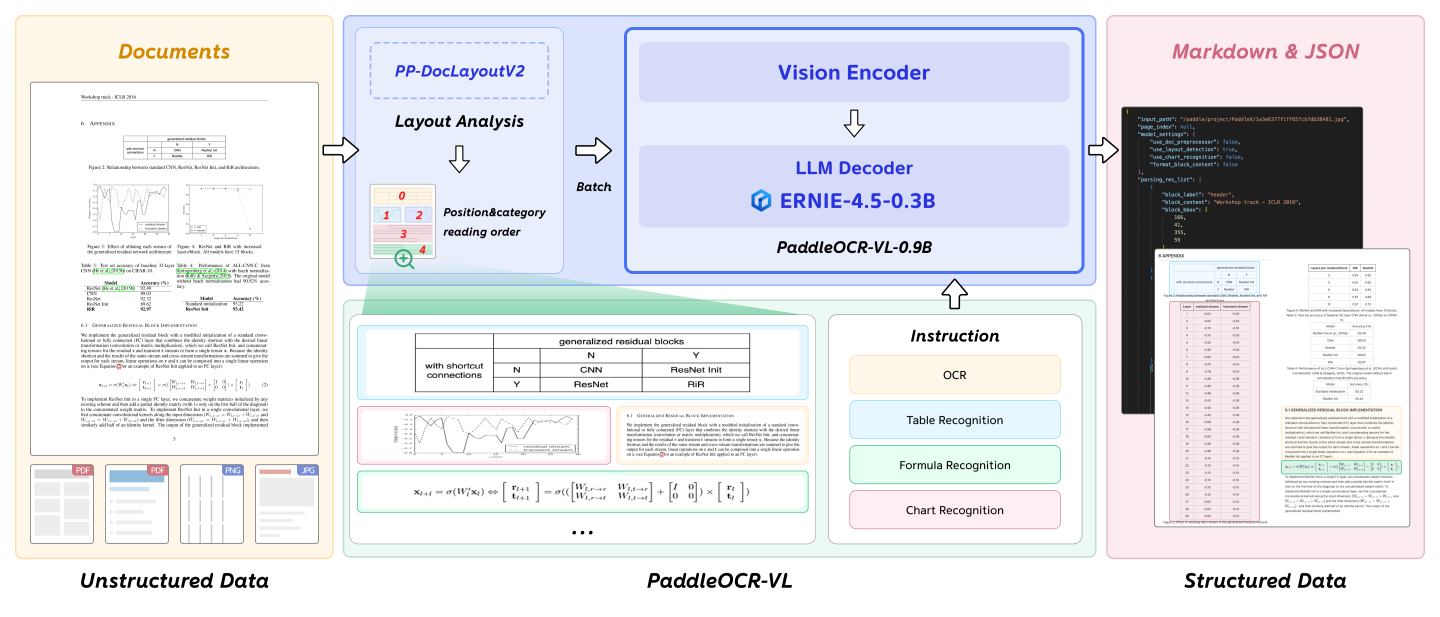} 

\caption{
    \centering
    The overview of PaddleOCR-VL.
}
\label{fig:model_overview}
\end{figure}

PaddleOCR-VL decomposes the complex task of document parsing into a two stages, as illustrated in Figure~\ref{fig:model_overview}. The first stage, PP-DocLayoutV2, is responsible for layout analysis, where it localizes semantic regions and predicts their reading order. Subsequently, the second stage, PaddleOCR-VL-0.9B, leverages these layout predictions to perform fine-grained recognition of diverse content, including text, tables, formulas, and charts. Finally, a lightweight post-processing module aggregates the outputs from both stages and formats the final document into structured Markdown and JSON.

\subsubsection{Layout Analysis}

Considering that end-to-end approaches based on VLM rely on long-sequence autoregressive processes, which result in high latency and memory consumption, and increase the risk of unstable layout analysis and hallucinations—problems that are particularly pronounced in multi-column or mixed text–graphic layouts—we employ a dedicated lightweight model for layout analysis, focusing specifically on element detection, classification, and reading order prediction.

Specifically, we decouple the layout analysis process by introducing an independent model, PP-DocLayoutV2, dedicated solely to this task. PP-DocLayoutV2 consists of an object detection model (RT-DETR~\cite{zhao2024detrs}) for elements localization and classification, as well as a lightweight pointer network~\cite{hou2024relation} with six transformer layers to accurately predict the reading order of layout elements. 

This separation enables us to fully leverage the advanced capabilities of the vision model, which typically requires lower input image resolution, and contains significantly fewer parameters. As a result, it achieves stable and accurate layout analysis, without the instability issues that may arise in end-to-end approaches.

\begin{figure}[h]
\centering
\includegraphics[width=\linewidth]{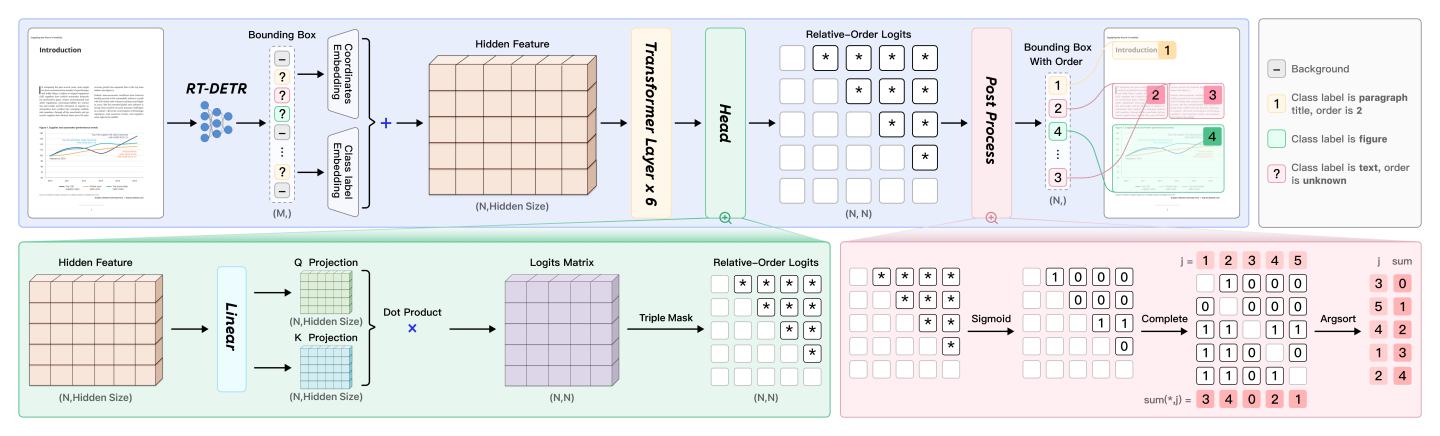} 

\caption{
    \centering
    Architecture of layout analysis model.
}
\label{fig:pp-doclayoutv2}
\end{figure}

Architecturally, PP‑DocLayoutV2 is composed of two sequentially connected networks, as shown in Figure~\ref{fig:pp-doclayoutv2}. The first is an RT‑DETR-based~\cite{zhao2024detrs} detection model that performs layout element detection and classification. The detected bounding boxes and class labels are then passed to a subsequent pointer network, which is responsible for ordering these layout elements.

Specifically, we first apply per-class thresholds to select foreground proposals for the ordering network. The selected proposals are embedded using absolute 2D positional encodings and class label embeddings. Additionally, the encoder attention incorporates a geometric bias mechanism from Relation‑DETR~\cite{hou2024relation} to explicitly model pairwise geometric relationships among elements. The pairwise relation head linearly projects element representations into query and key vectors, then computes bilinear similarities to produce pairwise logits, resulting in an $N \times N$ matrix that represents the relative order between each pair of elements. Finally, a deterministic win‑accumulation decoding algorithm recovers a topologically consistent reading order for the detected layout elements.

In comparison to other specialized models, such as LayoutReader~\cite{wang2021layoutreader}, our model achieves higher performance with fewer parameters by efficiently extending RT-DETR~\cite{zhao2024detrs} with a pointer network.

\subsubsection{Element-level Recognition}
\label{PaddleOCR-VL-0.9B arch}

\begin{figure}[h]
\centering
\includegraphics[width=\linewidth]{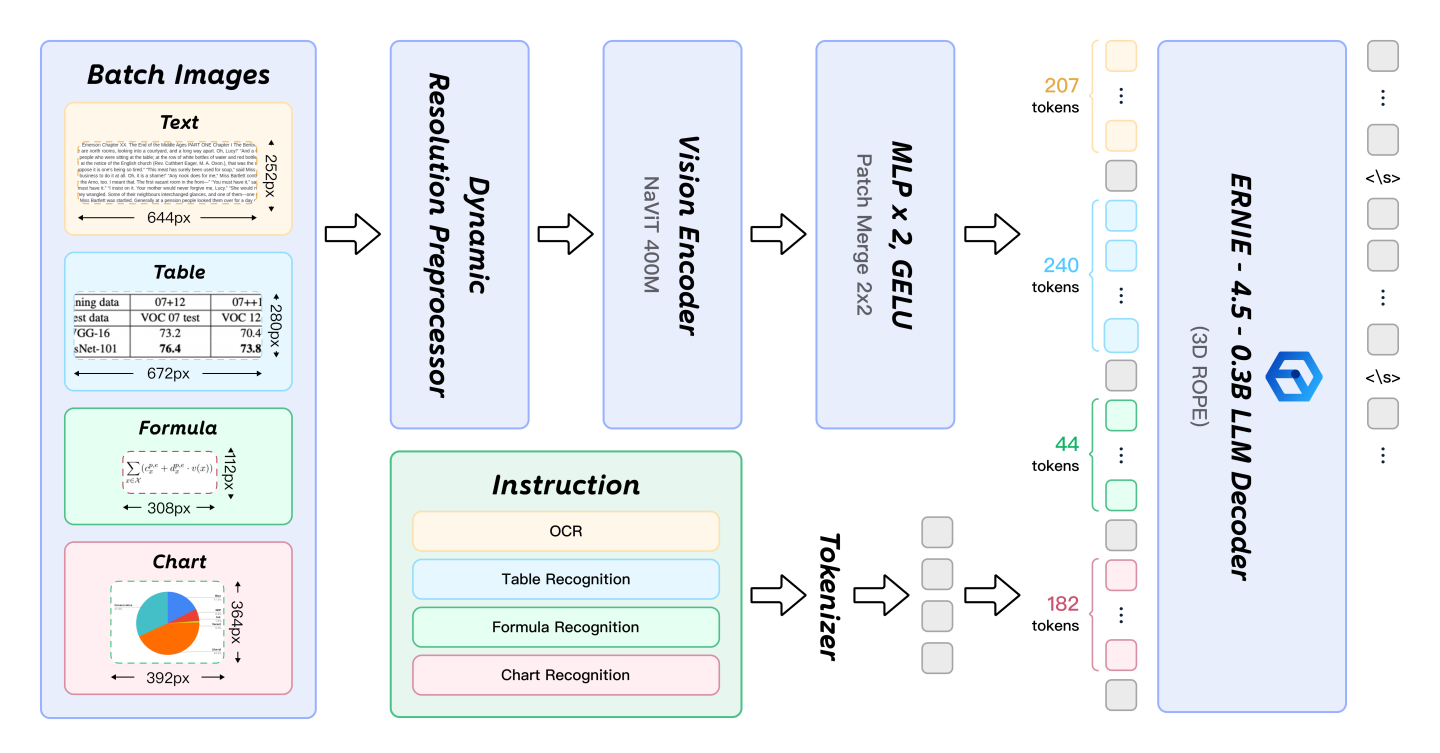} 

\caption{
    \centering
    Architecture of PaddleOCR-VL-0.9B.
}
\label{fig:model_vl}
\end{figure}

We systematically explore architecture configurations optimized for high accuracy and low computational overhead, and propose the PaddleOCR-VL-0.9B as shown in Figure~\ref{fig:model_vl}.

We adopted an architectural style inspired by LLaVA~\cite{liu2023visual}, integrating a pre-trained vision encoder with a dynamic resolution preprocessor, a randomly initialized 2-layer MLP projector, and a pre-trained large language model. Our architecture achieves a balance the scale of vision and language models to optimize performance in multi-elements recognition tasks.

Compared to earlier document parsing models based on fixed-resolution or tiling-based approaches~\cite{liu2025points,MinerU2,wei2024general}, our approach utilizes native dynamic high-resolution preprocessing. For the vision encoder, we employed a NaViT-style~\cite{dehghani2023patch} encoder initialized from Keye-VL’s~\cite{team2025kwai} vision model, which support native-resolution inputs. This design enables the vision-language model to handle images of arbitrary resolution without distortion, yielding fewer hallucinations and stronger performance on text-intensive tasks.

The projector is a randomly initialized 2-layer MLP with GELU~\cite{hendrycks2016gaussian} activation, incorporating a merge size of 2 to efficiently bridge visual features from the encoder to the language model's embedding space.

In auto-regressive language models, the entire sequence is generated by predicting one token at a time. This approach means that the size of the decoder is directly linked to the overall inference latency, so a smaller model will decode faster. With this in mind, we use the ERNIE-4.5-0.3B~\cite{ernie2025technicalreport} model, an open-source language model that balances a relatively small number of parameters with strong inference efficiency. In our implementation, we further enhance positional representation by incorporating a 3D-RoPE\cite{bai2025qwen2}.

The combination of NaViT~\cite{dehghani2023patch} with ERNIE-4.5-0.3B~\cite{ernie2025technicalreport} has led to significant performance improvements in documents parsing, achieving minimal memory usage and faster inference speed.

\subsection{Training Recipe}

 The following sections introduce the training details of these two modules: PP-DocLayoutV2 for layout analysis and PaddleOCR-VL-0.9B for element recognition.

\subsubsection{Layout Analysis}

\label{Layout Analysis}

We employ the PP-DocLayoutV2 model to perform layout element localization, classification, and reading order prediction. PP-DocLayoutV2 extends RT-DETR~\cite{zhao2024detrs} by incorporating an additional pointer network~\cite{hou2024relation}, which is responsible for predicting the reading order of detected elements. The training process adopts a two-stage strategy: we first train the core RT-DETR~\cite{zhao2024detrs} model for layout detection and classification. Afterward, we freeze its parameters and independently train the pointer network for reading order prediction.

For the first stage, we follow the training strategy of RT-DETR~\cite{zhao2024detrs}. Specifically, we initialize the model with PP-DocLayout\_Plus-L~\cite{sun2025pp} pretrained weights and train it for 100 epochs on our self-constructed dataset comprising over 20,000 high-quality samples.

For the second stage, specifically, the model outputs a matrix representing the pairwise ordering relationships between any two elements, and the Generalized Cross Entropy Loss \cite{zhang2018generalized} is computed with respect to the ground truth labels, as this loss function demonstrates increased robustness in scenarios where pre-annotated data are mixed into the dataset. We utilize a constant learning rate 2e-4 and the AdamW optimizer to train 200 epochs.

\subsubsection{Element-level Recognition}

\label{Element-level Recognition}

As described in Section \ref{PaddleOCR-VL-0.9B arch}, PaddleOCR-VL-0.9B consists of three modules: a vision encoder, a projector, and a language model. We adopt a post-adaptation strategy using pre-trained models. Specifically, the vision model is initialized with Keye-VL's weights, and the language model is initialized with ERNIE-4.5-0.3B's weights. The model is trained based on the ERNIEKit~\cite{ERNIEkit} repository and the training methodology for our VLM is divided into two stages, as outlined in Table \ref{training}.

\begin{table}[h]
\centering
 \fontsize{7}{7}\selectfont
\renewcommand{\arraystretch}{1.2}
\begin{tabular}{l|cc}
\toprule
\textbf{Stages} & \textbf{Stage 1} & \textbf{Stage 2} \\  \midrule
Training Samples & 29M & 2.7M \\
Max Resolution & 1280 $\times$ 28 $\times$ 28 & 2048 $\times$ 28 $\times$ 28 \\
Sequence length & 16384 & 16384 \\
Trainable components & All & All \\
Batch sizes & 128 & 128 \\
Data Augmentation & Yes & Yes \\
Maximum LR & $5 \times 10^{-5}$ & $5 \times 10^{-6}$ \\
Minimum LR & $5 \times 10^{-6}$ & $5 \times 10^{-7}$ \\
Epoch & 1 & 2 \\ \bottomrule
\end{tabular}
\caption{Training settings in stage 1 and stage 2.}
\label{training}
\end{table}

\textbf{Stage 1}: The initial stage focuses on pre-training alignment, where the model learns to associate visual information from images with corresponding textual representations. This crucial step is performed on a massive dataset comprising 29 million high-quality image-text pairs. During this phase, which runs for one epoch, the model is trained to establish a coherent understanding between diverse visual inputs and their semantic textual content. The training utilizes a batch size of 128, a sequence length of 16384, and supports a maximum image resolution of 1280$\times$28$\times$28, with data augmentation enabled to improve robustness. For optimization, the learning rate is scheduled between a maximum of $5 \times 10^{-5}$ and a minimum of $5 \times 10^{-6}$. The primary objective is to align the feature spaces of the vision encoder and the language model, enabling them to jointly process multimodal information effectively. This large-scale pre-training allows the model to capture intricate visual patterns, common textual structures, and their interdependencies across a vast range of contexts, laying a strong foundation for subsequent specialized tasks.

\textbf{Stage 2}: Following pre-training, the model undergoes \textbf{instruction fine-tuning} to adapt its general multimodal understanding to specific downstream elements recognition tasks. This stage utilizes a meticulously curated dataset of 2.7 million samples, which is intentionally designed to be highly rich and diverse in its distribution. The training is conducted over two epochs, maintaining the 128 batch size and 16384 sequence length, but increasing the maximum resolution to 2048$\times$28$\times$28 to handle more detailed inputs. A finer learning rate is adopted, with the maximum and minimum values set to $5 \times 10^{-6}$ and $5 \times 10^{-7}$, to carefully adjust the model on specialized data. The richness of this dataset encompasses a wide variety of document types, languages, writing systems, and visual complexities pertinent to real-world scenarios. During this fine-tuning phase, the model is trained with explicit instructions for four types of tasks:
\begin{enumerate}[label=\arabic*.]
    \item \textbf{OCR:} This task fine-tunes the model to accurately identify and extract textual content from images, encompassing individual characters, words, text lines, text blocks and simple layout structure of page-level texts.
    \item \textbf{Table Recognition:} The model learns to parse tabular structures within documents. This involves accurately extracting cell contents, identifying rows and columns, and recognize the logical relationships between different table elements, ultimately generating structured representations based on OTSL~\cite{lysak2023optimized} format.
    \item \textbf{Formula Recognition:} This instruction focuses on enabling the model to recognize and interpret mathematical and scientific formulas. It aims to convert their visual representation into a structured \LaTeX format and distinguishes between inline \textbackslash(...\textbackslash) and display \textbackslash[...\textbackslash]  equations.
    \item \textbf{Chart Recognition:} This task trains the model to recognition information from various types of charts, such as bar charts, line graphs, and pie charts and convert Markdown format tables.
\end{enumerate}

\section{Dataset}

To build our high-quality and diverse training dataset, we propose a systematic methodology for constructing such datasets. As illustrated in Figure \ref{fig:dataset}, we gather a diverse set of data from multiple sources to ensure comprehensive coverage. High-quality labels are then generated through automated annotation using large models, which guarantees precision and consistency. Additionally, we refine the training data by integrating challenging examples, which enhances the model’s performance and robustness. Each of these crucial steps is detailed in the following sections.

\begin{figure}[h]
\centering
\includegraphics[width=\linewidth]{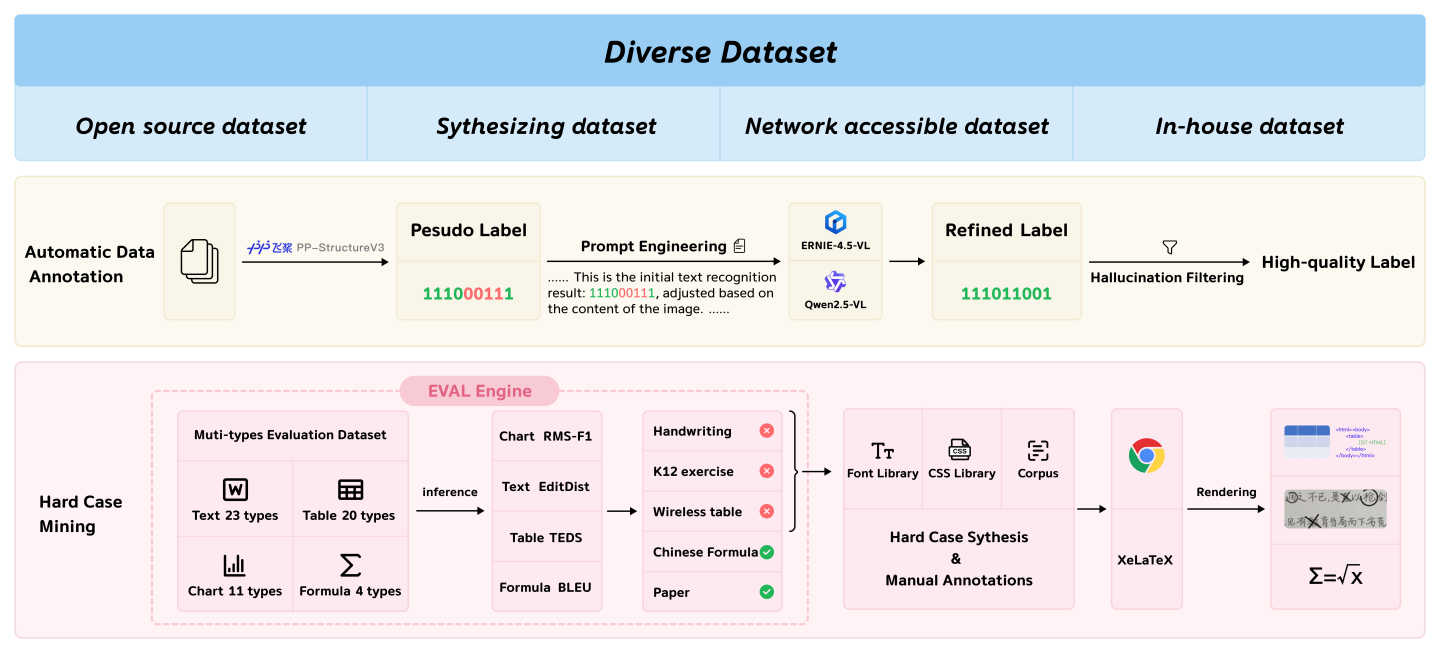} 

\caption{
    \centering
    The construction process of training data for PaddleOCR-VL-0.9B.
}
\label{fig:dataset}
\end{figure}

\subsection{Data Curation}
\label{Curation}

To ensure the breadth and diversity of the dataset, data is collected from four main sources: open-source dataset, synthesizing dataset, network accessible dataset, and in-house dataset.

\begin{enumerate}[label=\arabic*.]
     \item \textbf{Open Source Dataset:} As the foundation of our dataset, we systematically aggregated and curated a wide array of established public datasets. For textual content, we sourced data from the canonical dataset CASIA-HWDB~\cite{liu2011casia}. Our mathematical expression data is derived from UniMER-1M~\cite{unimernet} and MathWriting~\cite{MathWriting}. To ensure comprehensive coverage of data visualizations, we incorporated a rich spectrum of chart and graph datasets, including ChartQA~\cite{chartqa}, PlotQA~\cite{plotqa}, Chart2Text~\cite{chart2text}, DVQA~\cite{dvqa}, Unichart~\cite{masry2023unichart}, Beagle~\cite{beagle}, ChartINFO~\cite{chartinfo}, visText~\cite{tang2023vistext}, and ExcelChart~\cite{excelchart}. Each of these sources underwent an initial filtering and cleaning protocol to rectify or discard noisy and low-quality annotations.
    \item \textbf{Data Synthesizing Dataset:}  Due to the naturally imbalanced distribution of public data, we employed a data synthesizing strategy to produce large volumes of missing data types at low cost, providing our proposed model with the unbiased document parsing performance.
    \item \textbf{Network Accessible Dataset:}  To improve model generalization and robustness against the complexities of unstructured real-world documents, we amassed an extensive corpus of publicly accessible data harvested from the Internet. This public collection was deliberately curated to encompass a rich spectrum of document types and visual styles. It includes academic papers, newspapers, formal scientific journal articles, scanned handwritten documents, diverse examination papers, and slides, etc. The integration of these varied sources proved instrumental in significantly broadening the stylistic, structural, and domain diversity of our training data, thereby mitigating the risk of overfitting to clean, canonical datasets.
    \item \textbf{In-house Dataset:} Through years of research in the field of OCR, we have accumulated extensive datasets with diverse data types across all tasks of document parsing. We incorporate all in-house datasets into training with precisely controlled proportions, which have become unnecessary factors that enable our models to achieve outstanding performance.
\end{enumerate}

\subsection{Automatic Data Annotation}
\label{Automatic Data Annotation}

After acquiring the raw data, we utilize an automatic data annotations process for large-scale labeling. Initially, we employ the expert model, PP-StructureV3, to conduct preliminary processing on the data, generating pseudo labels that may contain some inaccuracies. Subsequently, through prompt engineering, we create prompts that include the original images and their associated pseudo labels, which are then submitted to more advanced multimodal large language models, ERNIE-4.5-VL~\cite{ernie2025technicalreport} and Qwen2.5VL~\cite{bai2025qwen2}. These sophisticated models refine and enhance the initial results by analyzing the image content, resulting in improved labels. Finally, to ensure the quality of the labels, the system performs a hallucination filtering step, which eliminates any potentially incorrect content generated by the large models, thereby producing reliable and high-quality labels.

\subsection{Hard Cases Mining}
\label{Hard Cases Mining}

To overcome performance bottlenecks in specific complex scenarios, we propose a hard case mining process for targeted performance improvement. We firstly develop a eval engine for various types. We created substantial evaluation data with precisely labeled data obtained through manual annotation. Theses evaluation datasets are categorized into several types: text data includes 23 categories such as Chinese, English, printed, handwritten, Japanese, Latin, and emojis; table data includes 20 categories such as limited tables, unlimited tables, handwritten tables, checklists, invoices, and rotated tables; formula data includes 4 categories such as Chinese and English formulas, handwritten and printed, simple, and complex; chart data includes 11 categories such as Chinese and English charts, line charts, and bar charts, sourced from diverse origins to cover different document. By inference on this evaluation set and using corresponding professional metrics (e.g., EditDist for Text, TEDS~\cite{teds} for Tables, RMS-F1~\cite{deplot} for Charts, and BLEU~\cite{bleu_score} for Formulas), we can accurately identify hard cases where the model performs poorly. Finally, for these identified weaknesses, the system utilizes a rich set of resources (such as Font Library, CSS Library, Corpus) and rendering tools (like XeLaTeX and web browsers) to synthetically generate a large volume of new, high-quality hard cases.

\section{Evaluation}

\label{sec:experiments} 

To thoroughly assess the effectiveness of PaddleOCR-VL, we compared it against leading general vision language models and specialized document parsing models across multiple public benchmarks and in-house benchmarks. We conducted comprehensive performance comparisons in two aspects: page-level document parsing and element-level recognition, which are detailed in Sections \ref{Page-level Evaluation} and \ref{subsec:element_level_evaluation}. Page-level involves analyzing entire pages of a document to parsing their overall content, structure and layout, while element-level is dedicated exclusively to assessing the recognition of specific elements, such as text, tables, formulas, and charts, within the document.

\subsection{Page-level Evaluation}
\label{Page-level Evaluation}

This section details the evaluation of end-to-end document parsing capabilities using the following three benchmarks, aiming to measure its overall performance in real-world document scenarios. 

\paragraph{OmniDocBench v1.5} To comprehensively evaluate the document parsing capabilities, we conducted extensive experiments on the OmniDocBench v1.5~\cite{niu2025mineru2} benchmark. It is an expansion of version v1.0, adding 374 new documents for a total of 1,355 document pages. It features a more balanced distribution of data in both Chinese and English, as well as a richer inclusion of formulas and other elements. The evaluation method has been updated, with formulas assessed using the CDM method. The overall metric is a weighted combination of the metrics for text, formulas, and tables. 

Table~\ref{tab:omni15_performance} demonstrate that PaddleOCR-VL achieves SOTA performance, outperforming existing pipeline tools, general VLMs, and other specialized document parsing models across all key metrics. Specifically, our model achieves a top-ranking overall score of 92.86, surpassing the next best model, MinerU2.5-1.2B (90.67). Moreover, our model establishes new SOTA results in the sub-tasks, including the lowest Text-Edit distance~\cite{lcvenshtcin1966binary} of 0.035, the highest Formula-CDM score of 91.22, the leading scores of 90.89 and 94.76 in Table-TEDS and Table-TEDS-S, and the best readering ordering scores of 0.043, respectively. These results underscore its superior accuracy in text recognition, formula recognition, and complex table structure analysis.

\begin{table}[H]
    \centering

    \resizebox{\textwidth}{!}{%
    \renewcommand{\arraystretch}{1.2}
    \begin{tabular}{l|ll|c|c c c c c}
        \toprule
        \textbf{Model Type} & \textbf{Methods} & \textbf{Parameters} & \textbf{Overall$\uparrow$} & \textbf{Text\textsuperscript{Edit}$\downarrow$} & \textbf{Formula\textsuperscript{CDM}$\uparrow$} & \textbf{Table\textsuperscript{TEDS}$\uparrow$} & \textbf{Table\textsuperscript{TEDS-S}$\uparrow$} & \textbf{Reading Order\textsuperscript{Edit}$\downarrow$} \\    \midrule
        \multirow{3}{*}{\textbf{Pipeline Tools}} & Marker-1.8.2~\cite{vik2024marker} & - & 71.30 & 0.206 & 76.66 & 57.88 & 71.17 & 0.250 \\
        & Mineru2-pipeline~\cite{MinerU2} & - & 75.51 & 0.209 & 76.55 & 70.90 & 79.11 & 0.225 \\
        & PP-StructureV3~\cite{cui2025paddleocr} & - & 86.73 & 0.073 & 85.79 & 81.68 & 89.48 & 0.073 \\
        \midrule
        \multirow{5}{*}{\textbf{General VLMs}} & GPT-4o~\cite{achiam2023gpt} & - & 75.02 & 0.217 & 79.70 & 67.07 & 76.09 & 0.148 \\
        & InternVL3-76B~\cite{zhu2025internvl3} & 76B & 80.33 & 0.131 & 83.42 & 70.64 & 77.74 & 0.113 \\
        & InternVL3.5-241B~\cite{wang2025internvl35} & 241B & 82.67 & 0.142 & 87.23 & 75.00 & 81.28 & 0.125 \\
        & Qwen2.5-VL-72B~\cite{bai2025qwen2} & 72B & 87.02 & 0.094 & 88.27 & 82.15 & 86.22 & 0.102 \\
        & Gemini-2.5 Pro~\cite{gemini25} & - & 88.03 & 0.075 & 85.82 & 85.71 & 90.29 & 0.097 \\
        \midrule
        \multirow{11}{*}{\textbf{Specialized VLMs}} & Dolphin~\cite{feng2025dolphin} & 322M & 74.67 & 0.125 & 67.85 & 68.70 & 77.77 & 0.124 \\
        & OCRFlux-3B~\cite{OCRFlux2025} & 3B & 74.82 & 0.193 & 68.03 & 75.75 & 80.23 & 0.202 \\
        & Mistral OCR~\cite{mistral} & - & 78.83 & 0.164 & 82.84 & 70.03 & 78.04 & 0.144 \\
        & POINTS-Reader~\cite{liu2025points} & 3B & 80.98 & 0.134 & 79.20 & 77.13 & 81.66 & 0.145 \\
        & olmOCR-7B~\cite{poznanski2025olmocr} & 7B & 81.79 & 0.096 & 86.04 & 68.92 & 74.77 & 0.121 \\
        & MinerU2-VLM~\cite{MinerU2} & 0.9B & 85.56 & 0.078 & 80.95 & 83.54 & 87.66 & 0.086 \\
        & Nanonets-OCR-s~\cite{Nanonets-OCR-S} & 3B & 85.59 & 0.093 & 85.90 & 80.14 & 85.57 & 0.108 \\
           & MonkeyOCR-pro-1.2B~\cite{li2025monkeyocr} & 1.9B & 86.96 & 0.084 & 85.02 & 84.24 & 89.02 & 0.130 \\
        & MonkeyOCR-3B~\cite{li2025monkeyocr} & 3.7B & 87.13 & 0.075 & 87.45 & 81.39 & 85.92 & 0.129 \\
        & dots.ocr~\cite{dotsocr} & 3B & 88.41 & 0.048 & 83.22 & 86.78 & 90.62 & 0.053 \\
      & MonkeyOCR-pro-3B~\cite{li2025monkeyocr} & 3.7B & 88.85 & 0.075 & 87.25 & 86.78 & 90.63 & 0.128 \\
        & MinerU2.5~\cite{niu2025mineru2} & 1.2B & \cellcolor{cyan!15}\underline{90.67} & \cellcolor{cyan!15}\underline{0.047} & \cellcolor{cyan!15}\underline{88.46} & \cellcolor{cyan!15}\underline{88.22} & \cellcolor{cyan!15}\underline{92.38} & \cellcolor{cyan!15}\underline{0.044} \\ 

        & \textbf{PaddleOCR-VL} &0.9B & \cellcolor{red!15}\textbf{92.86} & \cellcolor{red!15}\textbf{0.035} & \cellcolor{red!15}\textbf{91.22} & \cellcolor{red!15}\textbf{90.89} & \cellcolor{red!15}\textbf{94.76} & \cellcolor{red!15}\textbf{0.043} \\
        \bottomrule
    \end{tabular}%
    }
   \caption{Comprehensive evaluation of document parsing on OmniDocBench v1.5. Results are reported by OmniDocBench~\cite{ouyang2025omnidocbench} unless Ours.}
       \label{tab:omni15_performance}
\end{table}

\paragraph{OmniDocBench v1.0} A publicly available benchmark dataset specifically is designed to evaluate real-world document parsing capabilities. It comprises 981 PDF pages, spanning 9 distinct document types, 4 layout styles, and 3 language categories. 

Based on the experimental results presented in Table \ref{tab:omni_performance}, PaddleOCR-VL demonstrates superior performance with an average overall edit distance of 0.115, demonstrating its superior capability in document parsing. The model excels in formula edit distance (0.241 EN, 0.316 ZH), and achieves the SOTA performance (0.062) and a comparable SOTA performance (0.041) for Chinese and English text edit distance respectively, showcasing its accuracy in handling textual and formulaic data. Although the model exhibits slightly lower performance in the English Table TEDS (88.0), this can be largely attributed to typo-related annotation errors in OmniDocBench v1.0. Nevertheless, it demonstrates a clear advantage in the Chinese Table TEDS (92.14). Regarding the reading order edit distance, the model achieves the best performance in Chinese (0.063) and a comparable SOTA result in English (0.045), emphasizing its capability to maintain structural integrity and logical document flow.

\begin{table}[H]
  \centering
  \resizebox{\textwidth}{!}{%
  \renewcommand{\arraystretch}{1.2}\begin{tabular}{l|l| c| cc|cc|cc|cc|cc|ccc} 
   
 \toprule
    \multirow{2}{*}{\textbf{Method Type}} & \multirow{2}{*}{\textbf{Methods}} & \multirow{2}{*}{\textbf{\shortstack{AvgOverall\textsuperscript{Edit} $\downarrow$}}} & \multicolumn{2}{c|}{\textbf{Overall\textsuperscript{Edit}$\downarrow$}} & \multicolumn{2}{c|}{\textbf{Text\textsuperscript{Edit}$\downarrow$}} & \multicolumn{2}{c|}{\textbf{Formula\textsuperscript{Edit}$\downarrow$}} & \multicolumn{2}{c|}{\textbf{Table\textsuperscript{TEDS}$\uparrow$}} & \multicolumn{2}{c|}{\textbf{Table\textsuperscript{Edit}$\downarrow$}} & \multicolumn{2}{c}{\textbf{Reading Order\textsuperscript{Edit}$\downarrow$}} \\
    
 \cline{4-5}  \cline{6-7} \cline{8-9} \cline{10-11} \cline{12-13} \cline{14-15}
        & & & \textbf{EN} & \textbf{ZH} & \textbf{EN} & \textbf{ZH} & \textbf{EN} & \textbf{ZH} & \textbf{EN} & \textbf{ZH} & \textbf{EN} & \textbf{ZH} & \textbf{EN} & \textbf{ZH} \\
    \midrule
    \multirow{9}{*}{\textbf{Pipeline Tools}} 
      & Docling-2.14.0~\cite{Docling_Team_Docling} & 0.749 & 0.589 & 0.909 & 0.416 & 0.987 & 0.999 & 1 & 61.3 & 25.0 & 0.627 & 0.810 & 0.313 & 0.837 \\
         & OpenParse-0.7.0~\cite{open-parse} & 0.730 & 0.646 & 0.814 & 0.681 & 0.974 & 0.996 & 1 & 64.8 & 27.5 & 0.284 & 0.639 & 0.595 & 0.641 \\
        & Unstructured-0.17.2~\cite{unstructured} & 0.651 & 0.586 & 0.716 & 0.198 & 0.481 & 0.999 & 1 & 0 & 0.1 & 1 & 0.998 & 0.145 & 0.387 \\
        & Pix2Text-1.1.2.3~\cite{Pix2Text} & 0.424 & 0.320 & 0.528 & 0.138 & 0.356 & 0.276 & 0.611 & 73.6 & 66.2 & 0.584 & 0.645 & 0.281 & 0.499 \\
   
    & Marker-1.7.1~\cite{vik2024marker} & 0.397 & 0.296 & 0.497 & 0.085 & 0.293 & 0.374 & 0.688 & 67.6 & 54.0 & 0.609 & 0.678 & 0.116 & 0.329 \\
    & Mathpix~\cite{mathpix} & 0.278 & 0.191 & 0.364 & 0.105 & 0.381 & 0.306 & 0.454 & 77.0 & 67.1 & 0.243 & 0.320 & 0.108 & 0.304 \\
     & MinerU-pipeline~\cite{wang2024mineru} & 0.203 & 0.162 & 0.244 & 0.072 & 0.111 & 0.313 & 0.581 & 77.4 & 79.5 & 0.166 & 0.150 & 0.097 & 0.136 \\
    & PP-StructureV3~\cite{cui2025paddleocr} & 0.176 & 0.145 & 0.206 & 0.058 & 0.088 & 0.295 & 0.535 & 77.2 & 83.9 & 0.159 & 0.109 & 0.069 & 0.091 \\
    \midrule
    \multirow{5}{*}{\textbf{General VLMs}}

    & InternVL2-76B~\cite{InternVL} & 0.442 & 0.440 & 0.443 & 0.353 & 0.290 & 0.543 & 0.701 & 63.0 & 60.2 & 0.547 & 0.555 & 0.317 & 0.228 \\
    & GPT-4o~\cite{achiam2023gpt} & 0.316 & 0.233 & 0.399 & 0.144 & 0.409 & 0.425 & 0.606 & 72.0 & 62.9 & 0.234 & 0.329 & 0.128 & 0.251 \\

    & InternVL3-78B~\cite{zhu2025internvl3} & 0.257 & 0.218 & 0.296 & 0.117 & 0.210 & 0.380 & 0.533 & 69.0 & 73.9 & 0.279 & 0.282 & 0.095 & 0.161 \\
   & Qwen2.5-VL-72B~\cite{bai2025qwen2} & 0.238 & 0.214 & 0.261 & 0.092 & 0.180 & 0.315 & 0.434 & 81.4 & 83.0 & 0.341 & 0.262 & 0.106 & 0.168 
    \\    & Gemini2.5-Pro~\cite{gemini25} & 0.180 & 0.148 & 0.212 & 0.055 & 0.168 & 0.356 & 0.439 & 85.8 & 86.4 & 0.130 & 0.119 & 0.049 & 0.121 \\
    \midrule
    \multirow{13}{*}{\textbf{Specialized VLMs}} 
        & Nougat~\cite{blecher2023nougat} & 0.713 & 0.452 & 0.973 & 0.365 & 0.998 & 0.488 & 0.941 & 39.9 & 0.0 & 0.572 & 1 & 0.382 & 0.954 \\
  
    & SmolDocling-256M~\cite{nassar2025smoldocling} & 0.655 & 0.493 & 0.816 & 0.262 & 0.838 & 0.753 & 0.997 & 44.9 & 16.5 & 0.729 & 0.907 & 0.227 & 0.522 \\
       & olmOCR-7B~\cite{poznanski2025olmocr} & 0.398 & 0.326 & 0.469 & 0.097 & 0.293 & 0.455 & 0.655 & 68.1 & 61.3 & 0.608 & 0.652 & 0.145 & 0.277 \\
 
     & GOT~\cite{wei2024general} & 0.349 & 0.287 & 0.411 & 0.189 & 0.315 & 0.360 & 0.528 & 53.2 & 47.2 & 0.459 & 0.520 & 0.141 & 0.280 \\
  & OCRFlux-3B~\cite{OCRFlux2025} & 0.294 & 0.238 & 0.349 & 0.112 & 0.256 & 0.447 & 0.716 & 69.0 & 80.0 & 0.269 & 0.162 & 0.126 & 0.263 \\
      & Nanonets-OCR-s~\cite{Nanonets-OCR-S} & 0.289 & 0.283 & 0.295 & 0.134 & 0.231 & 0.518 & 0.546 & 76.8 & 79.4 & 0.343 & 0.201 & 0.135 & 0.200 \\
    & Dolphin~\cite{feng2025dolphin} & 0.259 & 0.205 & 0.313 & 0.092 & 0.204 & 0.447 & 0.606 & 76.1 & 66.9 & 0.193 & 0.282 & 0.088 & 0.160 \\
    & MinerU2-VLM~\cite{MinerU2} & 0.186 & 0.133 & 0.238 & 0.045 & 0.115 & 0.273 & 0.506 & 82.1 & 83.4 & 0.150 & 0.209 & 0.066 & 0.122 \\

        & MonkeyOCR-pro-1.2B~\cite{li2025monkeyocr} & 0.184 & 0.146 & 0.221 & 0.068 & 0.118 & 0.272 & 0.452 & 81.3 & 85.5 & 0.149 & 0.134 & 0.093 & 0.179 \\
            
    & MonkeyOCR-pro-3B~\cite{li2025monkeyocr} & 0.172 & 0.138 & 0.206 & 0.067 & 0.107 & \cellcolor{cyan!15}\underline{0.246} & 0.421 & 81.5&  87.5 & 0.139 & 0.111 & 0.100 & 0.185 \\
    & dots.ocr~\cite{dotsocr} & 0.143 & 0.125 & \cellcolor{cyan!15}\underline{0.160} & \cellcolor{red!15} \textbf{0.032} & \cellcolor{cyan!15}\underline{0.066} & 0.329 & \cellcolor{cyan!15}\underline{0.416} & \cellcolor{red!15}\textbf {88.6} & 89.0 & 0.099 & 0.092 & \cellcolor{red!15}\textbf{0.040} & \cellcolor{cyan!15}\underline{0.067} \\

    & MinerU2.5~\cite{niu2025mineru2} & \cellcolor{cyan!15}\underline{0.143} & \cellcolor{cyan!15}\underline{0.111} & 0.174 & 0.050 & 0.074 & 0.258 & 0.473 & \cellcolor{cyan!15}\underline{88.3} & \cellcolor{cyan!15}\underline{89.2} & \cellcolor{red!15}\textbf {0.089} & \cellcolor{cyan!15}\underline{0.083} & \cellcolor{cyan!15}\underline {0.045} & 0.068 \\
    
    &  \textbf{PaddleOCR-VL}  & \cellcolor{red!15} \textbf{0.115} & \cellcolor{red!15}\textbf{0.105} & \cellcolor{red!15}\textbf{0.126} & \cellcolor{cyan!15}\underline{0.041} & \cellcolor{red!15}\textbf{0.062} & \cellcolor{red!15}\textbf{0.241} & \cellcolor{red!15}\textbf{0.316} & 88.0 & \cellcolor{red!15}\textbf{92.1} & \cellcolor{cyan!15}\underline{0.093} & \cellcolor{red!15}\textbf{0.062} &\cellcolor{cyan!15}\underline{0.045} & \cellcolor{red!15}\textbf{0.063} \\
    \bottomrule
  \end{tabular}%
}
  \caption{Comprehensive evaluation of document parsing on OmniDocBench v1.0. Results are reported by OmniDocBench~\cite{ouyang2025omnidocbench} unless MinerU2.5 and Ours.}
  \label{tab:omni_performance}
\end{table}

\paragraph{olmOCR-Bench} olmOCR-Bench~\cite{poznanski2025olmocr} includes 1,402 PDF documents and 7,010 test cases, addressing diverse document types and extraction challenges. It offers a detailed evaluation framework for PDF content extraction by assessing tools and models through simple, clear, and machine-verifiable unit tests. This approach avoids biased evaluations and soft metric comparisons, allowing for the detection of subtle but significant extraction errors.

Table \ref{tab:ocr_performance} highlights the outstanding performance of PaddleOCR-VL in the olmOCR-Bench evaluation, achieving the highest overall score of 80.0 ± 1.0. It excels in various categories, leading in ArXiv (85.7), Headers and Footers (97.0) and securing second place in Multi-column text (79.9), Long Tiny Text (85.7). These results highlight the proposed model's capability to effectively manage diverse document types, reinforcing its status as a top solution in document parsing and its reliability in complex OCR tasks.

\begin{table*}[h]
  \centering
 \setlength{\tabcolsep}{8pt}

  \resizebox{\textwidth}{!}{%
  \renewcommand{\arraystretch}{1.2}
  \begin{tabular}{l|c|cccccccc}
    \toprule
    \multirow{2}{*}{\textbf{Methods}} & \multicolumn{9}{c}{\textbf{Unit Test Pass Rate $\uparrow$}}\\
    \cline{2-10}&\textbf{Overall} &\textbf{ArXiv} & \textbf{Old Scans Math} & \textbf{Tables} & \textbf{Old Scans} & \textbf{Headers and Footers} & \textbf{Multi column} & \textbf{Long Tiny Text} & \textbf{Base}  \\
    \midrule
   GOT~\cite{wei2024general}&48.3 ± 1.1 & 52.7 & 52.0 & 0.2 & 22.1 & 93.6 & 42.0 & 29.9 & 94.0  \\
   Gemini Flash 2 (No Anchor)~\cite{gemini25}& 57.8 ± 1.1 & 32.1 & 56.3 & 61.4 & 27.8 & 48.0 & 58.7 & 84.4 & 94.0  \\
   
    MinerU-pipeline~\cite{wang2024mineru}& 61.5 ± 1.1  & 75.4 & 47.4 & 60.9 & 17.3 & \cellcolor{cyan!15}\underline{96.6} & 59.0 & 39.1 & 96.6 \\
       Gemini Flash 2 (Anchored)~\cite{gemini25} & 63.8 ± 1.2  & 54.5 & 56.1 & 72.1 & 34.2 & 64.7 & 61.5 & 71.5 & 95.6\\
    Nanonets-OCR-s~\cite{Nanonets-OCR-S}& 64.5 ± 1.1  & 67.0 & 68.6 & 77.7 & 39.5 & 40.7 & 69.9 & 53.4 & \cellcolor{cyan!15}\underline{99.3} \\
     Qwen2.5-VL-7B (No Anchor)~\cite{bai2025qwen2} & 65.5 ± 1.2  & 63.1 & 65.7 & 67.3 & 38.6 & 73.6 & 68.3 & 49.1 & 98.3\\
    GPT-4o (No Anchor)~\cite{achiam2023gpt} & 68.9 ± 1.1& 51.5 & \cellcolor{red!15}\textbf {75.5} & 69.1 & 40.9 & 94.2 & 68.9 & 54.1 & 96.7  \\
    GPT-4o (Anchored)~\cite{achiam2023gpt}& 69.9 ± 1.1 & 53.5 & \cellcolor{cyan!15}\underline{74.5} & 70.0 & 40.7 & 93.8 & 69.3 & 60.6 & 96.8  \\
   
    Marker-1.8.2~\cite{vik2024marker}&70.1 ± 1.1  & 76.0 & 57.9 & 57.6 & 27.8 & 84.9 & 72.9 & 84.6 & 99.1  \\
    
    olmOCR v0.1.75 (No Anchor)~\cite{poznanski2025olmocr}& 74.7 ± 1.1  & 71.5 & 71.4 & 71.4 & \cellcolor{red!15}\textbf {42.8} & 94.1 & 77.7 & 71.0 & 97.8 \\
    
    olmOCR v0.1.75 (Anchored)~\cite{poznanski2025olmocr}& 75.5 ± 1.0  & 74.9 & 71.2 & 71.0 & \cellcolor{cyan!15}\underline{42.2} & 94.5 & 78.3 & 73.3 & 98.3  \\
    
    MonkeyOCR-pro-3B~\cite{li2025monkeyocr}& 75.8 ± 1.0  & \cellcolor{cyan!15}\underline{83.8} & 68.8 & 74.6 & 36.1 & 91.2 & 76.6 & 80.1 & 95.3 \\
    
    MinerU2.5~\cite{niu2025mineru2} & 77.5 ± 1.0 & 81.1 & 74.0 & \cellcolor{cyan!15}\underline{85.1} & 33.8 & 96.3 & 65.5 & \cellcolor{red!15}\textbf{89.8} & 94.4 \\
    dots.ocr~\cite{dotsocr} & \cellcolor{cyan!15}\underline{79.1 ± 1.0} & 82.1 & 64.2 & \cellcolor{red!15}\textbf {88.3} & 40.9 & 94.1 & \cellcolor{red!15}\textbf {82.4} & 81.2 & \cellcolor{red!15}\textbf {99.5} \\
    
     \textbf{PaddleOCR-VL}&\cellcolor{red!15}\textbf {80.0 ± 1.0} & \cellcolor{red!15}\textbf { 85.7} & 71.0 & 84.1 & 37.8 & \cellcolor{red!15}\textbf{97.0} & \cellcolor{cyan!15}\underline{79.9} & \cellcolor{cyan!15}\underline{85.7} & 98.5  \\
    \bottomrule
  \end{tabular}
  }
    \caption{Comprehensive evaluation of document parsing on olmOCR-Bench. Results are reported by olmOCR-Bench~\cite{poznanski2025olmocr} unless MinerU2.5 and Ours.}
     \label{tab:ocr_performance}
\end{table*}

\subsection{Element-level Evaluation}
\label{subsec:element_level_evaluation}

This section centers on evaluating the element-level capabilities of PaddleOCR VL 0.9B. We thoroughly assessed four tasks: text, tables, formulas, and charts using both public competition data and in-house data.

\subsubsection{Text Recognition} 

For text recognition, we utilize three benchmarks to validate the effectiveness of models based on the edit distance metric. 

\paragraph{OmniDocBench-OCR-block:} From the 1355 images of OmniDocBench v1.5, we extracted all text-related sub-images based on layout detection labels, removing any with null annotations. This process resulted in a total of 17,148 block-level images. This evaluation set is named OmniDocBench-OCR-block, with the ground truth still sourced from OmniDocBench. This evaluation set can more accurately assess the model’s text recognition performance on without being affected by layout detection. We use the average normalized edit distance for evaluation.
        
In Table \ref{tab:performance_ocr_overall}, we present a comprehensive comparison of performance across various document types using different models. Our model, PaddleOCR-VL, consistently demonstrates superior performance, achieving the lowest error rates in almost all categories. Specifically, PaddleOCR-VL achieves the best results in the PPT2PDF (0.049), Academic Literature (0.021), Book (0.045), Colorful Textbook (0.081), Exam Paper (0.115), Magazine (0.020), Newspaper (0.034), Note (0.081), and Research Report (0.033) categories. These results highlight PaddleOCR-VL’s robust and versatile capability in handling diverse document types, establishing it as the leading method in the OmniDocBench-OCR-block performance evaluation.

\begin{table}[H]
  \centering
  \setlength{\tabcolsep}{4pt}
  \renewcommand{\arraystretch}{1.2}
    \centering
    \fontsize{7}{7}\selectfont
  \begin{tabular}{l|c *{8}{c}} 
    \toprule
       \multirow{2}{*}{\textbf{Methods}} & \multicolumn{9}{c}{\textbf{Edit Distance $\downarrow$}}
      \\\cline{2-10} & \textbf{PPT2PDF} &
    \multicolumn{1}{p{1.4cm}}{\centering\textbf{Academic\newline Literature}} & 
    \textbf{Book} &
    \multicolumn{1}{p{1.4cm}}{\centering\textbf{Colorful\newline Textbook}} & 
    \multicolumn{1}{p{1.4cm}}{\centering\textbf{ Exam\newline Paper}} & 
    \textbf{Magazine} & \textbf{Newspaper} & \textbf{Note} &
    \multicolumn{1}{p{1.4cm}}{\centering\textbf{Research\newline Report}} \\ 
    \midrule
    Qwen2.5-VL-72B~\cite{bai2025qwen2}  & \cellcolor{cyan!15}\underline{0.054} & \cellcolor{cyan!15}\underline{0.023} & \cellcolor{cyan!15}\underline{0.061} & \cellcolor{cyan!15}\underline{0.084} & 0.195 & 0.032 & \cellcolor{cyan!15}\underline{0.056} & \cellcolor{cyan!15}\underline{0.118} & \cellcolor{cyan!15}\underline{0.040} \\
    MonkeyOCR-pro-3B~\cite{li2025monkeyocr} & 0.058 &  \cellcolor{red!15}\textbf {0.021} & 0.064 & 
    0.096 & \cellcolor{cyan!15}\underline{0.116} & \cellcolor{cyan!15}\underline{0.023} & 0.058 & 0.124 & 0.052 \\
    MinerU2.5~\cite{niu2025mineru2}   & 0.195 & 0.089 & 0.111 & 0.234 & 0.194 & 0.147 & \cellcolor{cyan!15}\underline{0.056} & 0.142 & 0.094 \\
    Dolphin~\cite{feng2025dolphin}  & 0.237 & 0.095 & 0.135 & 0.347 & 0.248 & 0.233 & 0.121 & 0.309 & 0.213 \\
    \textbf{PaddleOCR-VL}   &  \cellcolor{red!15}\textbf {0.049} & \cellcolor{red!15}\textbf {0.021} & \cellcolor{red!15}\textbf {0.045} & \cellcolor{red!15}\textbf {0.081} & \cellcolor{red!15}\textbf {0.115} & \cellcolor{red!15}\textbf {0.020} & \cellcolor{red!15}\textbf {0.034} &
     \cellcolor{red!15}\textbf {0.081} &
     \cellcolor{red!15}\textbf{0.033} \\
    \bottomrule
    \end{tabular}
  \caption{Overall Comparison of OmniDocBench-OCR-block Performance.}
  \label{tab:performance_ocr_overall}
\end{table}

\paragraph{In-house-OCR:} This is our self-built line-level text evaluation dataset which contains 107452 samples with high-quality labels. The dataset includes various text types such as handwritten Chinese, handwritten English, printed Chinese, printed English, traditional Chinese, ancient texts, general scenarios, Pinyin, obscure characters, vertical text, single characters, emojis, and artistic fonts. It also comprises evaluation sets for 109 languages, such as Latin and Japanese. 

Table \ref{tab:performance_comparison} provides a detailed evaluation of performance across multiple languages and text types. In the Multilingual Metrics (Table \ref{tab:performance_multilingual}), the model demonstrates outstanding accuracy with the lowest edit distances in all evaluated scripts: Arabic(0.122), Korean(0.052), Tamil(0.043), Greek(0.135), Thai(0.081), Telugu (0.114), Devanagari (0.097), Cyrillic (0.109), Latin (0.013), and Japanese (0.096), indicating superior capability in handling diverse languages. Similarly, in the Text Type Metrics (Table \ref{tab:performance_text_type}), it excels in various text types, achieving the lowest error rates in categories like Handwritten CN (0.089), Handwritten EN (0.042), Printed CN (0.035), Printed EN (0.016), Traditional Chinese (0.048), Ancient Texts(0.198), General Scene (0.067), Pinyin (0.113), Rare Characters (0.001), Vertical Text (0.005), Single Characters (0.027), Emoji (0.057), and Art Font (0.165). These impressive results underscore the model's robust performance and versatility, establishing it as the leading OCR solution in this benchmark comparison.

\begin{table}[H]
  \centering

  \fontsize{10}{12}\selectfont
   
  \begin{subtable}[t]{1\textwidth}
    \centering
    
    \resizebox{\textwidth}{!}{%
      \renewcommand{\arraystretch}{1.2}
    \setlength{\tabcolsep}{9pt}
    \begin{tabular}{l|cccccccccc}
      \toprule
      \textbf{Methods} & \multicolumn{10}{c}{\textbf{Edit Distance $\downarrow$}}
      \\\cline{2-11}& \textbf{Arabic} & \textbf{Korean} & \textbf{Tamil}& \textbf{Greek} & \textbf{Thai} & \textbf{Telugu} & \textbf{Devanagari} & \textbf{Cyrillic} & \textbf{Latin} & \textbf{Japanese} \\
      \midrule
      Qwen2.5-VL-72B~\cite{bai2025qwen2} & \cellcolor{cyan!15}\underline{0.405}&\cellcolor{cyan!15}\underline{0.056}& \cellcolor{cyan!15}\underline{0.389}& \cellcolor{cyan!15}\underline{0.165}& \cellcolor{cyan!15}\underline{0.194} & \cellcolor{cyan!15}\underline{0.758} &  \cellcolor{cyan!15}\underline{0.164} &  \cellcolor{cyan!15}\underline{0.220} &  \cellcolor{cyan!15}\underline{0.021} &  \cellcolor{cyan!15}\underline{0.181} \\
      Dolphin~\cite{feng2025dolphin} & 0.682 & 0.699 & 0.912 & 0.691 & 0.709 & 0.832 & 0.818 & 0.549 & 0.037 & 0.309 \\
       MonkeyOCR-pro-3B~\cite{li2025monkeyocr} & 0.601 & 0.182 & 0.921 & 0.449 & 0.876 & 0.909 & 0.896 & 0.387 & 0.036 & 0.262 \\
    MinerU2.5~\cite{niu2025mineru2} & 0.978 & 0.917 & 0.957 & 0.661 & 0.880& 0.937 & 0.915 & 0.832 & 0.063 & 0.588 \\
       \textbf{PaddleOCR-VL} & \cellcolor{red!15}\textbf {0.122} & \cellcolor{red!15}\textbf {0.052} & \cellcolor{red!15}\textbf {0.043} & \cellcolor{red!15}\textbf {0.135} & \cellcolor{red!15}\textbf {0.081} & \cellcolor{red!15}\textbf {0.011} & \cellcolor{red!15}\textbf {0.097} & \cellcolor{red!15}\textbf {0.109} & \cellcolor{red!15}\textbf {0.013} & \cellcolor{red!15}\textbf {0.086} \\
      \bottomrule
    \end{tabular}
    }
 
    \caption{Multilingual Metrics.}
       \label{tab:performance_multilingual}
  \end{subtable}

  \vspace{0.5cm} 

\begin{subtable}[t]{1\textwidth}
  \centering
  \resizebox{\textwidth}{!}{%
  \setlength{\tabcolsep}{3pt}
      \renewcommand{\arraystretch}{1.2}
  \begin{tabular}{l|ccccccccccccccc}

    \toprule
   \multirow{2}{*}{ \textbf{Methods}} & \multicolumn{13}{c}{\textbf{Edit Distance $\downarrow$}}
      \\ \cline{2-14} &
    \multicolumn{1}{p{1.3cm}}{\centering\textbf{Hand-\newline written\newline CN}} &
    \multicolumn{1}{p{1.3cm}}{\centering\textbf{Hand-\newline written\newline EN}} &
    \multicolumn{1}{p{1.3cm}}{\centering\textbf{Printed\newline CN}} &
    \multicolumn{1}{p{1.3cm}}{\centering\textbf{Printed\newline EN}} &
    \multicolumn{1}{p{1.3cm}}{\centering\textbf{Trad.\newline Chinese}} &
    \multicolumn{1}{p{1.3cm}}{\centering\textbf{Ancient\newline Texts}} &
    \multicolumn{1}{p{1.3cm}}{\centering\textbf{General\newline Scene}} &
    \textbf{Pinyin} &
    \multicolumn{1}{p{1.3cm}}{\centering\textbf{Rare\newline Char.}} &
    \multicolumn{1}{p{1.3cm}}{\centering\textbf{Vertical\newline Text}} &
    \multicolumn{1}{p{1.3cm}}{\centering\textbf{Single\newline Char.}} &
    \textbf{Emoji} &
    \multicolumn{1}{p{1.3cm}}{\centering\textbf{Art\newline Font}} \\
    \midrule
    Dolphin~\cite{feng2025dolphin} & 0.236 & 0.145 & 0.074 & 0.025 & \cellcolor{cyan!15}\underline{0.095} & \cellcolor{cyan!15}\underline{0.218} & \cellcolor{cyan!15}\underline{0.113} & 0.183 & 0.092 & 0.190 & 0.202 & 0.225 & 0.230 \\
    MonkeyOCR-pro-3B~\cite{li2025monkeyocr} & 0.253 & 0.071 & 0.048 & 0.023 & 0.295 & 0.529 & 0.144 & \cellcolor{cyan!15}\underline{0.165} & 0.063 & \cellcolor{cyan!15}\underline{0.086} & 0.110 & 0.184 & 0.263 \\
  
    Qwen2.5-VL-72B~\cite{bai2025qwen2} &\cellcolor{cyan!15}\underline{ 0.188} & \cellcolor{cyan!15}\underline{0.047} & \cellcolor{cyan!15}\underline{0.037} & \cellcolor{cyan!15}\underline{0.018} & 0.100 & 0.387 & 0.122 & 0.186 & \cellcolor{cyan!15}\underline{0.034} & 0.090 & \cellcolor{cyan!15}\underline{0.041} & \cellcolor{cyan!15}\underline{0.134} & \cellcolor{cyan!15}\underline{0.220} \\
      MinerU2.5~\cite{niu2025mineru2} & 0.370 & 0.088 & 0.041 & 0.023 & 0.232 & 0.950 & 0.179 & 0.256 & 0.048 & 0.962 & 0.097 & 0.174 & 0.337 \\
    \textbf{ PaddleOCR-VL} & \cellcolor{red!15}\textbf {0.089} & \cellcolor{red!15}\textbf {0.042} & \cellcolor{red!15}\textbf {0.035} & \cellcolor{red!15}\textbf {0.016} & \cellcolor{red!15}\textbf {0.048} & \cellcolor{red!15}\textbf {0.198} & \cellcolor{red!15}\textbf {0.067} & \cellcolor{red!15}\textbf {0.113} & \cellcolor{red!15}\textbf {0.001} & \cellcolor{red!15}\textbf {0.005} & \cellcolor{red!15}\textbf {0.027} & \cellcolor{red!15}\textbf {0.057} & \cellcolor{red!15}\textbf {0.165} \\
    \bottomrule
  \end{tabular}
  }
  \caption{Text Type Metrics.}  
  \label{tab:performance_text_type}
  
\end{subtable}
   \caption{Comparison of In-house-OCR Edit Distance Performance.}
 \label{tab:performance_comparison}

\end{table}

\paragraph{Ocean-OCR-Handwritten:} This is a line and paragraph levels handwritten evaluation dataset designed for comprehensive handwriting recognition assessment. It contains 400 samples, evenly divided into four subsets of 100 images each. The dataset covers both real and synthetic handwriting in Chinese and English. Real samples are collected from established handwriting datasets such as CASIA-HWDB~\cite{liu2011casia}, GNHK~\cite{lee2021gnhk}, and BRUSH~\cite{kotani2020generating}, while synthetic samples are generated to simulate diverse writing styles, character densities, and layouts. The benchmark aims to provide balanced and fine-grained evaluation for handwritten text recognition across different scripts and writing conditions.

Table \ref{table:handwritten_ocr} presents a comparison of OCR performance for handwritten English and Chinese text on the Ocean-OCR-Bench. Our model demonstrates superior performance across all metrics in both languages. For English, it achieves the best edit distance of 0.118 and excels in F1-score, Precision, Recall, BLEU, and METEOR, establishing itself as the leading model. In Chinese, PaddleOCR-VL sets a benchmark with an edit distance of 0.034 and leads in all other metrics, showcasing its outstanding precision and reliability.

\begin{table}[H]
 
    \centering
      \fontsize{7}{7}\selectfont

     \renewcommand{\arraystretch}{1.2}
    \begin{tabular}{l|cc|cc|cc|cc|cc|cc}
        \toprule
        \multirow{2}{*}{\textbf{Methods}}   & \multicolumn{2}{c|}{\textbf{Edit Distance $\downarrow$}}  & \multicolumn{2}{c|}{\textbf{F1-score $\uparrow$}} & \multicolumn{2}{c|}{\textbf{Precision$\uparrow$}} & \multicolumn{2}{c|}{\textbf{Recall$\uparrow$}} & \multicolumn{2}{c|}{\textbf{BLEU$\uparrow$}} & \multicolumn{2}{c}{\textbf{METEOR$\uparrow$}} \\
        \cline{2-3} \cline{4-5} \cline{6-7} \cline{8-9} \cline{10-11} \cline{12-13}  
        & \textbf{EN} & \textbf{ZH} & \textbf{EN} & \textbf{ZH} & \textbf{EN} & \textbf{ZH} & \textbf{EN} & \textbf{ZH} & \textbf{EN} & \textbf{ZH} & \textbf{EN} & \textbf{ZH}\\
        
        \midrule
    
        InternVL2.5-4B~\cite{InternVL}& 0.197 & 0.240 & 0.661 & 0.741 & 0.674 & 0.754 & 0.655 & 0.734 & 0.406 & 0.473 & 0.652 & 0.687 \\

        MiniCPM-V2.6-8B~\cite{yao2024minicpm}& 0.147 & 0.175 & 0.727 & 0.810 & 0.747 & 0.811 & 0.714 & 0.812 & 0.443 & 0.583 & 0.727 & 0.774\\
        Qwen2-VL-7B~\cite{qwen2}& \cellcolor{cyan!15}\underline{ 0.127} & 0.113 &  \cellcolor{cyan!15}\underline{0.760} & 0.881 &  \cellcolor{cyan!15}\underline{0.773} & 0.884 &  0.754 & \cellcolor{cyan!15}\underline{0.884} &  0.490 & 0.666 & 0.756 & 0.859\\
        \midrule
        GOT~\cite{wei2024general}& 0.616 & 0.402 & 0.283 & 0.568 & 0.309 & 0.618 & 0.273 & 0.544 & 0.151 & 0.295 & 0.255 & 0.492 \\
       PaddleOCR~\cite{cui2025paddleocr} & 0.418 & 0.325 & 0.237 & 0.664 & 0.232 & 0.646 & 0.263 & 0.700 & 0.069 & 0.431 & 0.236 & 0.648 \\ 
        TextIn & 0.358 & 0.180 & 0.362 & 0.840 & 0.368 & 0.869 & 0.362 & 0.822 & 0.098 & 0.567 & 0.337 & 0.751 \\
        Ocean-OCR~\cite{chen2025ocean} & 0.145 &  \cellcolor{cyan!15}\underline{0.106} & \cellcolor{red!15}\textbf {0.774} &  \cellcolor{cyan!15}\underline{0.885} & \cellcolor{red!15}\textbf {0.780} &  \cellcolor{cyan!15}\underline{0.912} &\cellcolor{red!15}\textbf { 0.782} &  0.862 &  \cellcolor{cyan!15}\underline{0.532} &  \cellcolor{cyan!15}\underline{0.736} &  \cellcolor{cyan!15}\underline{0.772} &  \cellcolor{cyan!15}\underline{0.885} \\
        \midrule
        MinerU2.5~\cite{niu2025mineru2} & 0.238 & 0.356 & 0.558 & 0.619 & 0.547 & 0.623  & 0.574 &  0.622 & 0.344 & 0.489  & 0.553 & 0.601 \\

         \textbf{PaddleOCR-VL} & \cellcolor{red!15}\textbf {0.118} & \cellcolor{red!15}\textbf {0.034} & 0.750 & \cellcolor{red!15}\textbf {0.957} & 0.748 & \cellcolor{red!15}\textbf {0.959} &  \cellcolor{cyan!15}\underline{0.753} & \cellcolor{red!15}\textbf {0.957} & \cellcolor{red!15}\textbf {0.551} &\cellcolor{red!15}\textbf { 0.856} &\cellcolor{red!15}\textbf { 0.787} & \cellcolor{red!15}\textbf {0.936} \\
        \bottomrule
    \end{tabular}
   \caption{Comparison of performance on English(EN) and Chinese(ZH) OCR for handwritten recognition on Ocean-OCR-Bench. Results are reported by Ocean-OCR~\cite{chen2025ocean} unless MinerU2.5 and Ours.}
   \label{table:handwritten_ocr}
\end{table}

\subsubsection{Table Recognition.} For table recognition, we utilize two benchmarks to validate the effectiveness of PaddleOCR-VL-0.9B based on TEDS~\cite{teds} and Edit Distance.

\paragraph{OmniDocBench-Table-block:} To evaluate the table recognition performance of PaddleOCR-VL, we crop 512 tables from OmniDocBench v1.5 datasets. 

As shown in Table \ref{result_on_omnidoc_table_v1_5}, our PaddleOCR-VL leads in the OmniDocBench-Table-block benchmark, surpassing all competitors. It achieves a top overall TEDS of 0.9195, reflecting high accuracy in capturing table structure and content. Its structural TEDS of 0.9543 highlights its ability to parse complex structures, while the lowest Overall Edit Distance of 0.0561 indicates minimal recognition errors. These results confirm PaddleOCR-VL's superior performance and establish it as the benchmark for accurate table recognition.

\begin{table}[H] 
    \centering
\fontsize{7}{7}\selectfont

    \setlength{\tabcolsep}{6pt} 
      \renewcommand{\arraystretch}{1.2}
    \begin{tabular}{l|ccc}
\toprule[.9pt]
\textbf{Methods} & \textbf{Overall TEDS$\uparrow$} & \textbf{Structural TEDS$\uparrow$} & \textbf{Overall Edit Dist$\downarrow$} \\
\midrule
MinerU2-VLM~\cite{MinerU2} &0.9002 &0.9369 &0.0734 \\

Seed1.6 &\cellcolor{cyan!15}\underline{0.9079}  &0.9489 &\cellcolor{cyan!15}\underline{0.0652}  \\

dots.ocr~\cite{dotsocr} &0.8194 &0.8442 &0.1508 \\

MinerU2.5~\cite{niu2025mineru2} &0.9005 &\cellcolor{cyan!15}\underline{0.9539} &0.0693 \\

\textbf{PaddleOCR-VL} &\cellcolor{red!15}\textbf{0.9195} &\cellcolor{red!15}\textbf{0.9543} &\cellcolor{red!15}\textbf{0.0561} \\

\bottomrule
\end{tabular}
\caption{Comparison of OmniDocBench-Table-block Performance}
\label{result_on_omnidoc_table_v1_5}
\end{table}

\paragraph{In-house-Table:} Our self-built evaluation set contains diverse array of table images with comprehensive annotations and type classifications. It includes 20 different table types such as Chinese, English, mixed Chinese-English, and tables with various characteristics like full, partial, or no borders. The collection also covers tables with formulas, dense data, book/manual formats, lists, academic papers, merged cells, as well as low-quality, watermarked, registration forms, statistical forms, research reports, financial reports, images, invoices, and handwritten tables, among others.

Table \ref{tab:performance_0801_refined} provides a comparison of different methods on the In-house-Table task, highlighting their performance across various metrics. We achieves the highest scores in Overall TEDS (0.8699), Structural TEDS (0.9066), Overall Edit Distance (0.9066) and Structural Edit Distance (0.9339). These results underscore PaddleOCR-VL's effectiveness and reliability in table recognition tasks.

\begin{table}[H]
    \centering
\fontsize{7}{7}\selectfont
   
      \renewcommand{\arraystretch}{1.2}
    \begin{tabular}{@{}l| l *{3}{c} @{}} 
        \toprule
        \textbf{Methods} & \textbf{Overall TEDS$\uparrow$} & \textbf{Structural TEDS$\uparrow$} & \textbf{Overall Edit Dist$\uparrow$} & \textbf{Structural Edit Dist$\uparrow$}\\
        \midrule
        MinerU2-VLM~\cite{MinerU2} & 0.8286 & 0.8730 & 0.8757 & 0.9088 \\
        MonkeyOCR~\cite{li2025monkeyocr} & 0.7396 & 0.7824 & 0.8174 & 0.8537 \\
        Nanonets-OCR-s~\cite{Nanonets-OCR-S} & 0.7824 & 0.8190 & 0.8377 & 0.8692 \\
        OCRFlux-3B~\cite{OCRFlux2025} & 0.7741 & 0.8071 & 0.8238 & 0.8617 \\
        Qwen2.5-VL-3B~\cite{bai2025qwen2} & 0.7398 & 0.7765 & 0.8132 & 0.8701 \\
        Qwen2.5-VL-7B~\cite{bai2025qwen2} & 0.7549 & 0.7926 & 0.8251 & 0.8819 \\
        Qwen2.5-VL-72B~\cite{bai2025qwen2} & 0.7762 & 0.8361 & 0.843 & 0.8987 \\
        dots.ocr~\cite{dotsocr} & 0.7547 & 0.7914 & 0.8047 & 0.8361 \\
        MinerU2.5~\cite{niu2025mineru2} & \cellcolor{cyan!15}\underline{0.8469} & \cellcolor{cyan!15}\underline{0.8955} & \cellcolor{cyan!15}\underline{0.8896} & \cellcolor{cyan!15}\underline{0.9239} \\
        \textbf{PaddleOCR-VL} & \cellcolor{red!15}\textbf {0.8699} & \cellcolor{red!15}\textbf {0.9066} & \cellcolor{red!15}\textbf {0.9066} & \cellcolor{red!15}\textbf{0.9339} \\

        \bottomrule
    \end{tabular}
        \caption{Comparison of In-house-Table Performance}
         \label{tab:performance_0801_refined}
\end{table}

\subsubsection{Formula Recognition.} For formula recognition, we validate the effectiveness our model based on the Character Detection Matching (CDM)~\cite{cdm} metric on OmniDocBench-Formula-block and In-house-Formula datasets.

\paragraph{OmniDocBench-Formula-block} 

Using the formula bounding boxes from OmniDocBench v1.5, 1050 formula sub-images were cropped. This step was taken to minimize the influence of layout detection on formula recognition. As shown in Table \ref{result_on_omnidoc_formula}, the model achieved state-of-the-art CDM score of 0.9453. 

\begin{table}[H] 
    \centering
\fontsize{7}{7}\selectfont

    \setlength{\tabcolsep}{6pt} 
      \renewcommand{\arraystretch}{1.2}
    \begin{tabular}{l|ccc}
\toprule[.9pt]
\textbf{Methods} & \textbf{Overall CDM $\uparrow$} & \textbf{EN CDM $\uparrow$} & \textbf{ZH CDM $\uparrow$} \\
\midrule
dots.ocr~\cite{dotsocr} &0.4641 &0.4868 &0.4414 \\

MinerU2-VLM~\cite{MinerU2} &0.8286 &0.9616 &0.6956 \\

MonkeyOCR-pro-1.2B~\cite{li2025monkeyocr} &0.8531 &0.9642 &0.7419 \\

MonkeyOCR-3B ~\cite{li2025monkeyocr} &0.8621 &\cellcolor{cyan!15}\underline{0.9718} &0.7524 \\

Qwen2.5-VL-72B~\cite{bai2025qwen2} &0.8747 &0.9574 &0.7920 \\

MinerU2.5~\cite{niu2025mineru2} &\cellcolor{cyan!15}\underline{0.9187} &\cellcolor{red!15}\textbf{0.9751} &\cellcolor{cyan!15}\underline{0.8623} \\

\textbf{PaddleOCR-VL} &\cellcolor{red!15}\textbf{0.9453} &0.9677 &\cellcolor{red!15}\textbf{0.9228} \\

\bottomrule
\end{tabular}
\caption{Comparison of OmniDocBench v1.5 Formula-block Performance. Due to dots.ocr \cite{dotsocr} easily recognizing cropped formulas as images, the score is relatively low.}
\label{result_on_omnidoc_formula}
\end{table}

\paragraph{In-house-Formula:} The self-constructed formula evaluation set contains 34,816 samples, covering common formula recognition scenarios such as academic papers, mathematics books, and primary and secondary school exam papers. Among them, there are 498 Chinese formulas and 34,318 English formulas. As shown in Table \ref{result_on_In-house_formula},  our model obtains the best performance of 0.9882 CDM score on the In-house-Formula dataset. These results collectively demonstrate the powerful recognition capability of PaddleOCR-VL in real-world complex formula scenarios.

\begin{table}[H] 
    \centering
\fontsize{7}{7}\selectfont

    \setlength{\tabcolsep}{6pt} 
      \renewcommand{\arraystretch}{1.2}
    \begin{tabular}{l|ccc}
\toprule[.9pt]
\textbf{Methods} & \textbf{Overall CDM $\uparrow$} & \textbf{EN CDM $\uparrow$} & \textbf{ZH CDM $\uparrow$} \\
\midrule
dots.ocr~\cite{dotsocr} &0.6737 &0.8066 &0.5408 \\

MinerU2-VLM~\cite{MinerU2} &0.9237 &0.9764 &0.8709 \\

MonkeyOCR-pro-1.2B~\cite{li2025monkeyocr} &0.9537 &0.9656 &0.9417 \\

MonkeyOCR-3B ~\cite{li2025monkeyocr} &0.9566 &0.9761 &0.9371 \\

Qwen2.5-VL-72B~\cite{bai2025qwen2} &0.9412 &0.9519 &0.9304 \\

MinerU2.5~\cite{niu2025mineru2} &\cellcolor{cyan!15}\underline{0.9770} &\cellcolor{cyan!15}\underline{0.9832} &\cellcolor{cyan!15}\underline{0.9708} \\

\textbf{PaddleOCR-VL} &\cellcolor{red!15}\textbf{0.9882} &\cellcolor{red!15}\textbf{0.9914}&\cellcolor{red!15}\textbf{0.9849} \\

\bottomrule
\end{tabular}
\caption{Comparison of In-house-Formula Performance. Due to dots.ocr \cite{dotsocr} easily recognizing cropped formulas as images, the score is relatively low.}
\label{result_on_In-house_formula}
\end{table}

\subsubsection{Chart Recognition.} For chart recognition, considering the limitations in dataset size, the imbalanced distribution of chart categories, and the poor annotation quality of publicly available test sets, we only utilize a in-house benchmark to validate the effectiveness of PaddleOCR-VL-0.9B based on the RMS-F1~\cite{deplot} score metric. As shown in Table \ref{tab:rms_f1_performance}, the proposed PaddleOCR-VL not only outperforms expert OCR VLMs but also surpasses some 72B-level multimodal language models.

\paragraph{In-house-Chart:} This in-house chart recognition evaluation set comprises 1,801 samples, all of which have underwent a rigorous manual review to ensure annotation correctness. The evaluation set is broadly categorized into 11 chart categories, including bar-line hybrid, pie, 100\% stacked bar, area, bar, bubble, histogram, line, scatterplot, stacked area, and stacked bar. Of these samples, 851 are in English and 950 are in Chinese. Prior to evaluation, both the predicted data table and the ground truth data table are normalized to a uniform markdown format to eliminate expression ambiguities.

\begin{table}[H]
\centering
\fontsize{7}{7}\selectfont

\setlength{\tabcolsep}{8pt}
 \renewcommand{\arraystretch}{1.2}
\begin{tabular}{@{} l| c c c @{}}
\toprule
\multirow{2}{*}{\textbf{Methods}} & \multicolumn{3}{c}{\textbf{RMS-F1 $\uparrow$}} \\
\cline{2-4}
& \textbf{Overall} & \textbf{EN} & \textbf{ZH} \\
\midrule
TinyChart~\cite{zhang2024tinychart}      & 0.2159          & 0.4726          & 0.0876 \\

GOT~\cite{wei2024general}              & 0.3160          & 0.1100          & 0.4190 \\
OneChart~\cite{onechart}      & 0.3716          & 0.1384          & 0.4882 \\
Qwen2.5-VL-3B~\cite{bai2025qwen2}   & 0.5942          & 0.5619          & 0.6103 \\
Qwen2.5-VL-7B~\cite{bai2025qwen2}    & 0.6821          & 0.5876          & 0.7293 \\
Qwen2.5-VL-72B~\cite{bai2025qwen2}    & 0.7300          & 0.6972          & 0.7464 \\

PP-StructureV3~\cite{cui2025paddleocr}   & \cellcolor{cyan!15}\underline{0.8060}          & \cellcolor{cyan!15}\underline{0.7963}         & \cellcolor{cyan!15}\underline{0.8109} \\
\textbf{PaddleOCR-VL}     & \cellcolor{red!15}\textbf {0.8440}          & \cellcolor{red!15}\textbf {0.8222}          & \cellcolor{red!15}\textbf {0.8549} \\
\bottomrule
\end{tabular}
\caption{Comparison of In-house-Chart Performance}
\label{tab:rms_f1_performance}
\end{table}

\subsection{Inference Performance}

To improve the inference performance of PaddleOCR-VL, we introduce multi-threading asynchronous execution into the inference workflow. The process is divided into three main stages---data loading (e.g., rendering PDF pages as images), layout model processing, and VLM inference---each running in a separate thread. Data is transferred between adjacent stages via queues, enabling concurrent execution for higher efficiency. In particular, for VLM inference, batch processing is only triggered when either the number of items in the queue reaches a predefined threshold or the waiting time for queued data exceeds a specified limit. This design allows blocks across different pages to be aggregated and processed together, thereby maximizing parallelism, especially when handling large volumes of files. We further deploy PaddleOCR-VL-0.9B on high-throughput inference and serving engines~\cite{kwon2023efficient,zheng2024sglang,PaddlePaddle_FastDeploy}, tuning parameters like max-num-batched-tokens and gpu-memory-utilization to balance inference throughput with GPU memory consumption.


We measured the end-to-end inference speed and GPU usage on the OmniDocBench v1.0 dataset, processing PDF files in batches of 512 on a single NVIDIA A100 GPU. By "end-to-end", we mean that the inference time was measured from providing the PDF file path as input to the complete generation of the Markdown text. For MonkeyOCR, dots.ocr, and MinerU, inference was run with the vLLM backend and the default configuration (including the KV cache settings). The generated Markdown text was tokenized with the "cl100k\_base" tokenizer to compute the number of output tokens. For dots.ocr specifically, 200 threads were used for concurrent page processing, and the Base64-encoded image content in the produced Markdown text was replaced with a dummy path (UUID-based, prefixed with "images/" and suffixed with ".png") to ensure a reasonable token count.

Table~\ref{tab:inference_performance} provides a comprehensive comparison of inference efficiency across different methods. The proposed PaddleOCR-VL demonstrates clear and consistent advantages in processing speed. When deployed with the FastDeploy\footnote{https://github.com/PaddlePaddle/FastDeploy} backend, it achieves 53.1\% higher page throughput and 50.9\% higher token throughput than the leading baseline, MinerU2.5, establishing itself as the most efficient solution overall. These results collectively confirm that PaddleOCR-VL attains state-of-the-art inference efficiency through a balanced optimization of speed, making it highly suitable for real-world, high-throughput document understanding scenarios.

\begin{table}[H]
  \centering
  \fontsize{7}{7}\selectfont
  \renewcommand{\arraystretch}{1.2}
  \begin{tabular}{l|cccccc}
    \toprule
    \textbf{Methods} & \textbf{Backend} & \textbf{Total Time (s)$\downarrow$} & \textbf{Pages/s$\uparrow$} & \textbf{Tokens/s$\uparrow$} 
    \\
    \midrule
    MonkeyOCR-pro-1.2B~\cite{li2025monkeyocr}& vLLM (v0.10.2) & 1456.4 & 0.6730 & 1120.3 & 
    \\
    dots.ocr~\cite{dotsocr} & vLLM (v0.10.2)& 2784.6 & 0.3522 & 532.9 & 
    \\
    MinerU2.5~\cite{niu2025mineru2} & vLLM (v0.10.2)& 927.3 & 1.0574 & 1647.9 & 
    \\
    PaddleOCR-VL& SGLang (v0.5.2)& 882.1 & 1.1115 & 1707.8 & 
    \\
    PaddleOCR-VL& vLLM (v0.10.2) & \cellcolor{cyan!15}\underline{728.7} & \cellcolor{cyan!15}\underline{1.3453} & \cellcolor{cyan!15}\underline{2067.6} & 
    \\
    PaddleOCR-VL & FastDeploy (v2.3) & \cellcolor{red!15}\textbf{605.6} & \cellcolor{red!15}\textbf{1.6184} & \cellcolor{red!15}\textbf{2486.4} & 
    \\
    \bottomrule
  \end{tabular}
  \caption{
    End-to-End Inference Performance Comparison. 
  }
  \label{tab:inference_performance}
\end{table}

\section{Conclusion}

This report introduces PaddleOCR-VL, an advanced and efficient model for document parsing that excels at both element-level and page-level recognition. Its core componets, PaddleOCR-VL-0.9B, built with a NaViT-style visual encoder and ERNIE-4.5-0.3B language model, it accurately recognizes complex elements such as text, tables, formulas, and charts in over 100 languages. PaddleOCR-VL achieves fast inference and low resource consumption, making it practical for real-world deployment. It outperforms existing pipeline solutions on many benchmarks and effectively handles challenging content including handwriting and historical documents, as well as converting chart visuals into structured data. Its broad multilingual support and strong performance have the potential to advance the application and development of multimodal document processing technologies, bringing innovation to automated analysis and information retrieval. This will significantly enhance the performance and stability of RAG systems, making information extraction from complex documents more efficient, thereby providing more reliable data support for future AI applications.
\bibliography{main}

\setcounter{figure}{0}
\makeatletter 
\renewcommand{\thefigure}{A\@arabic\c@figure}
\makeatother

\setcounter{table}{0}
\makeatletter 
\renewcommand{\thetable}{A\@arabic\c@table}
\makeatother

\clearpage 
\newpage
\appendix

\section*{Appendix}
\label{sec:appendix}

\section{Training Dataset Details}

This two-stage approach offers unique advantages in terms of data collection, as obtaining isolated element imagesalong with their annotations is more feasible than collecting complete document pages containing different elements. In the following sections, we will elaborate on the construction of multimodal model training data for text, tables, formulas, and charts.

\subsection{Text}

We have curated a large-scale dataset comprising 20 Million High-Quality Image-Text Pairs. As shown in Figure~\ref{fig:text_dataset}, the dataset generation follows a rigorous multi-stage pipeline which primarily involves:

\begin{figure}[h]
\centering
\includegraphics[width=\linewidth]{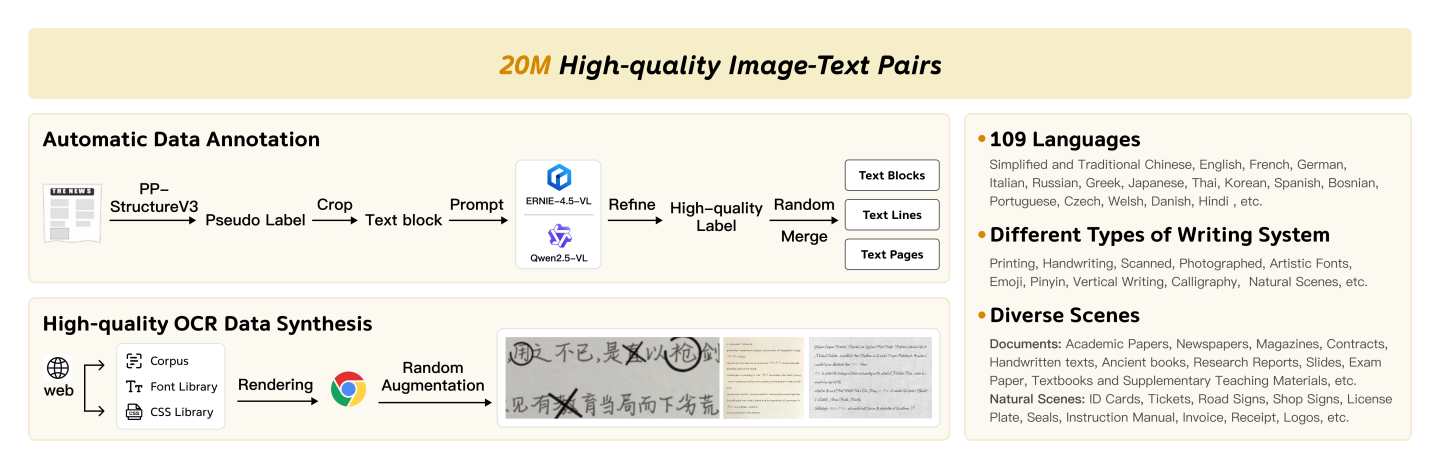} 

\caption{
    \centering
    The construction method and characteristics of the text training data for PaddleOCR-VL-0.9B. 
}
\label{fig:text_dataset}
\end{figure}

\begin{enumerate}[label=\arabic*.]

\item \textbf{Automatic Data Annotation:} We design an automatic annotation pipeline that integrates lightweight document-structure models with large multimodal language models. Specifically, PP-StructureV3 is employed as an expert model to perform layout analysis and text recognition, generating pseudo labels that are converted into prompts for multimodal models such as ERNIE-4.5-VL and Qwen2.5-VL to refine. Finally, the refined labels are aggregated and randomly merged at multiple granularities to produce 20 million high-quality image–text training samples.

\item \textbf{High-quality OCR Data Synthesis:} During data distillation, low label quality in challenging scenarios like messy handwriting and dense blurry text was addressed by expanding the dataset through synthetic generation. Utilizing diverse CSS styles, over 200 fonts, and various corpora, we rendered a large amount of images, thereby enhancing the model’s capabilities in these difficult scenarios.

\end{enumerate}

Ultimately, the data is meticulously annotated at three distinct hierarchical levels: text lines, text blocks, and text pages. With extensive language coverage of 109 languages, including major global ones like Chinese, English, French, and Hindi. It includes diverse scenes including Academic Papers, Newspapers, Handwritten texts, Ancient books, Id cards, tickets, seals, etc. Additionally, the dataset addresses compatibility with a variety of writing systems and text styles, covering Printing, Handwriting, Scanned text, Artistic Fonts, etc.

\subsection{Table}

As shown in Figure \ref{fig:table_dataset}, we constructed a large-scale dataset of over $5$ million high-quality image-table pairs. Our dataset construction employs three key strategies: automatic data annotation, potential annotation mining, and high-quality data synthesis. For coding efficiency, we adopt OTSL~\cite{lysak2023optimized} as the model’s target format instead of conventional HTML. The main dataset construction process is as follows:
\begin{figure}[h]
\centering
\includegraphics[width=\linewidth]{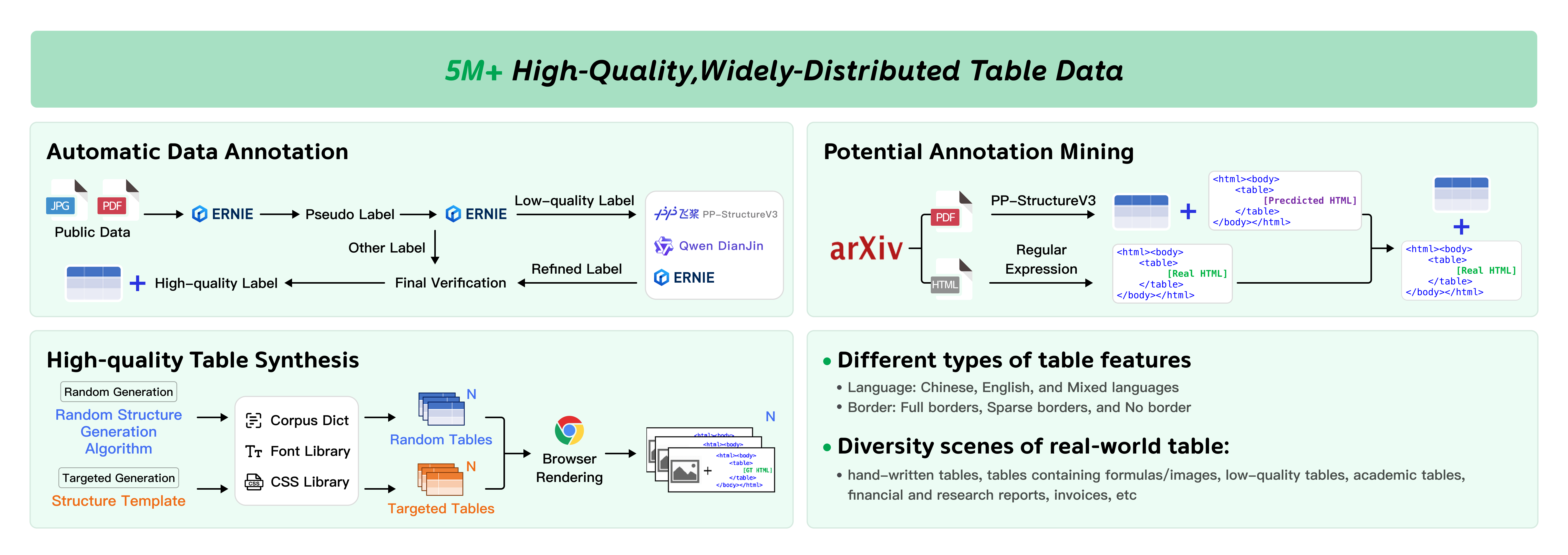} 

\caption{
    \centering
    The construction method and characteristics of the table training data for PaddleOCR-VL-0.9B.
}
\label{fig:table_dataset}
\end{figure}
\begin{enumerate}
\item \textbf{Automatic Data Annotation:} To enhance the performance of PaddleOCR-VL in table recognition, we built a large-scale, diverse dataset covering various languages, border styles, and table types. Tables are first located using PP-StructureV3~\cite{cui2025paddleocr}. For unlabeled images, we employed a multi-stage annotation pipeline: ERNIE-4.5-VL~\cite{ernie2025technicalreport} first generates pseudo-labels, which are then validated by a ERNIE-4.5-VL-28B-A3B~\cite{ernie2025technicalreport} as discriminative model. Rejected annotations are refined using DianJin-OCR-R1~\cite{chen2025dianjin} (for tools, we use ERNIE-4.5-VL and PP-StructureV3~\cite{cui2025paddleocr}). Finally, all annotations undergo rigorous rule-based verification, including n-gram analysis and HTML validation, to ensure only high-quality samples are used for training.
\item \textbf{Potential Annotation Mining:}

For public data with potential annotations (e.g., from arXiv), we extract tables and their corresponding official-supported HTML source code. We then employ a mechanism combining regular expression matching with contextual and sequential alignment to construct accurate table-HTML pairs. The extracted HTML subsequently undergoes rule-based filtering, yielding high-quality data samples ready for model training.
\item \textbf{High-quality Table Synthesis:} 

To overcome data imbalance and high annotation costs, we introduce an innovative high-quality table synthesis tool which constitutes the cornerstone of our table data collection pipeline. This tool enables both randomized synthesis for comprehensive data supplement and targeted synthesis to enhance recognition of specific table categories. Specifically, we first leverage LLMs to gather a diverse and extensive corpus.Then, our tool generates table training pairs through randomized configurations of structures, fonts, CSS styles, and textual content, while also supporting customized synthesis by specifying particular parameters to accurately simulate specialized table types. With a synthesis speed of $10,000$ samples per hour, our tool has produced over $5,500,000$ training instances, substantially enhancing our model's generalization capability and comprehensive performance in table recognition.
\end{enumerate}

Through the aforementioned data construction strategies, we build a comprehensive table dataset encompassing diverse table categories and recognition scenarios, thereby providing robust support for training our model in the table recognition task.

\subsection{Formula}

\par As shown in Figure \ref{fig:formula_dataset}, this dataset was developed using a range of strategies, including source code rendering, automatic data annotation, targeted synthesis of long-tail data, and public data collection. It encompasses a variety of formula scenarios, such as educational supplementary materials, test papers for primary and secondary schools, mathematical papers, PowerPoint courseware, university theses, financial research reports, and handwritten mathematical notes. The dataset features four types of formulas: Simple Printed Expressions, Complex Printed Expressions, Screen-Captured Expressions, and Handwritten Expressions, available in both Chinese and English. The main process for constructing the dataset is as follows:
\begin{figure}[h]
\centering

\includegraphics[width=\linewidth]{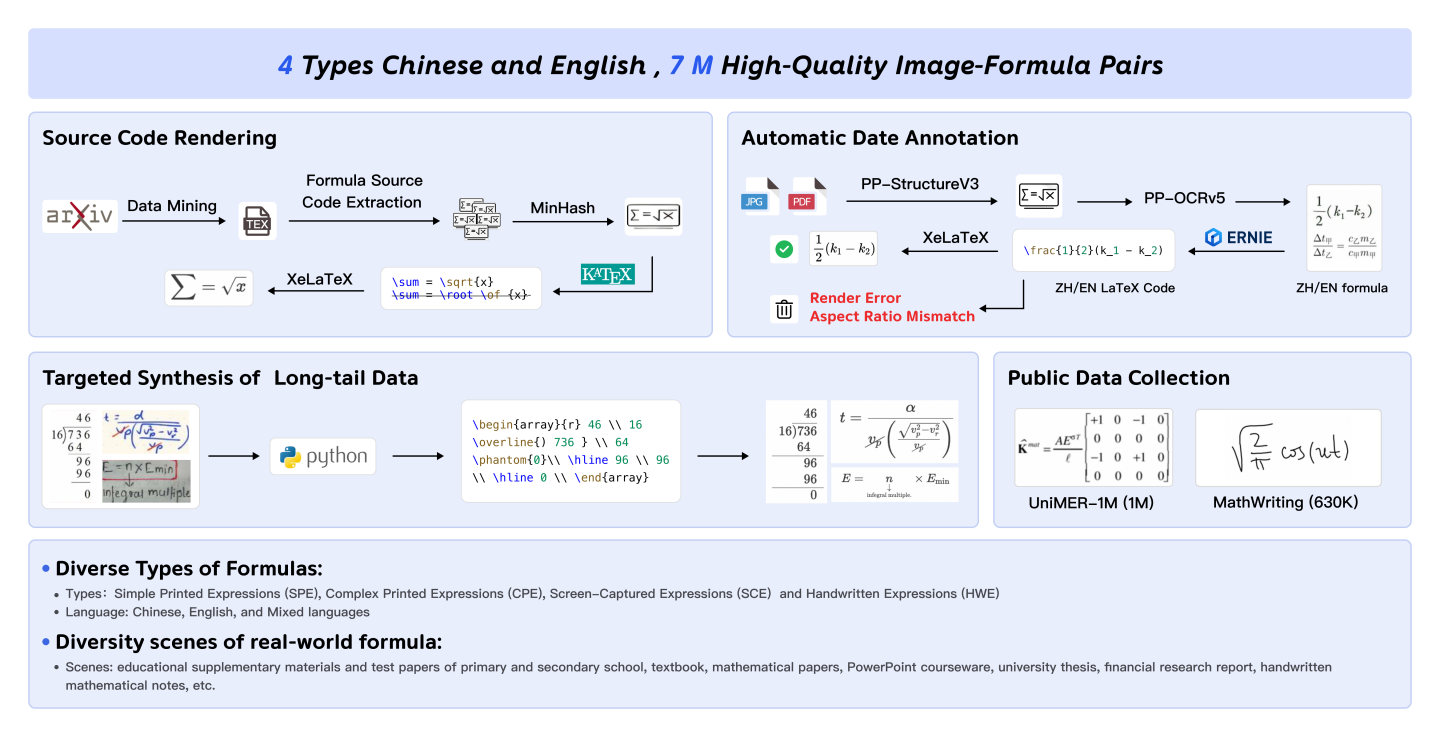} 

\caption{
    \centering
    The construction method and characteristics of the formula training data for PaddleOCR-VL-0.9B.
}
\label{fig:formula_dataset}
\end{figure}
\begin{enumerate}[label=\arabic*.]
\item \textbf{Source Code Rendering:} To enhance the model's adaptability to a wide variety of unusual formula structures, a large amount of paper source code was scraped from arXiv, and LaTeX code for the formulas was extracted using regular expressions. Then, MinHash was used to remove duplicate and highly similar formula source codes, and KaTeX was employed to normalize the formula source codes, thereby reducing their ambiguity. Finally, the formulas were re-rendered into images using a formula rendering engine.

\item \textbf{Automatic Data Annotation:} For real-world formula data from exam papers, educational materials, and handwritten notes, the process begins with the use of the layout analysis method PP-StructureV3 \cite{cui2025paddleocr} to identify the bounding boxes for formulas. Based on these bounding boxes, formula regions are cropped from the images. Subsequently, large multimodal language models, such as ERNIE-4.5-VL-28B-A3B~\cite{ernie2025technicalreport}, are employed to generate the LaTeX source code for these formulas. Given the rarity of Chinese formulas in real-world scenarios—where approximately 1 out of 100 formulas contains Chinese characters—PP-OCRv5~\cite{cui2025paddleocr} is utilized to recognize characters within the cropped regions, enabling targeted optimization when Chinese characters are detected. Due to the complex and diverse nature of real-world formulas, recognition errors may occur with existing large models. To address this, a LaTeX rendering engine is used to filter the formulas generated by these models. Specifically, image-formula pairs that cannot be successfully rendered by xelatex are discarded. For those that render successfully, a more in-depth screening is conducted by comparing metrics such as the aspect ratio between the recognized image and the rendered image.

\item \textbf{Targeted Synthesis of Long-tail Data:}  For certain long-tail formula structures, such as elementary school vertical calculations, formulas with strikethroughs, and handwritten formulas with explanatory arrows, existing multimodal large models struggle to accurately recognize them due to data distribution issues. To address this, LaTeX code is synthetically generated based on rules and inverse rendering is performed using a LaTeX rendering engine, thereby constructing image-formula matching pairs for these long-tail scenarios.

\item \textbf{Public Data Collection:} In order to enable the model to learn high-quality formula representations, a substantial amount of data has been collected from existing public datasets, including  UniMER-1M~\cite{unimernet} and MathWriting~\cite{MathWriting}. Specifically,  UniMER-1M is oriented towards real document scenarios and has gathered 1 million formula data from arXiv, Pix2tex~\cite{pix2tex}, CROHME~\cite{chrome2014,mouchere2016icfhr2016,chrome2019}, and HME100K~\cite{yuan2022syntax}. On the other hand, MathWriting is currently the largest handwritten mathematical formula dataset, comprising 230,000 real handwritten formula samples and 400,000 synthetic handwritten formula samples.

\end{enumerate}

\subsection{Chart}

\par We constructed a large-scale, bilingual (Chinese and English) dataset of over 0.8 million high-quality image-chart pairs. Our dataset construction employs four key strategies: public data collection and cleaning, automatic data annotation, data synthesis, and targeted long-tail data augmentation. The dataset covers a wide array of chart types from diverse sources, including academic papers, financial reports, and web pages. The main dataset construction process is as follows:

\begin{figure}[H]
\centering
\includegraphics[width=\linewidth]{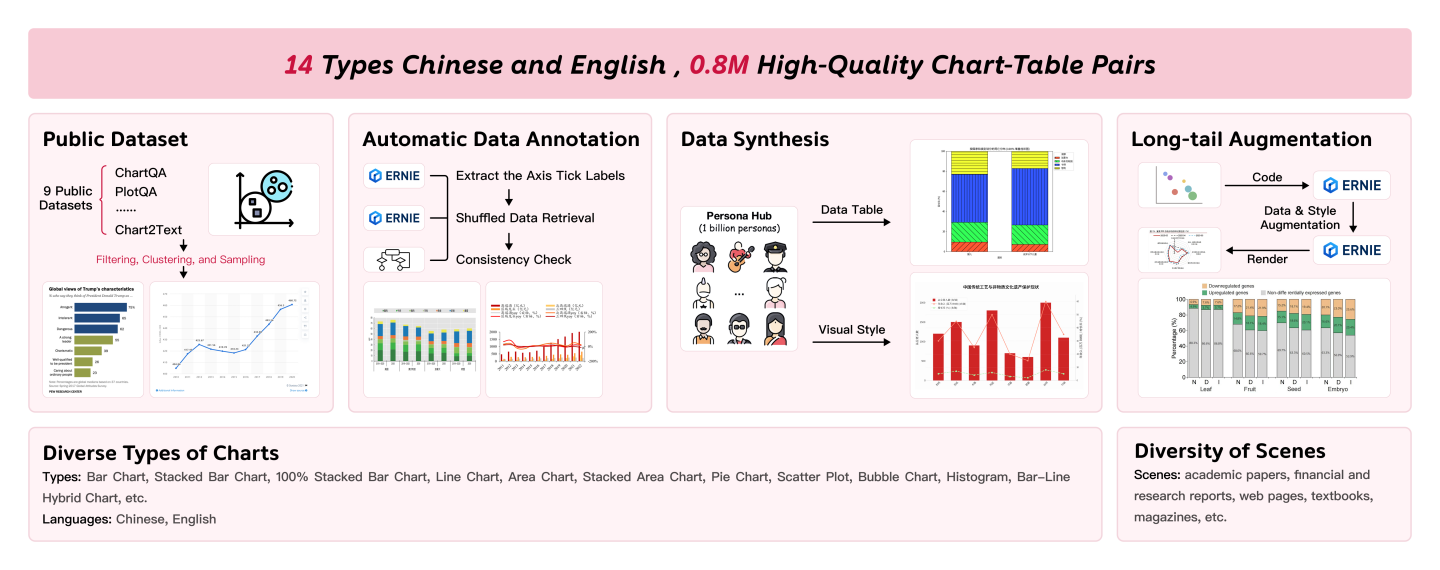} 

\caption{
    \centering
    The construction method and characteristics of the chart training data for PaddleOCR-VL-0.9B.
}
\label{fig:chart_dataset}
\end{figure}

\begin{enumerate}[label=\arabic*.]

\item \textbf{Public Data Collection and Cleaning:} We collected a large number of samples from public datasets, including ChartQA~\cite{chartqa}, PlotQA~\cite{plotqa}, Chart2Text~\cite{chart2text}, DVQA~\cite{dvqa}, Unichart~\cite{masry2023unichart}, Beagle~\cite{beagle}, ChartINFO~\cite{chartinfo}, visText~\cite{tang2023vistext}, and ExcelChart~\cite{excelchart}. However, the raw datasets suffered from poor annotation quality and extremely imbalanced data distributions. Inspired by~\cite{llavachart}, a meticulous data cleaning and filtering pipeline was implemented to remove noisy samples and ensure balanced clustering, resulting in a high-quality dataset of 220k samples.

\item \textbf{Automatic Data Annotation:} To annotate our large collection of unlabeled public and in-house data, we developed a two-stage annotation pipeline based on the Vision Large Language Model ERNIE-4.5-VL~\cite{ernie2025technicalreport}. In the first stage, the model extracts tick labels from the x- and y-axes; in the second, random permutations of these labels are used to query corresponding data points, framing annotation as a data retrieval task. A final consistency check ensures that only verified annotations are included in the training set, guaranteeing high reliability.

\item \textbf{Data Synthesis:} To capture diverse visual styles and enhance model generalization, we designed a three-stage data synthesis pipeline. It begins with a large collection of base data tables, followed by an LLM Persona~\cite{ge2024scaling} strategy using ERNIE-X1~\cite{ernie2025technicalreport}, which diversifies table content and generates persona-specific rendering code. This enables control over chart aesthetics such as color, font, and layout. Leveraging a billion distinct personas, the pipeline produces highly varied data structures and visual styles, substantially improving PaddleOCR-VL’s generalization across real-world charts. For rendering, we employ matplotlib and seaborn.

\item \textbf{Targeted Long-tail Data Augmentation:} To improve generalization on real-world long-tail samples, we designed a data augmentation pipeline based on seed charts. It first selects long-tail samples by their distinctive visual features, then uses ERNIE-4.5-VL~\cite{ernie2025technicalreport} to replicate their rendering code. ERNIE-X1~\cite{ernie2025technicalreport}, guided by a specific persona~\cite{ge2024scaling}, further diversifies the code by altering data tables and visual styles. Executing the modified code produces new augmented charts with corresponding data tables.

\end{enumerate}

\par Through the four data construction strategies mentioned above, the final chart dataset covers a wide range of application scenarios and a rich variety of chart styles, providing strong support for the training of chart models.

\clearpage 
\newpage

\section{Supported Languages}

PaddleOCR-VL supports a total of 109 languages. Table \ref{tab:performance_comparison} in the main text shows the text line recognition accuracy for different languages. Table~\ref{tab:supported_language} lists the correspondence between each language category and the specific supported languages.

\begin{table}[ht]
\centering
\begin{tabular}{>{\centering\arraybackslash}m{0.22\textwidth}|>{\centering\arraybackslash}m{0.75\textwidth}}
\toprule
\textbf{Language Category} & \textbf{Specific Languages} \\
\midrule
Chinese & Chinese \\
\midrule
English & English \\
\midrule
Korean & Korean \\
\midrule
Japanese & Japanese \\
\midrule
Thai & Thai\\
\midrule
Greek & Greek \\
\midrule
Tamil & Tamil \\
\midrule
Telugu & Telugu \\
\midrule
Arabic & Arabic, Persian, Uyghur, Urdu, Pashto, Kurdish, Sindhi, Balochi \\
\midrule
Latin & French, German, Afrikaans, Italian, Spanish, Bosnian, Portuguese, Czech, Welsh, Danish, Estonian, Irish, Croatian, Uzbek, Hungarian, Serbian (Latin), Indonesian, Occitan, Icelandic, Lithuanian, Maori, Malay, Dutch, Norwegian, Polish, Slovak, Slovenian, Albanian, Swedish, Swahili, Tagalog, Turkish, Latin, Azerbaijani, Kurdish, Latvian, Maltese, Pali, Romanian, Vietnamese, Finnish, Basque, Galician, Luxembourgish, Romansh, Catalan, Quechua \\
\midrule
Cyrillic & Russian, Belarusian, Ukrainian, Serbian (Cyrillic), Bulgarian, Mongolian, Abkhazian, Adyghe, Kabardian, Avar, Dargin, Ingush, Chechen, Lak, Lezgin, Tabasaran, Kazakh, Kyrgyz, Tajik, Macedonian, Tatar, Chuvash, Bashkir, Malian, Moldovan, Udmurt, Komi, Ossetian, Buryat, Kalmyk, Tuvan, Sakha, Karakalpak \\
\midrule
Devanagari & Hindi, Marathi, Nepali, Bihari, Maithili, Angika, Bhojpuri, Magahi, Santali, Newari, Konkani, Sanskrit, Haryanvi \\
\bottomrule
\end{tabular}
   \caption{Supported Languages}
       \label{tab:supported_language}
\end{table}

\clearpage 
\newpage

\section{Inference Performance on Different Hardware Configurations}

We measured the inference performance of PaddleOCR-VL on different hardware configurations, as summarized in Table~\ref{tab:inference_performance_hardware}.We used FastDeploy version 2.3.0, vLLM version 0.10.2, and SGLang version 0.5.2. As observed, PaddleOCR-VL demonstrates stable and efficient inference performance across a wide range of hardware and backend configurations, showing that the system can flexibly adapt to diverse computing environments.

\begin{table}[H]
  \centering

  \renewcommand{\arraystretch}{1.2}
  \begin{tabular}{l|c|ccccc}
    \toprule
    \textbf{Hardware} & \textbf{Backend} & \textbf{Total Time (s)$\downarrow$} & \textbf{Pages/s$\uparrow$} & \textbf{Tokens/s$\uparrow$} & \textbf{Avg. VRAM Usage (GB)$\downarrow$} \\
    \midrule
    \multirow{3}{*}{H800} & FastDeploy & 400.4 & 2.2250 & 3416.7 & 63.5 \\ & vLLM & 596.7 & 1.6442 & 2524.9 & 44.2 \\
                          & SGLang & 875.2 & 1.1198 & 1720.8 & 49.5 \\
    \midrule
    \multirow{3}{*}{A100} & FastDeploy & 605.6 & 1.6184 & 2486.4 & 62.8 \\ & vLLM & 728.7 & 1.3453 & 2067.6 & 40.1 \\
                          & SGLang & 882.1 & 1.1115 & 1707.8 & 49.7 \\
    \midrule
    \multirow{3}{*}{H20} & FastDeploy & 635.0 & 1.5432 & 2371.2 & 64.3 \\ & vLLM & 694.7 & 1.4112 & 2168.8 & 74.8 \\
                          & SGLang & 759.6 & 1.2906 & 1986.6 & 80.2 \\
    \midrule
    \multirow{3}{*}{L20} & FastDeploy & 802.0 & 1.2224 & 1881.9 & 40.8 \\ & vLLM & 904.9 & 1.0835 & 1665.1 & 25.2 \\
                          & SGLang & 953.6 & 1.0280 & 1579.0 & 31.3 \\
    \midrule
    \multirow{3}{*}{A10} & FastDeploy & 1197.2 & 0.8191 & 1260.9 & 21.8 \\ & vLLM & 1320.5 & 0.7426 & 1141.2 & 15.6 \\
                          & SGLang & 1446.2 & 0.6781 & 1043.8 & 19.9 \\
    \midrule
    \multirow{2}{*}{RTX 3060} & vLLM & 2694.1 & 0.3641 & 559.5 & 11.9 \\
                              & SGLang & 2765.2 & 0.3547 & 546.1 & 11.9 \\
    \midrule
    \multirow{2}{*}{RTX 4090D} & vLLM & 851.9 & 1.1507 & 1768.1 & 16.6 \\
                               & SGLang & 932.4 & 1.0514 & 1618.5 & 20.9 \\
    \bottomrule
  \end{tabular}%
    \caption{End-to-End Inference Performance}
      \label{tab:inference_performance_hardware}
\end{table}

\clearpage 
\newpage
\section{Real-world Samples}

This appendix showcases the parsing and recognition capabilities of our proposed algorithm across a variety of challenging scenarios.

\begin{figure}[H]
\centering
\includegraphics[width=0.95\linewidth]{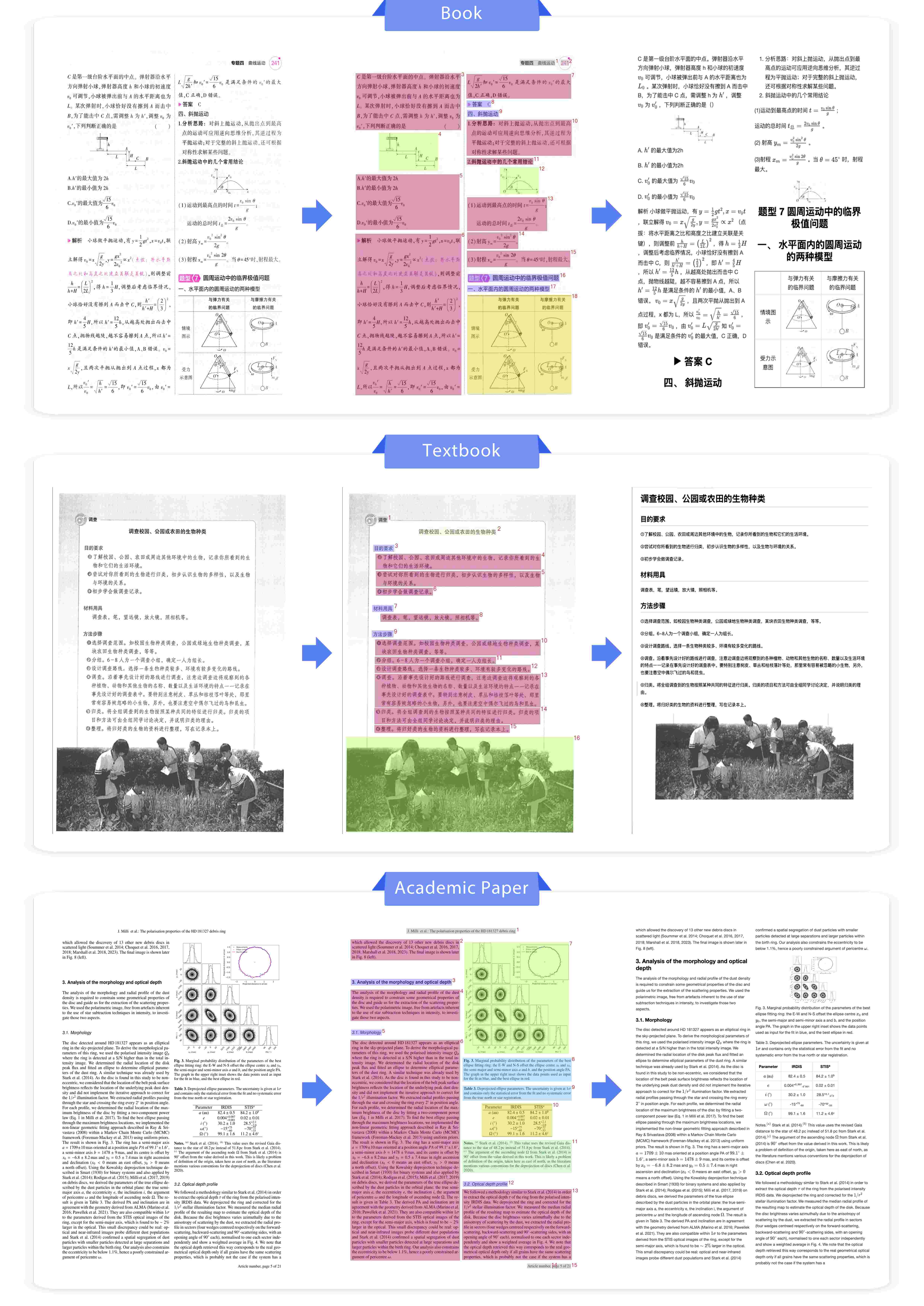} 

\caption{
    \centering
     The Layout and Markdown Output for Book, Textbook and Academic Paper.
}
\label{fig:overview1}
\end{figure}

\begin{figure}[H]
\centering
\includegraphics[width=0.95\linewidth]{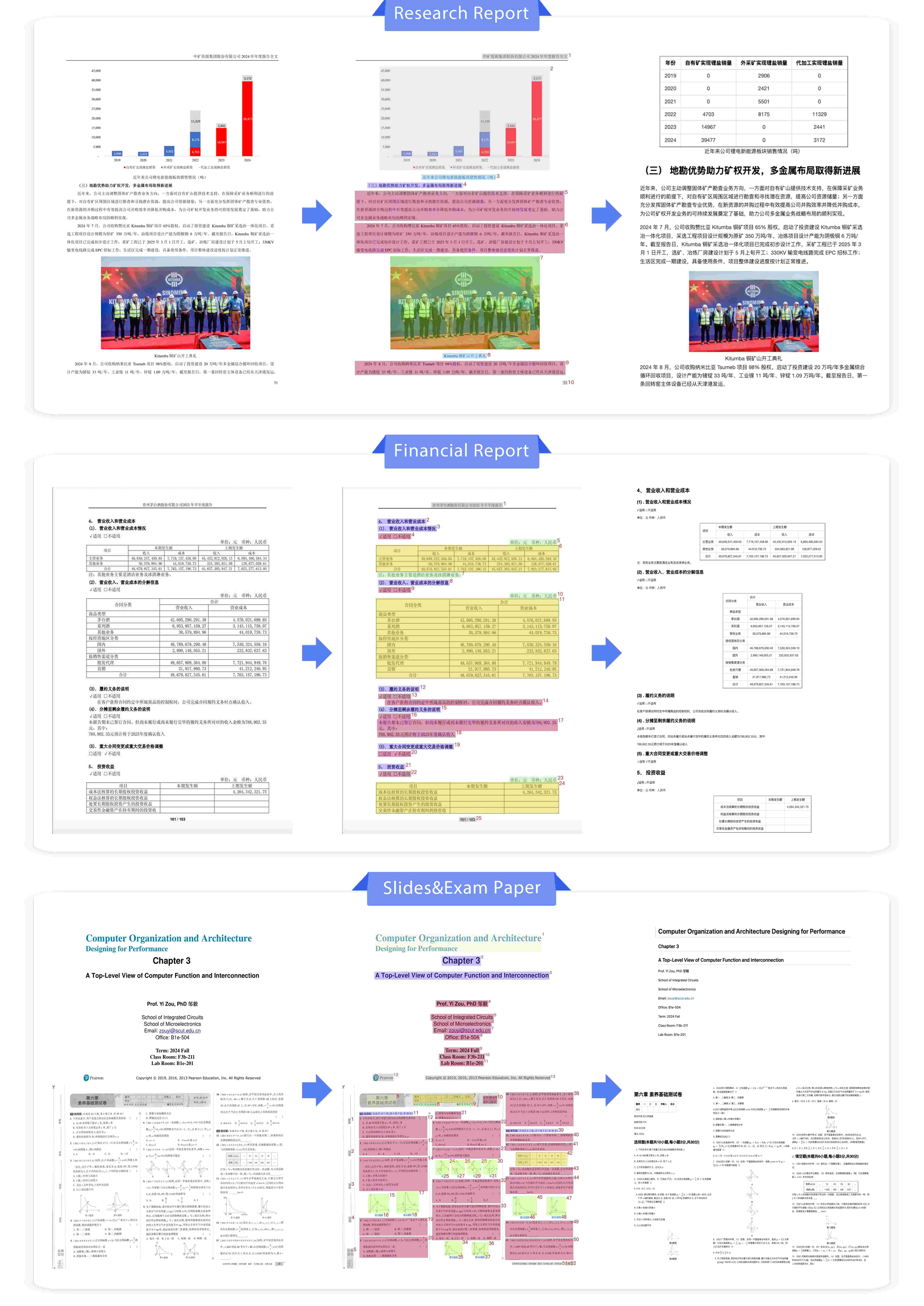} 

\caption{
    \centering
    The Layout and Markdown Output for Research Report(with chart recognition enabled), Financial Report, Slides and Exam Paper.
}
\label{fig:overview2}
\end{figure}

\begin{figure}[H]
\centering
\includegraphics[width=0.95\linewidth]{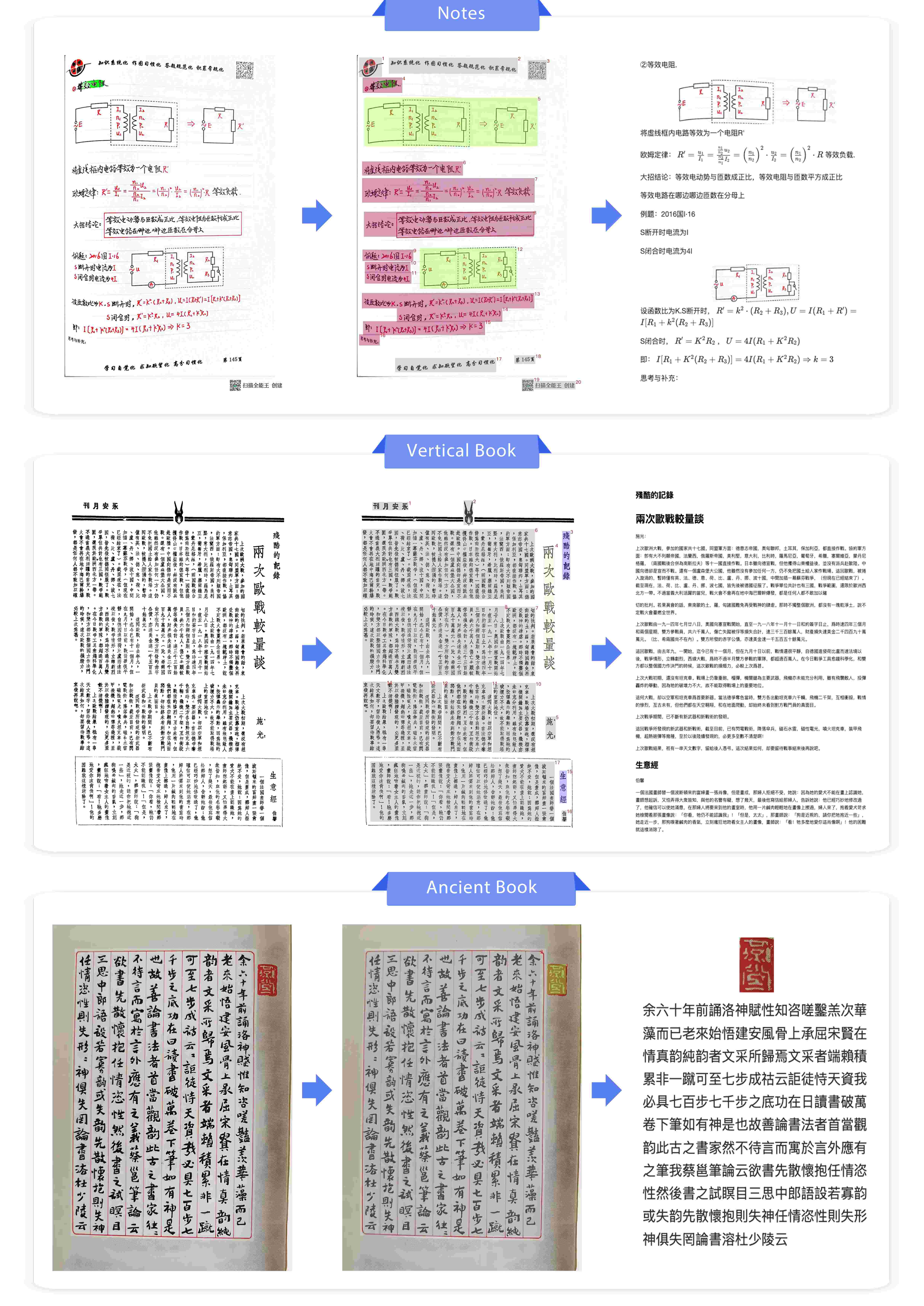} 

\caption{
    The Layout and Markdown Output for Notes, Vertical Book and Ancient Book.
}
\label{fig:overview3}
\end{figure}

\begin{figure}[H]
\centering
\includegraphics[width=0.95\linewidth]{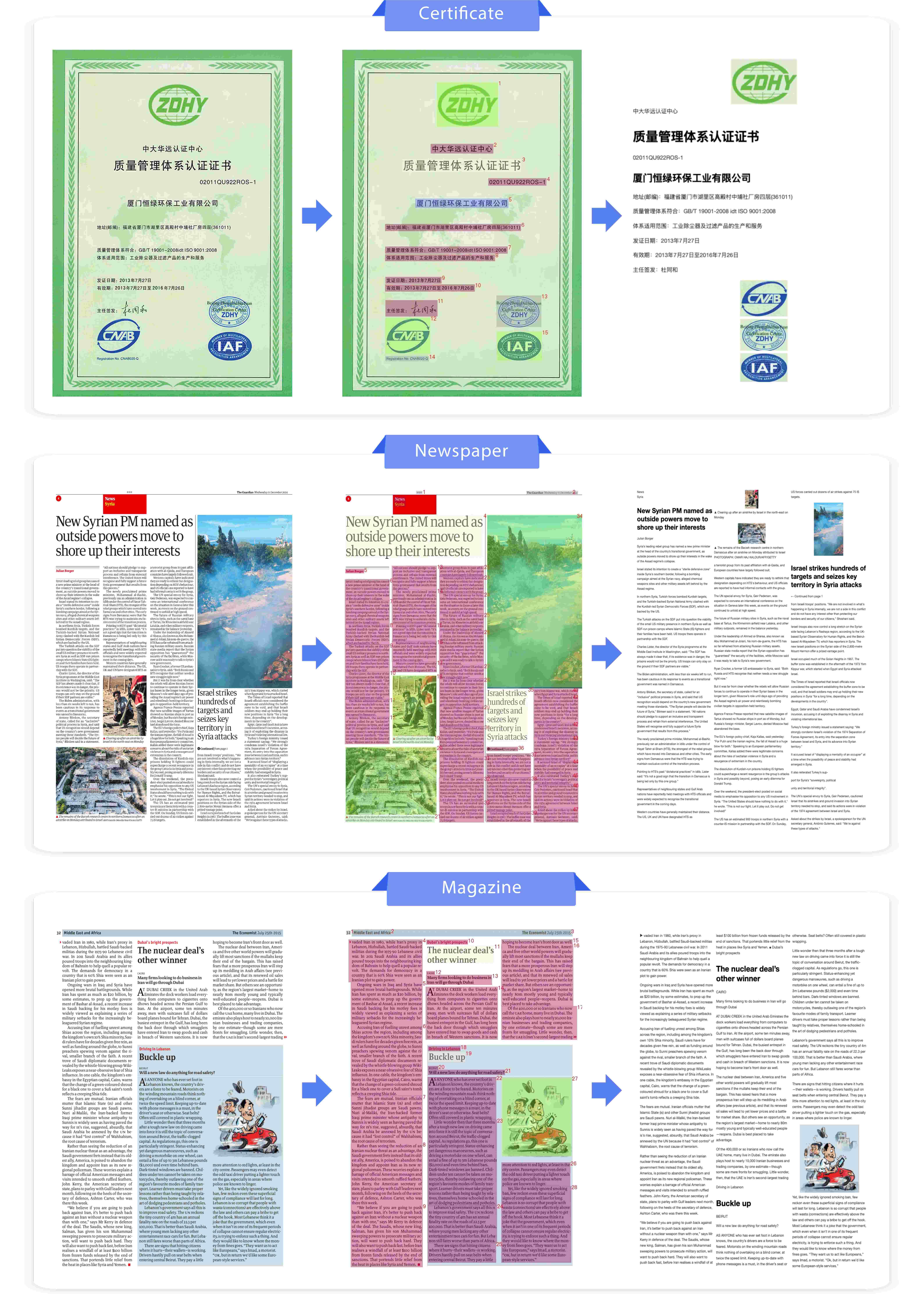} 

\caption{
    \centering
    The Layout and Markdown Output for Certificate, Newspaper and Magazine.
}
\label{fig:overview4}
\end{figure}

Section \ref{subsec:Comprehensive Document Parsing} demonstrates the overall document parsing capability of PaddleOCR-VL. Figures \ref{fig:overview1}-\ref{fig:overview4} are examples of parsing different types of documents in Markdown format.

\begin{figure}[H]
\centering
\includegraphics[width=0.92\linewidth]{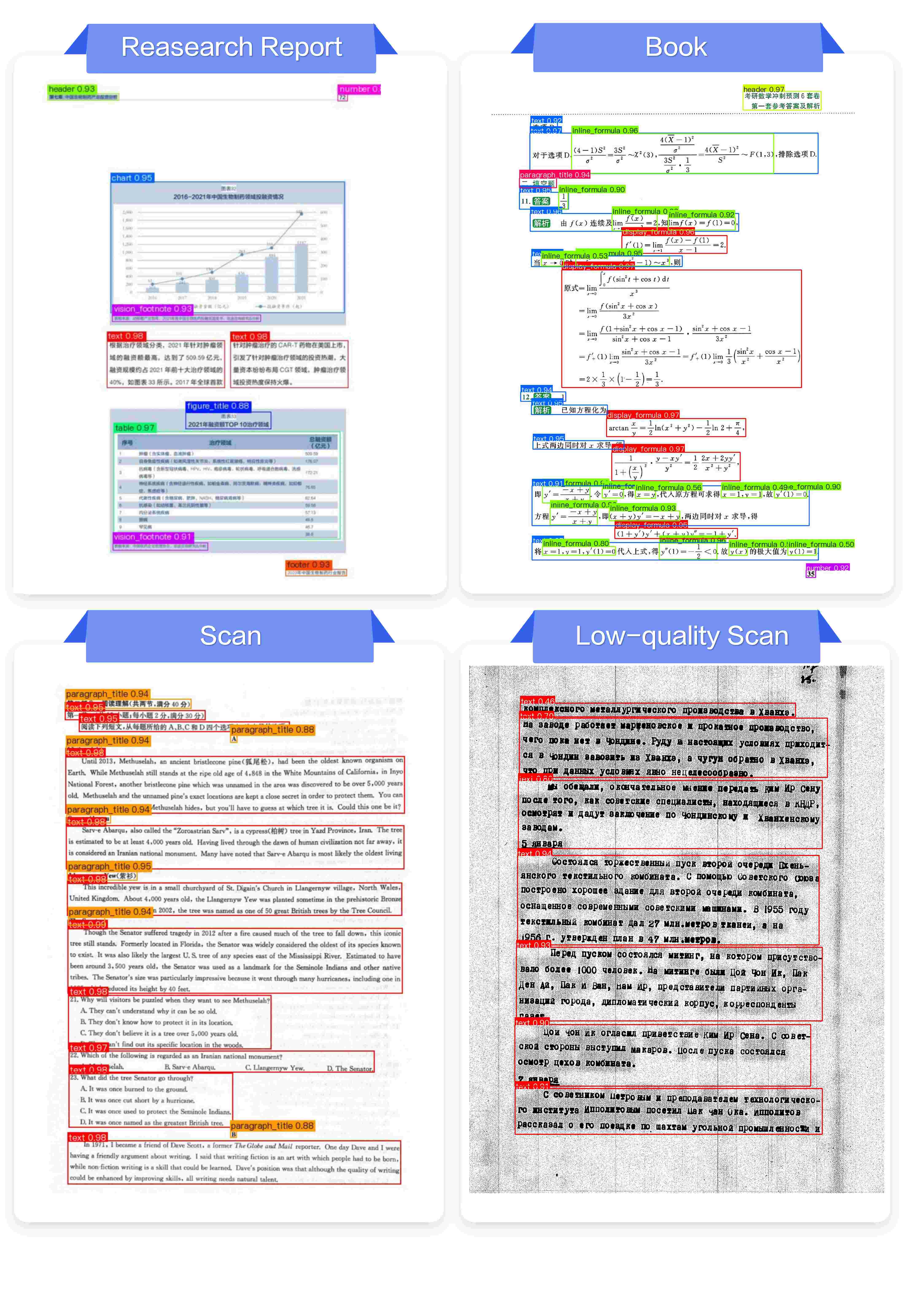} 

\caption{
    \centering
    The Layout Detection results for various types of documents.
}
\label{fig:layout01}
\end{figure}

\begin{figure}[H]
\centering
\includegraphics[width=0.95\linewidth]{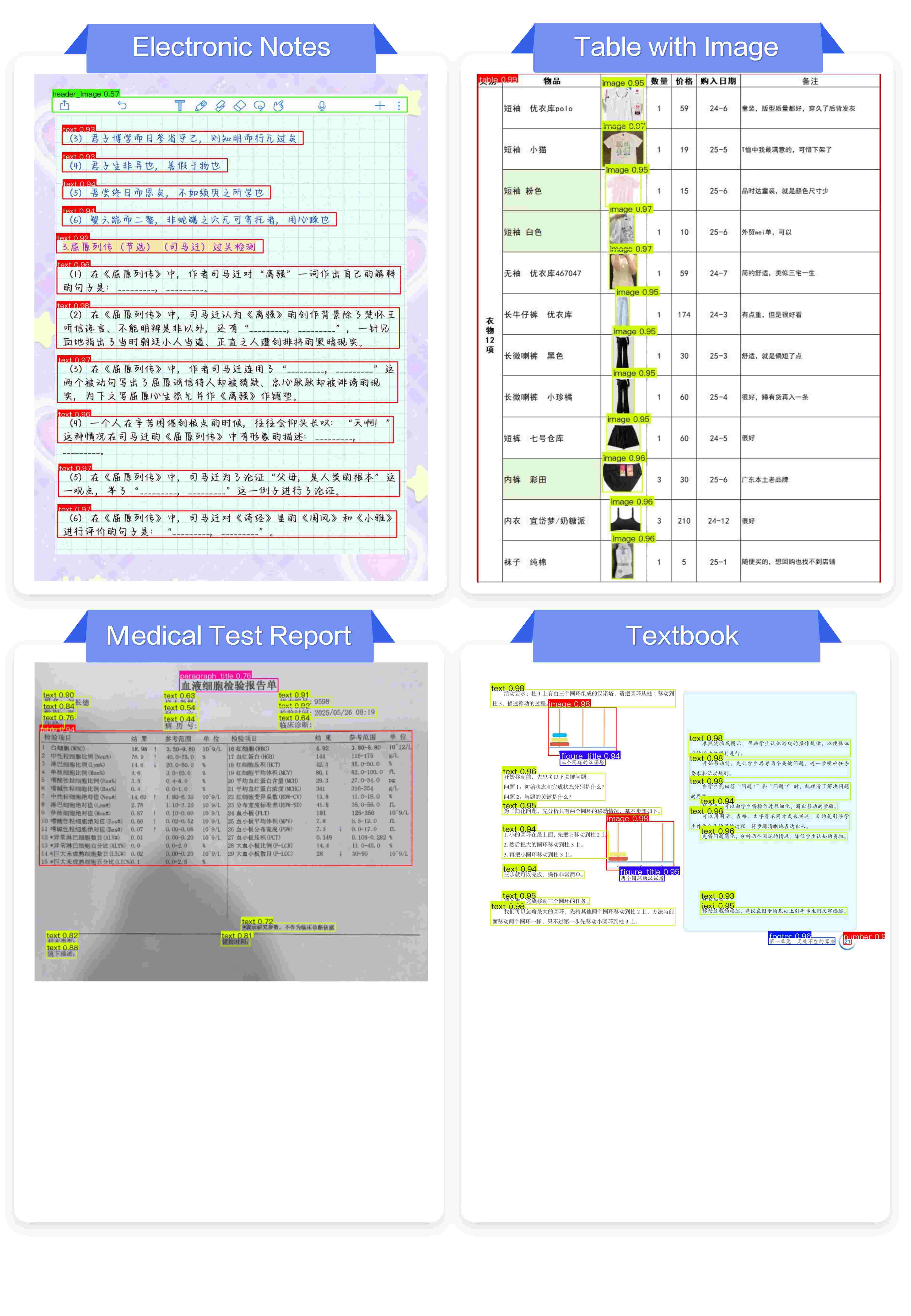} 

\caption{
    \centering
    The Layout Detection results for various types of documents.
}
\label{fig:layout02}
\end{figure}

\begin{figure}[H]
\centering
\includegraphics[width=0.95\linewidth]{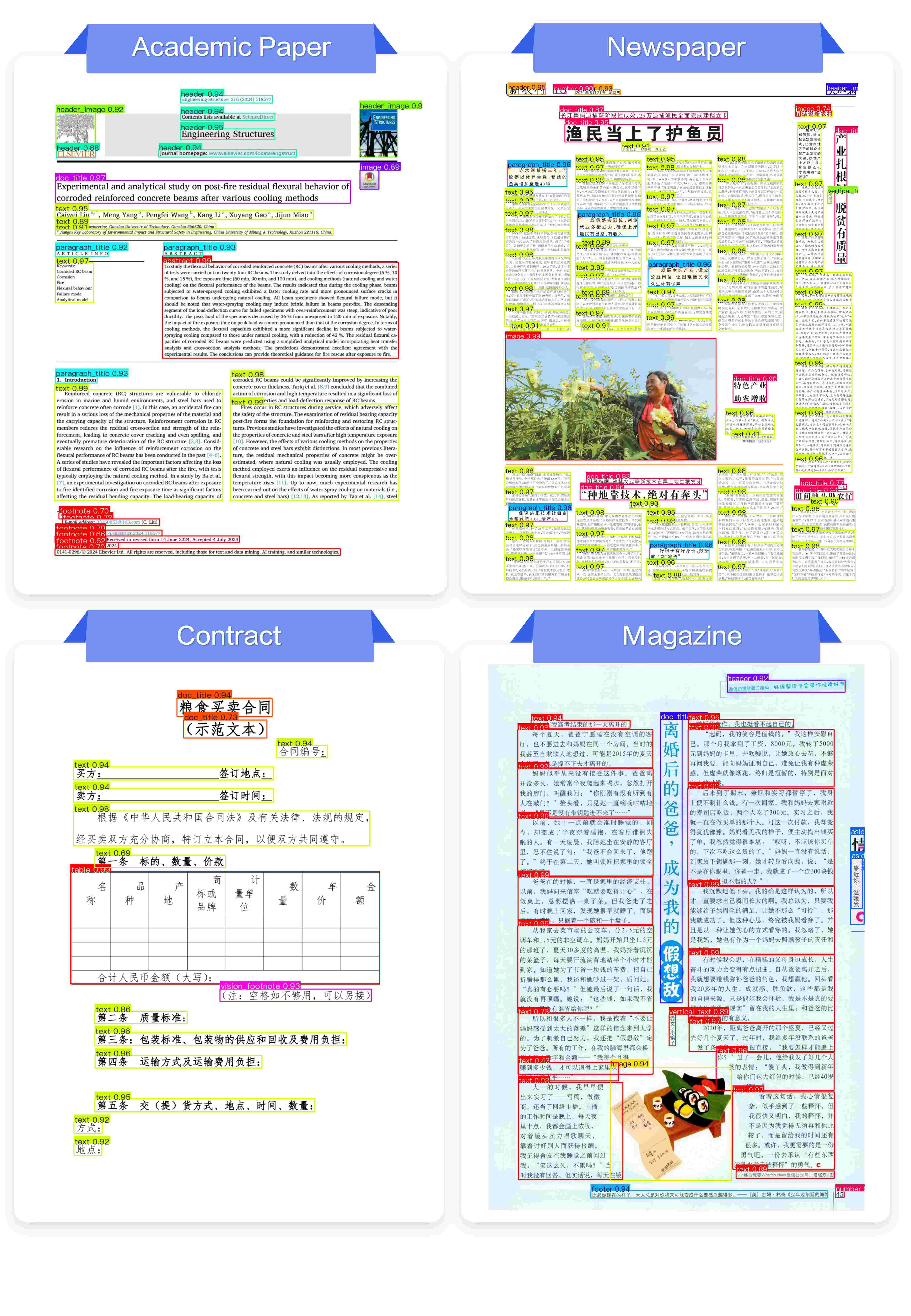} 

\caption{
    \centering
    The Layout Detection results for various types of documents.
}
\label{fig:layout03}
\end{figure}

Figures \ref{fig:layout01}-\ref{fig:layout03} in section \ref{subsec:Layout Detection} illustrate the superior ability of PaddleOCR-VL to process pages featuring intricate or challenging layouts.

\begin{figure}[H]
\centering
\includegraphics[width=0.92\linewidth]{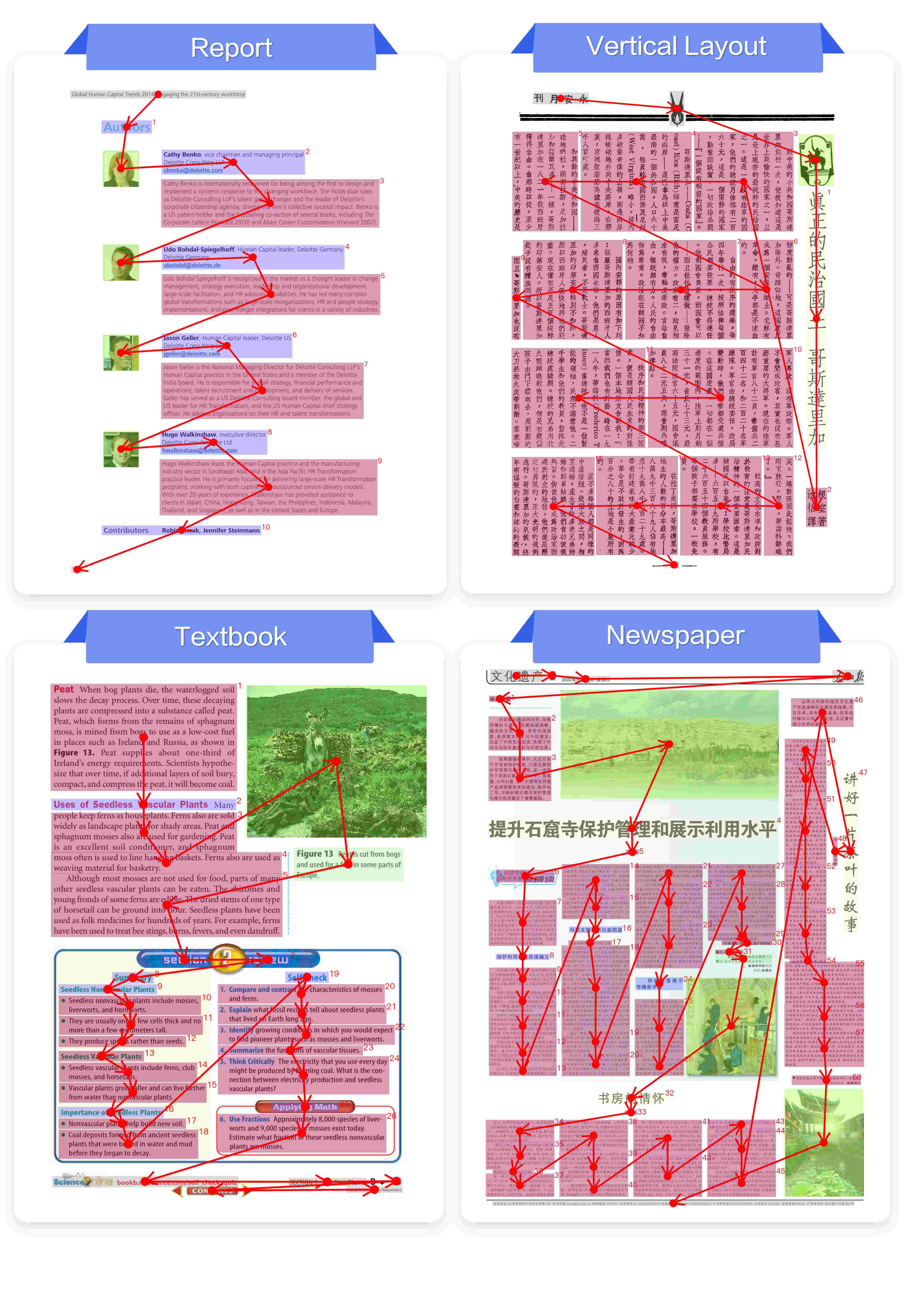} 

\caption{
    \centering
    The Reading Order results for various types of documents.
}
\label{fig:order_01}
\end{figure}

\begin{figure}[H]
\centering
\includegraphics[width=0.95\linewidth]{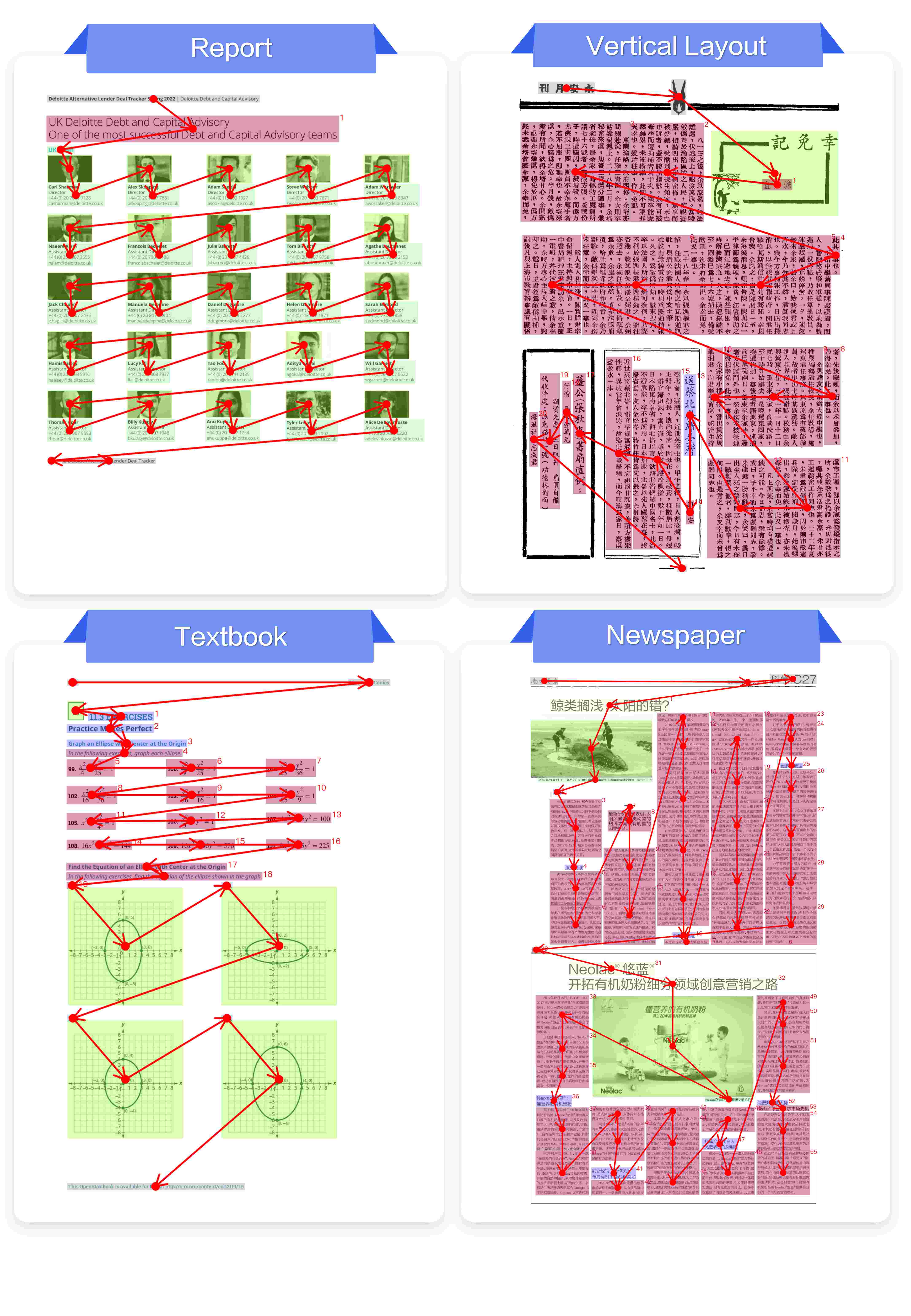} 

\caption{
    \centering
    The Reading Order results for various types of documents.
}
\label{fig:order_02}
\end{figure}

Figures \ref{fig:order_01} and \ref{fig:order_02} in section \ref{subsec:Reading Order} demonstrate that PaddleOCR-VL maintains excellent reading order when faced with complex layouts, such as those found in various reports, textbooks, newspapers, magazines, and even vertical documents.

\begin{figure}[H]
\centering
\includegraphics[width=0.92\linewidth]{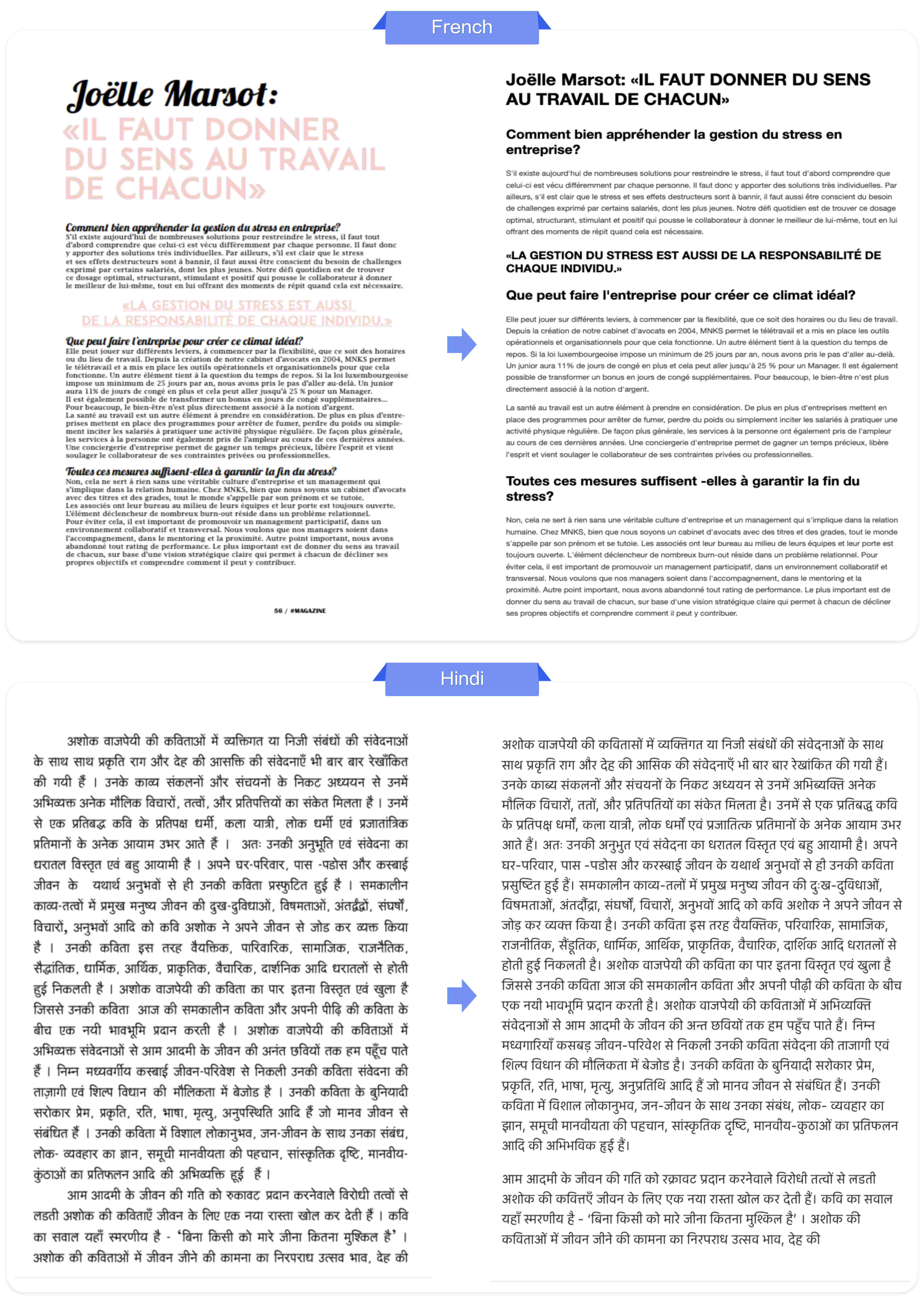} 

\caption{
    \centering
    The markdown output for French and Hindi documents.
}
\label{fig:text_french_hindi}
\end{figure}

\begin{figure}[H]
\centering
\includegraphics[width=0.95\linewidth]{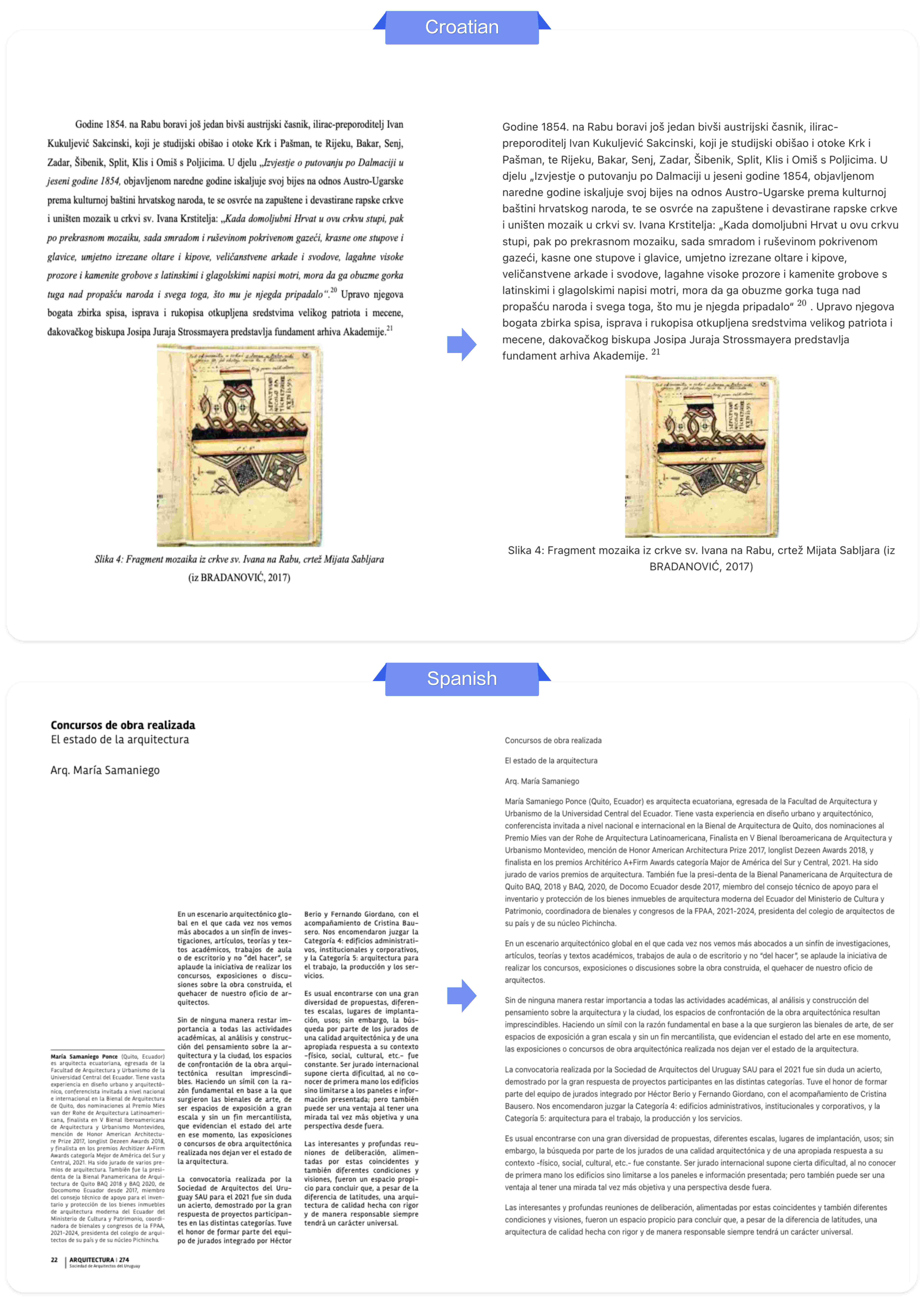} 

\caption{
    \centering
    The markdown output for Croatian and Spanish documents.
}
\label{fig:text_croatian_spanish}
\end{figure}

\begin{figure}[H]
\centering
\includegraphics[width=0.95\linewidth]{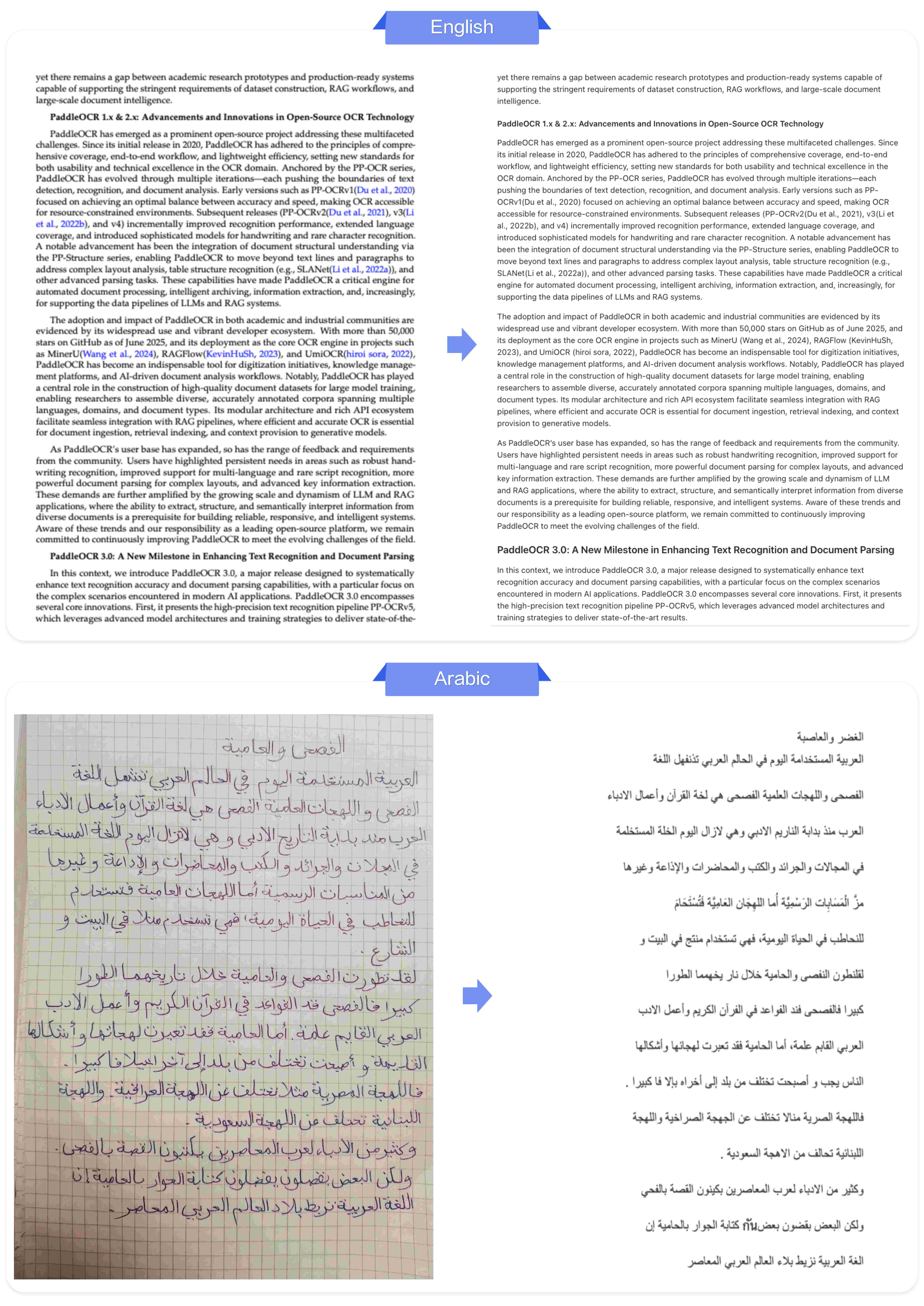} 

\caption{
    \centering
    The markdown output for English and Arabic documents.
}
\label{fig:ext_english_arabic}
\end{figure}

\begin{figure}[H]
\centering
\includegraphics[width=0.95\linewidth]{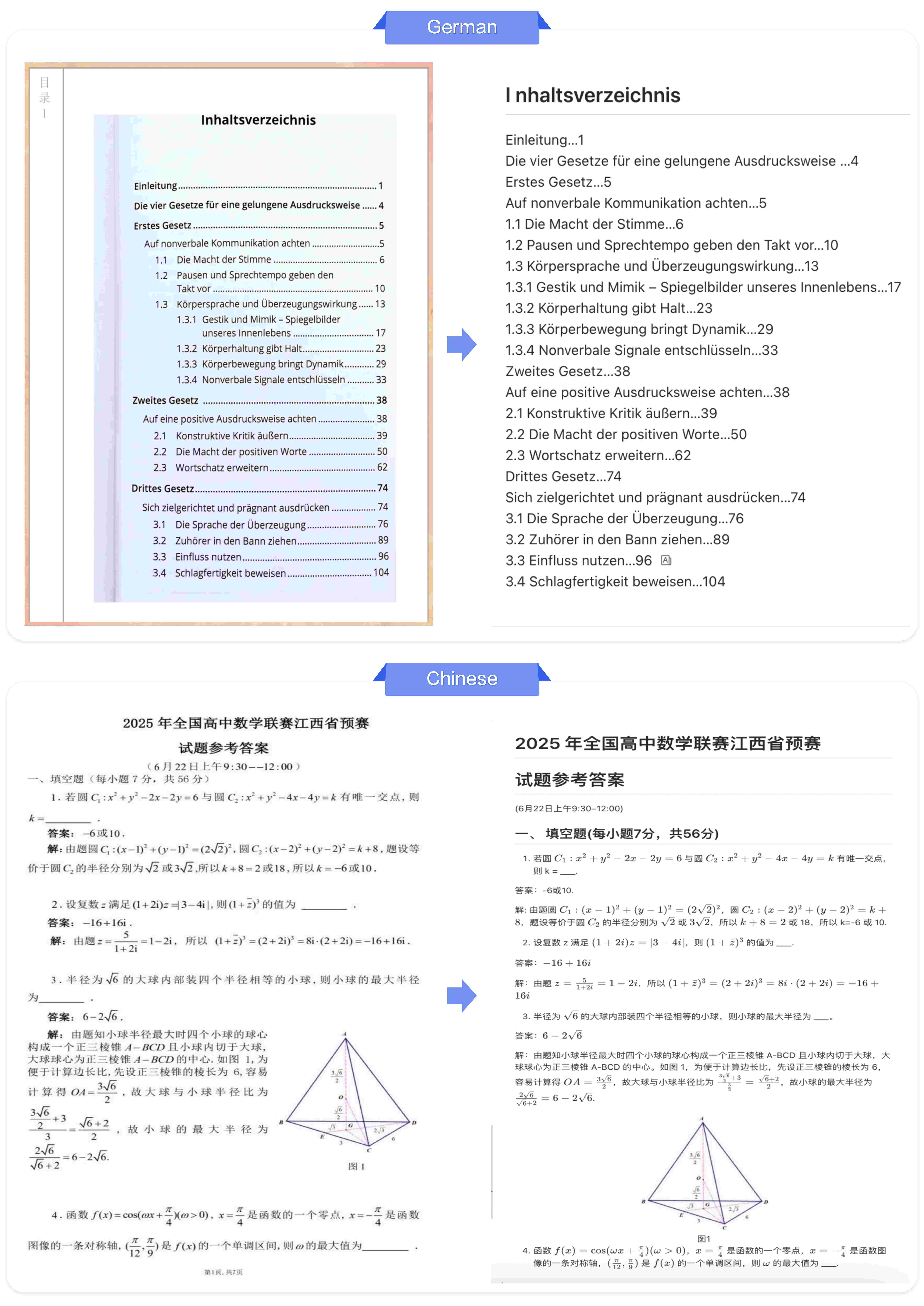} 

\caption{
    \centering
    The markdown output for German and Chinese documents.
}
\label{fig:text_german_chinese}
\end{figure}

\begin{figure}[H]
\centering
\includegraphics[width=0.95\linewidth]{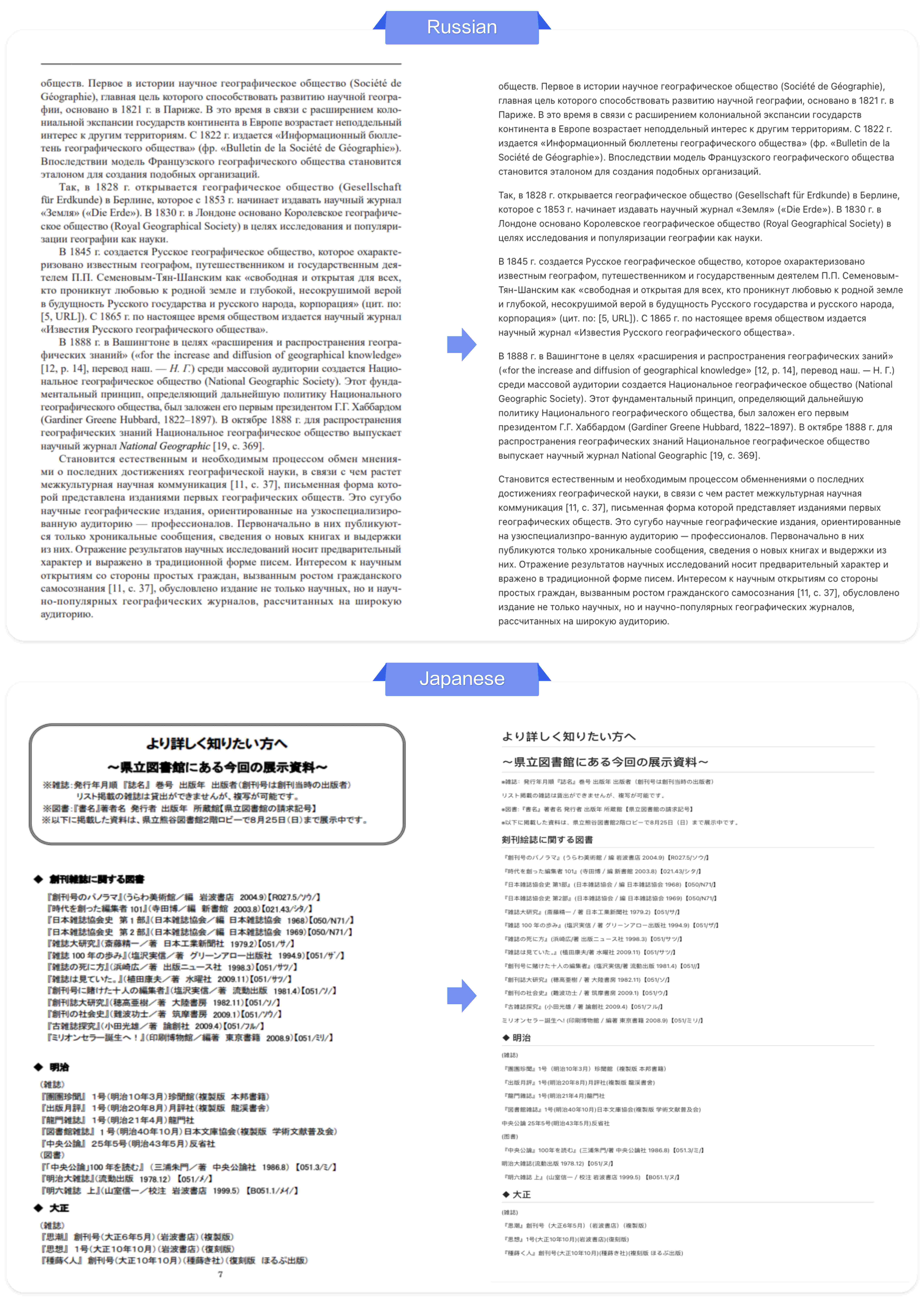} 

\caption{
    \centering
    The markdown output for Russian and Japanese documents.
}
\label{fig:text_russian_japanese}
\end{figure}

\begin{figure}[H]
\centering
\includegraphics[width=0.95\linewidth]{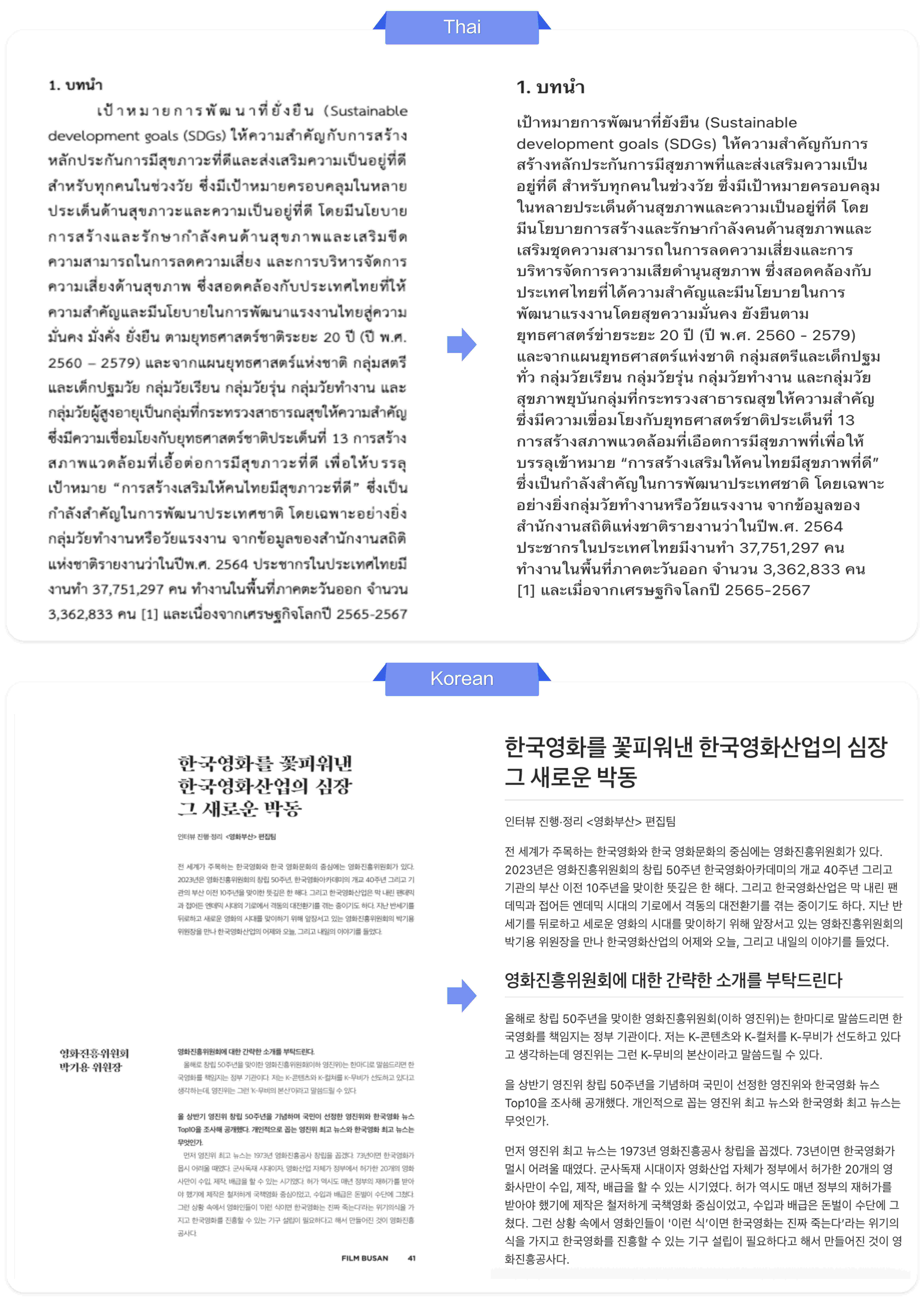} 

\caption{
    \centering
    The markdown output for Thai and Korean documents.
}
\label{fig:text_thai_korean}
\end{figure}

\begin{figure}[H]
\centering
\includegraphics[width=0.92\linewidth]{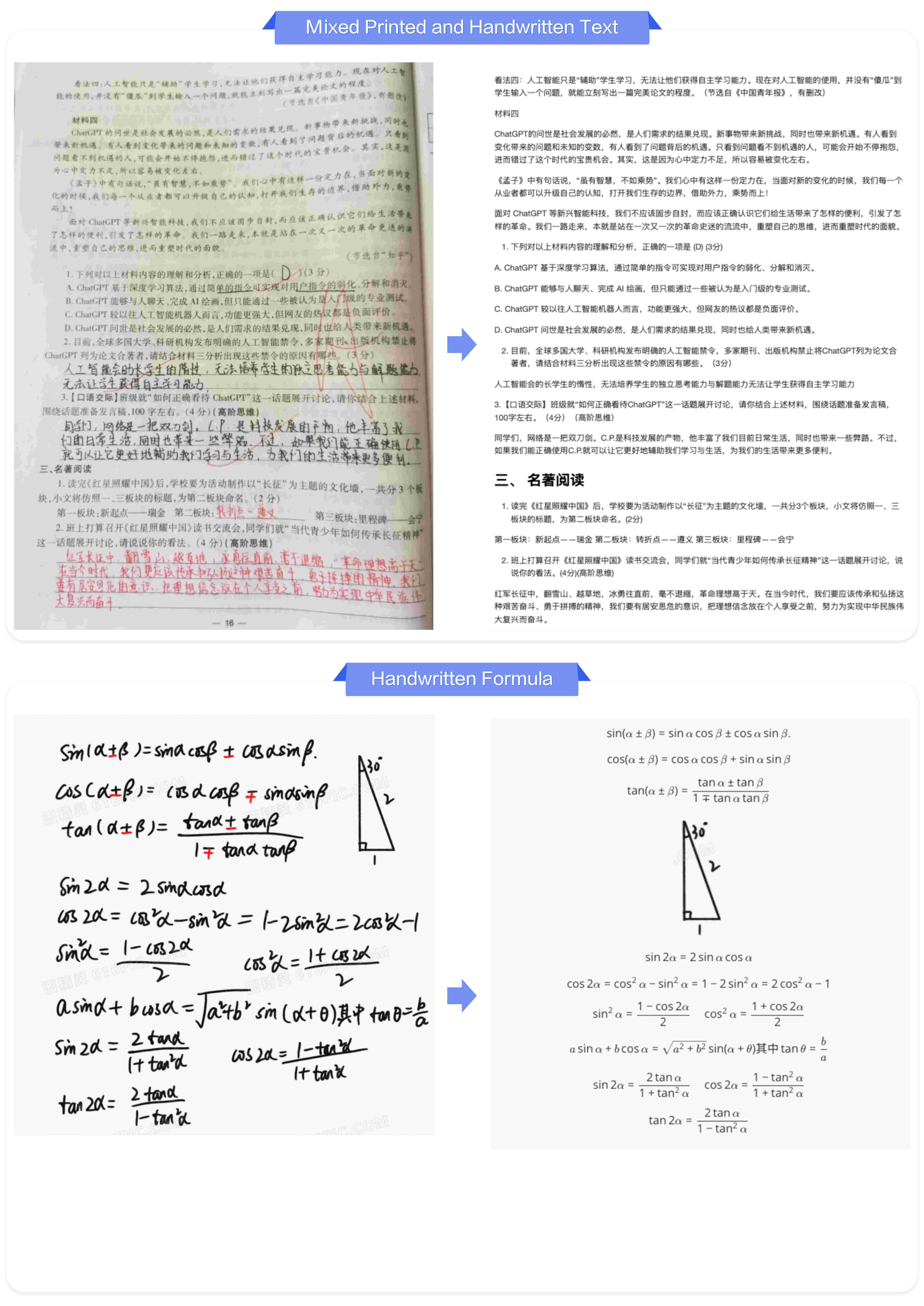} 

\caption{
    \centering
    The markdown output for Mixed Printed Handwritten Text and Handwritten Formula documents.
}
\label{fig:text_handwriting01}
\end{figure}

\begin{figure}[H]
\centering
\includegraphics[width=0.95\linewidth]{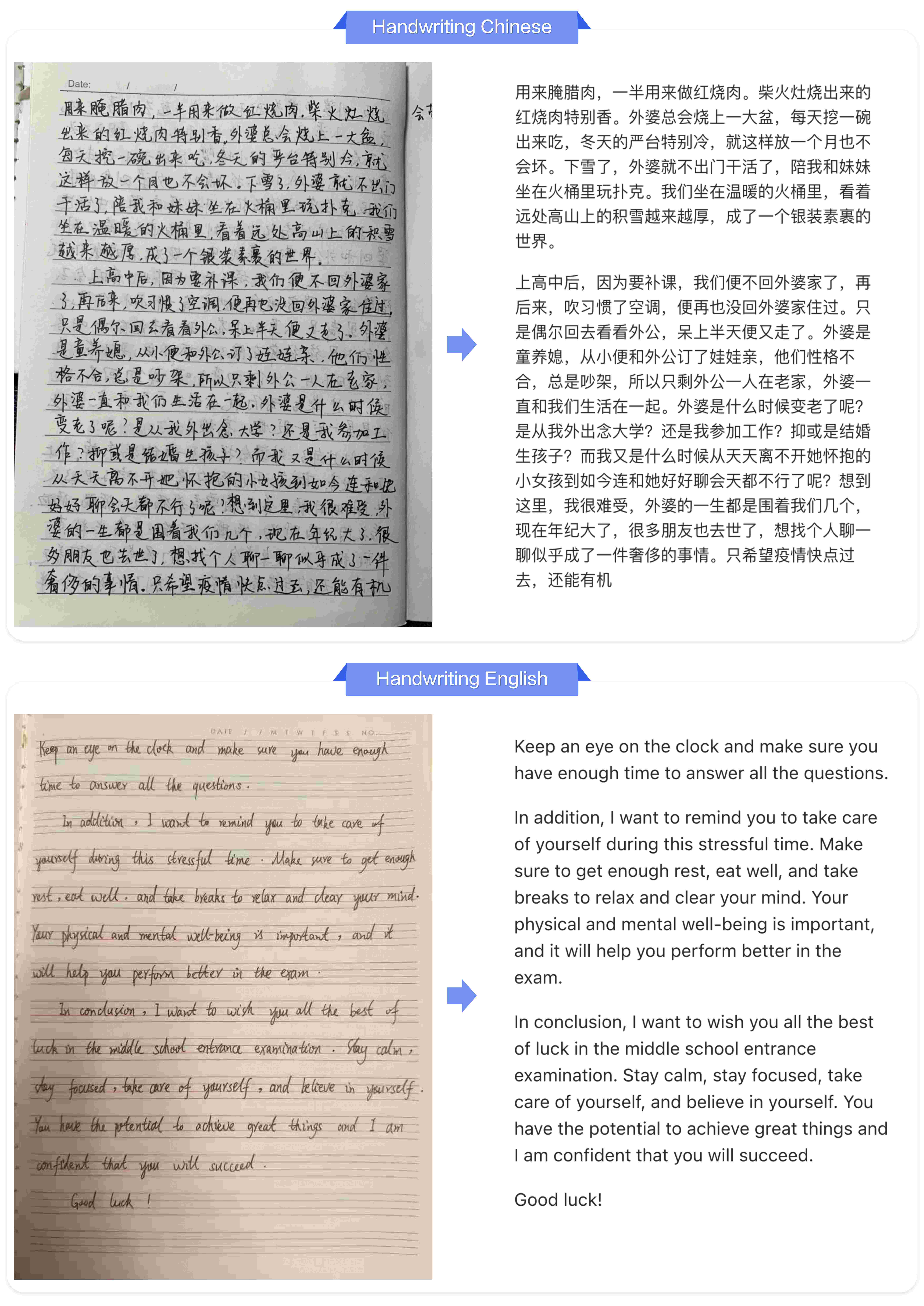} 

\caption{
    \centering
    The markdown output for Handwriting Chinese and Handwriting English documents.
}
\label{fig:text_handwriting02}
\end{figure}

\begin{figure}[H]
\centering
\includegraphics[width=0.95\linewidth]{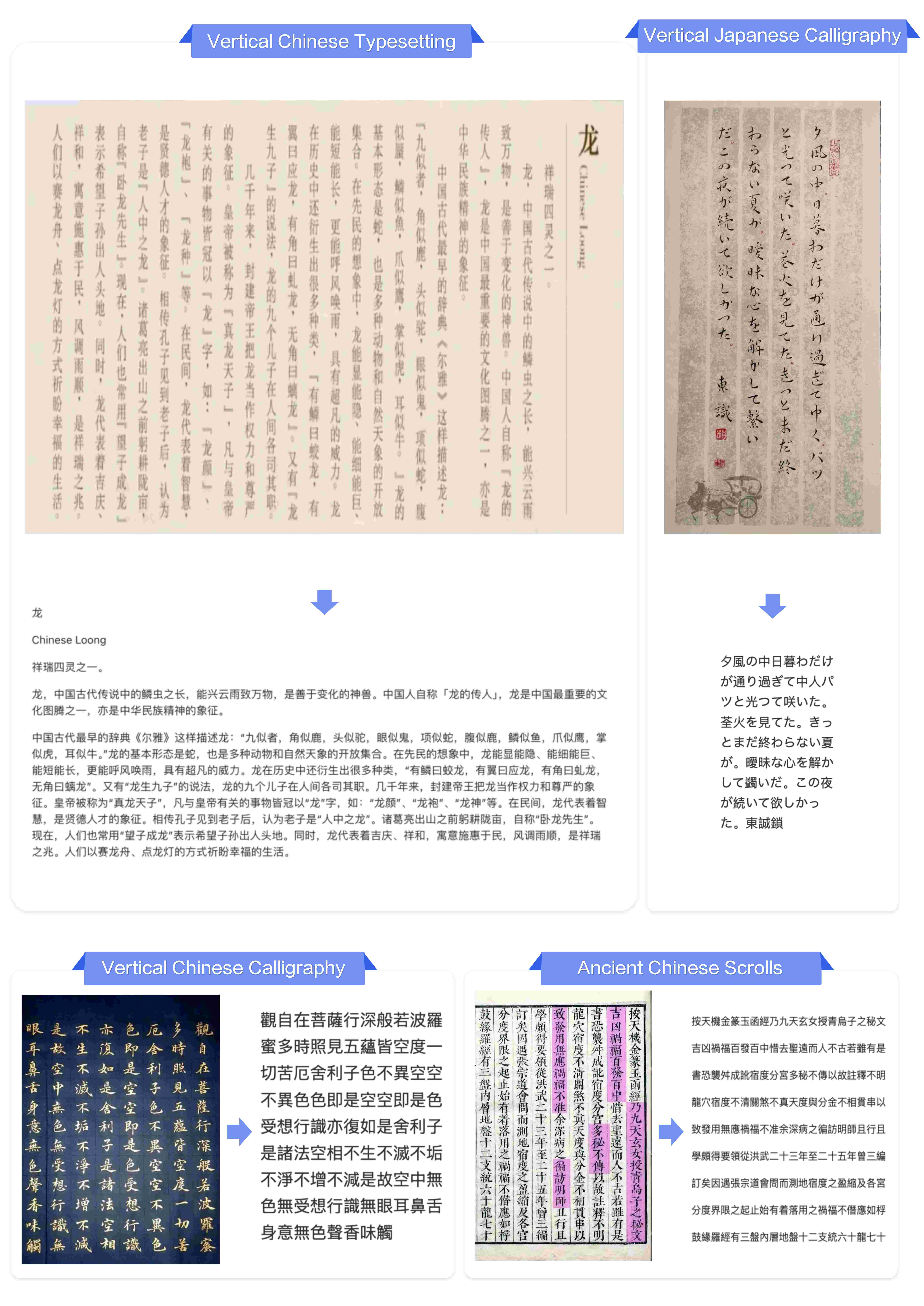} 

\caption{
    \centering
    The markdown output for various types of vertical documents.
}
\label{fig:text_vertical}
\end{figure}

Section \ref{subsec:Text Recognition} highlights the robust text recognition performance of PaddleOCR-VL in challenging cases, including multilingual text, handwriting text, and vertical text, which are presented in Figures \ref{fig:text_french_hindi}-\ref{fig:text_vertical}.

\begin{figure}[H]
\centering
\includegraphics[width=0.95\linewidth]{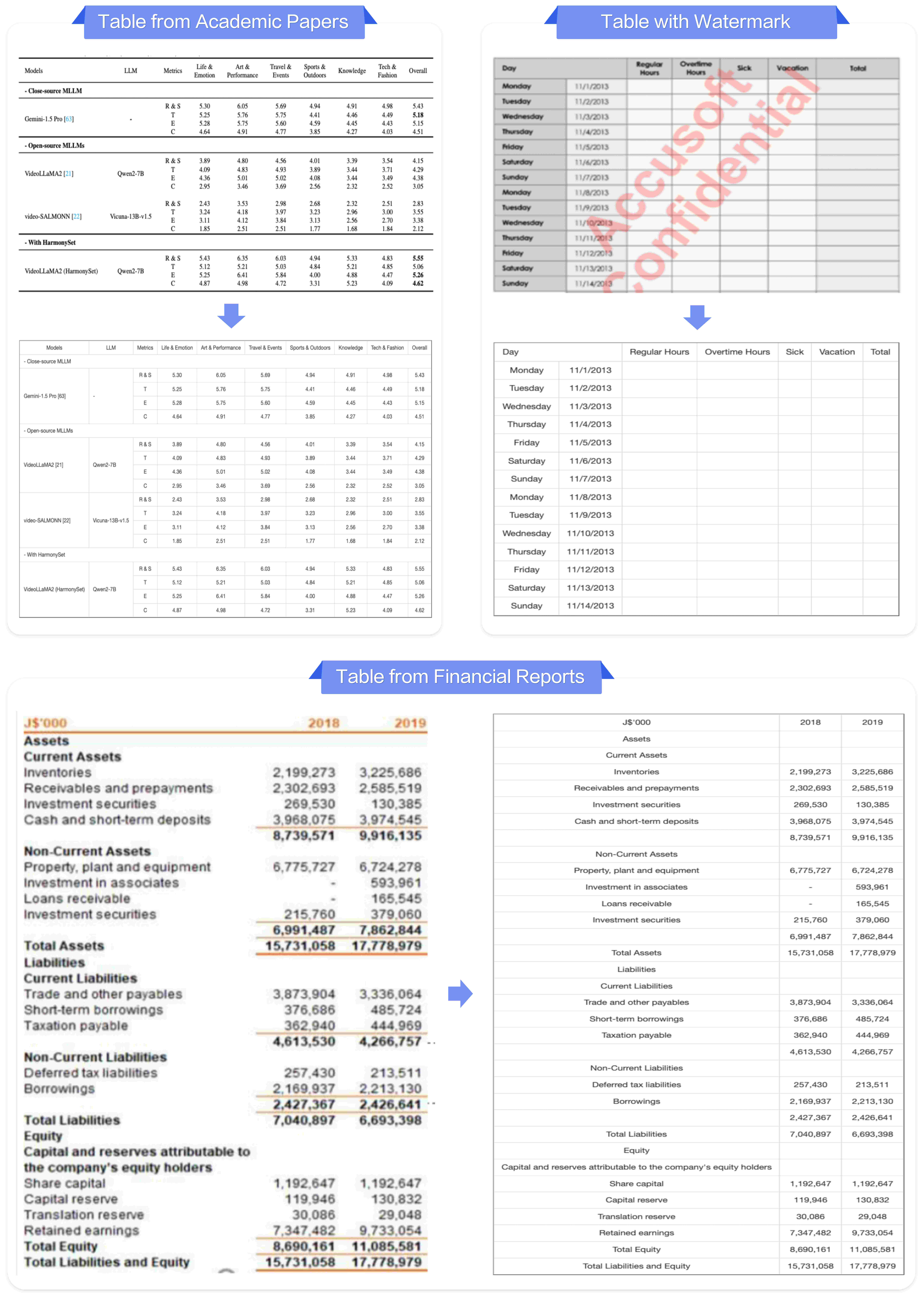} 

\caption{
    \centering
    The markdown output for various types of Tables.
}
\label{fig:table_01}
\end{figure}

\begin{figure}[H]
\centering
\includegraphics[width=0.95\linewidth]{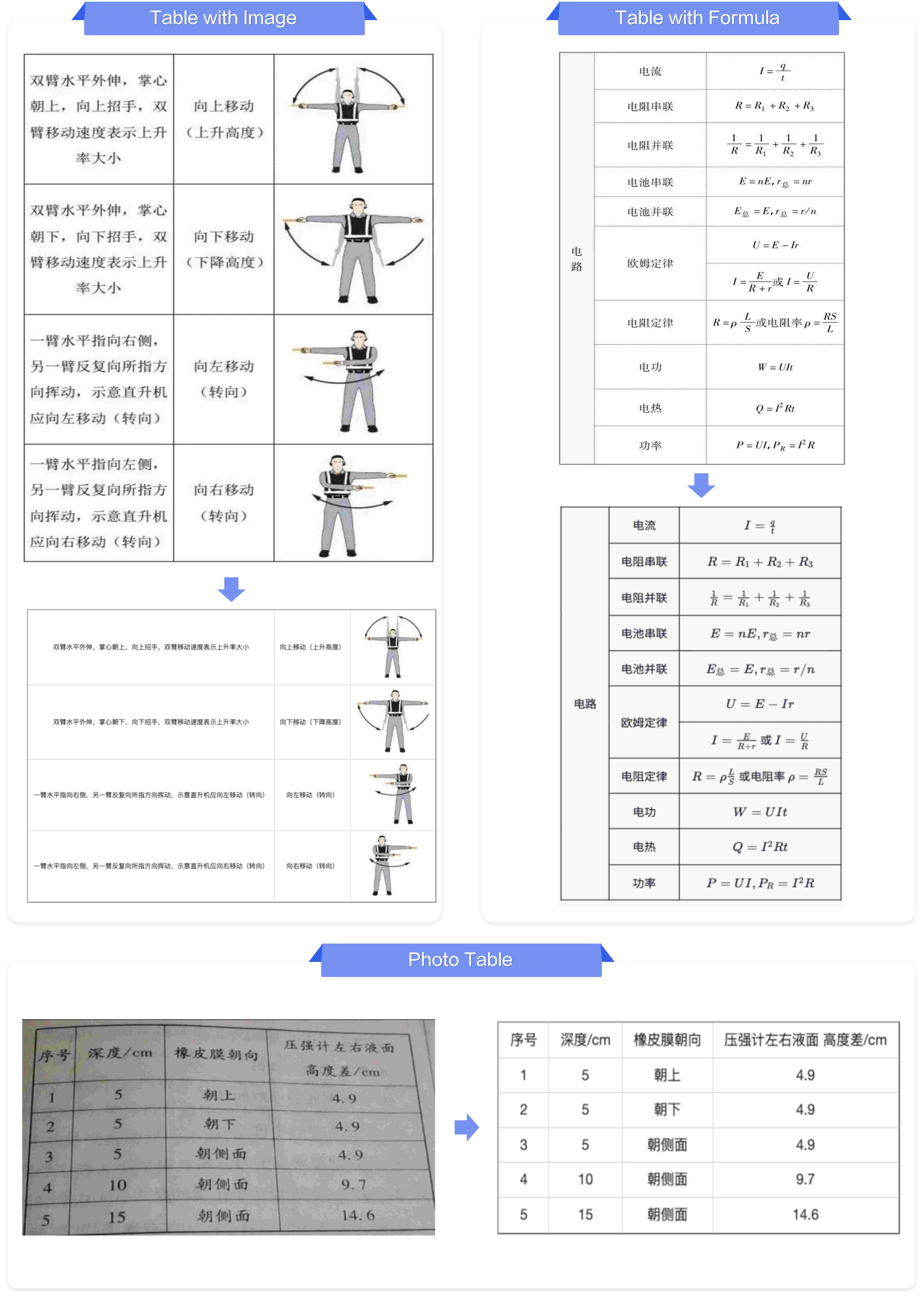} 

\caption{
    \centering
    The markdown output for various types of Tables.
}
\label{fig:table_02}
\end{figure}

The model's table recognition abilities are demonstrated in section \ref{subsec:Table Recognition}. Figures \ref{fig:table_01} and \ref{fig:table_02} showcase its robust handling of a wide array of table formats, including tables from academic papers, tables from financial reports, tables with watermark, tables with image, tables with formulas and photograph of tables.

\begin{figure}[H]
\centering
\includegraphics[width=0.95\linewidth]{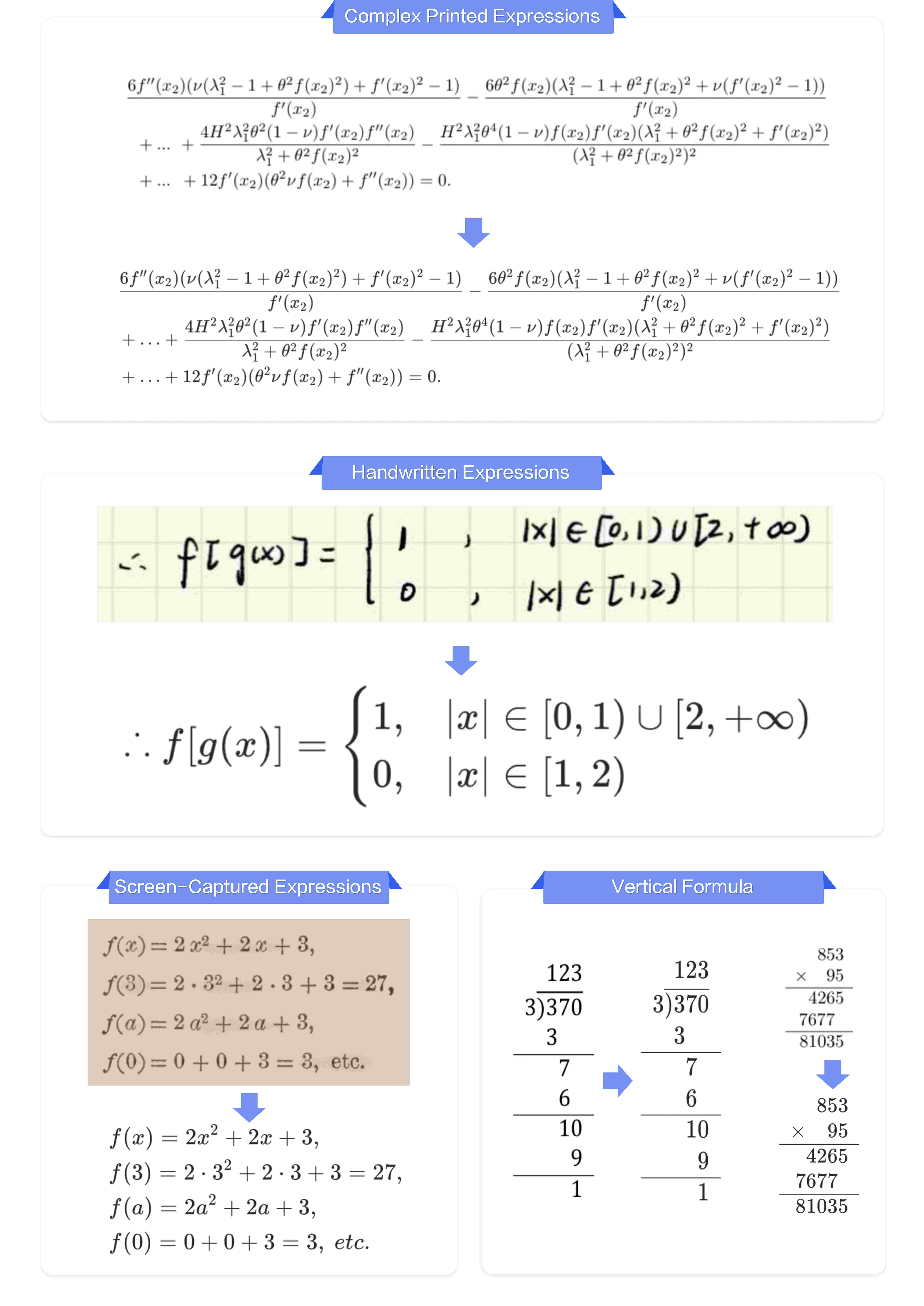} 

\caption{
    \centering
    The markdown output for various types of Formulas.
}
\label{fig:formula_EN}
\end{figure}

\begin{figure}[H]
\centering
\includegraphics[width=0.95\linewidth]{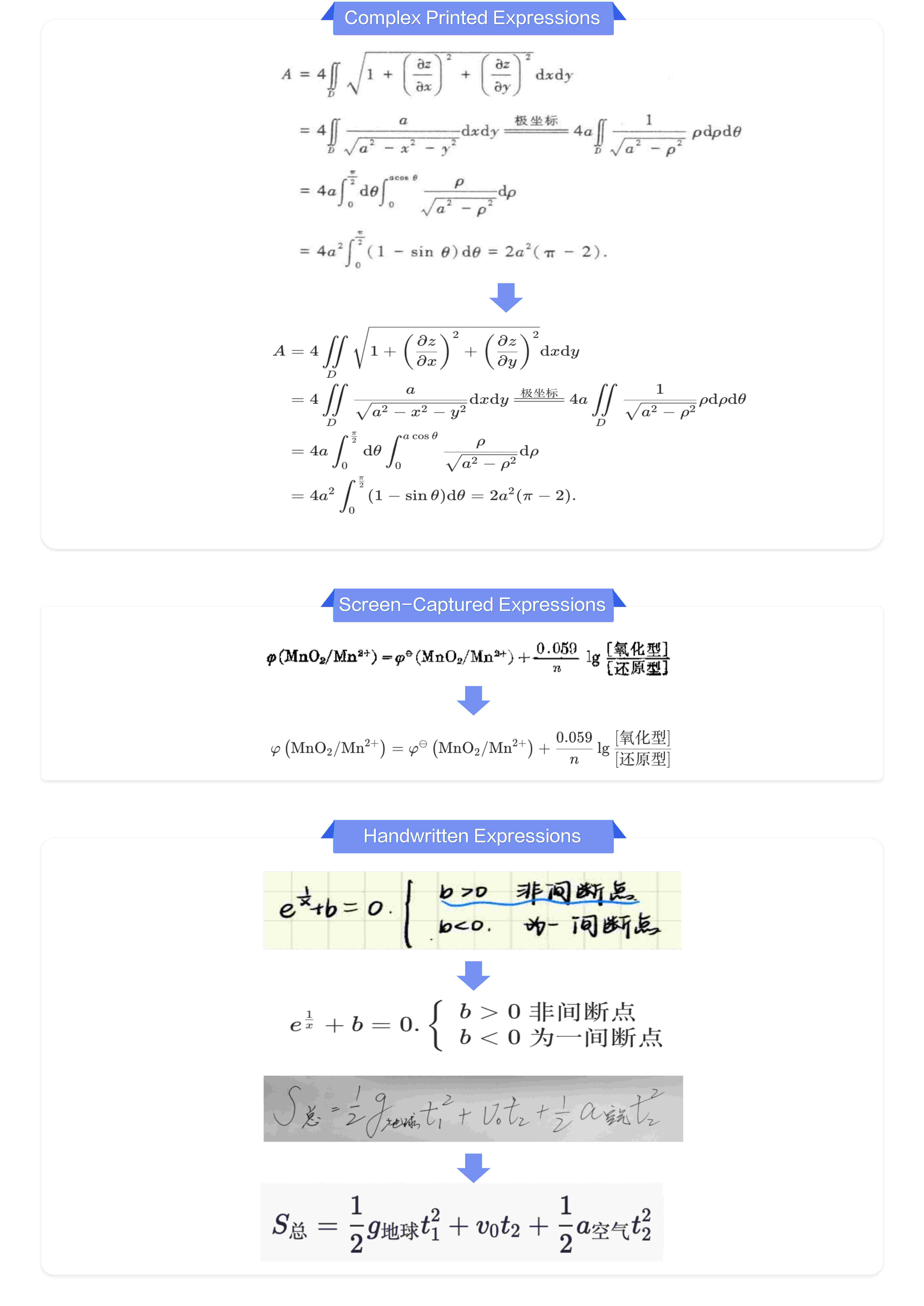} 

\caption{
    \centering
    The markdown output for various types of Formulas.
}
\label{fig:formula_ZH}
\end{figure}

Figures in section \ref{subsec:Formula Recognition} detail the formula recognition performance. Figure \ref{fig:formula_EN} demonstrates the ability to handle various types of english formulas including complex printed expressions, handwritten expressions screen-captured expressions and vertical formula, while Figure \ref{fig:formula_ZH} focuses on the ability to handle formulas that contain Chinese characters.

\begin{figure}[H]
\centering
\includegraphics[width=0.95\linewidth]{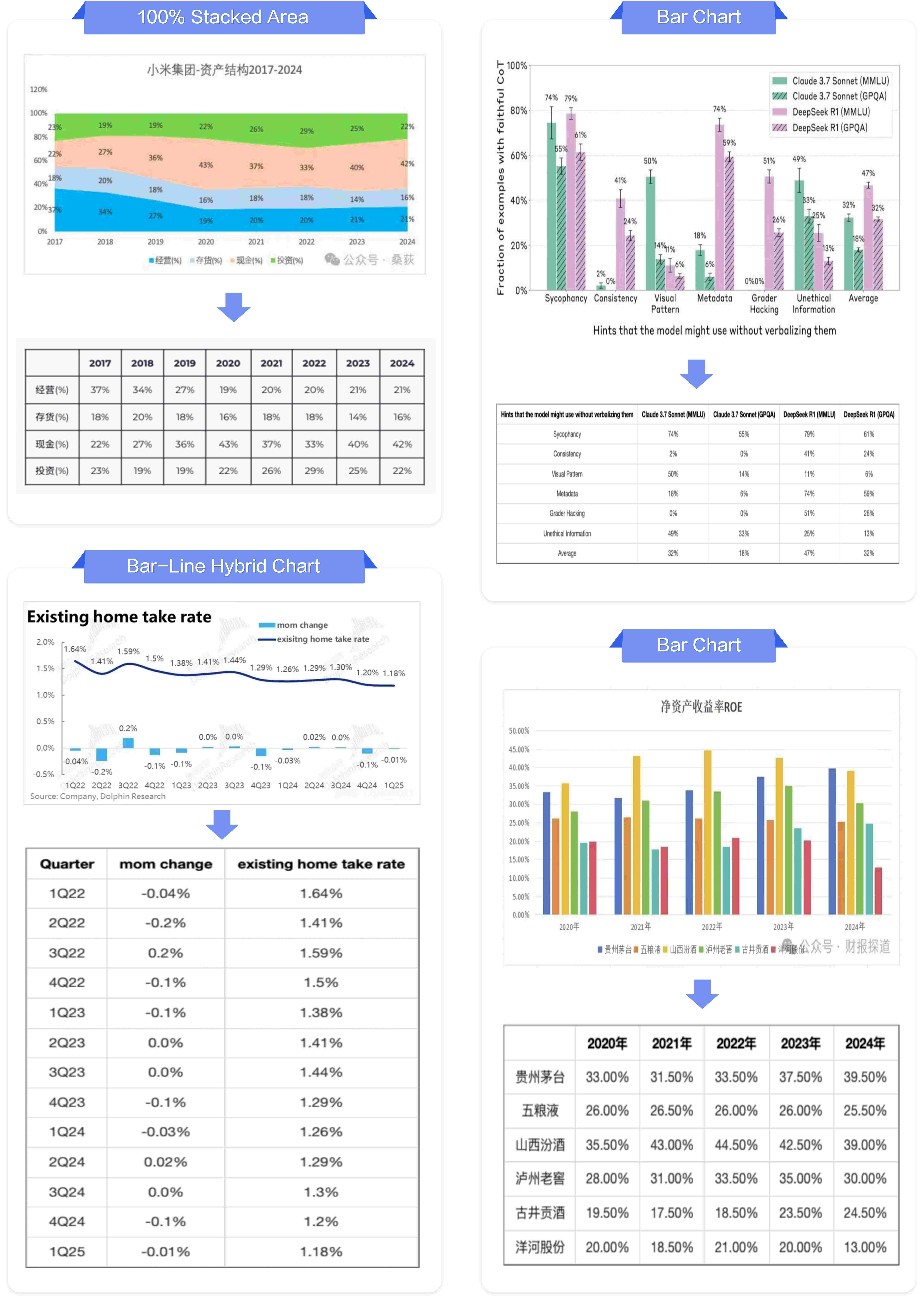} 

\caption{
    \centering
    The markdown output for various types of Charts.
}
\label{fig:chart_01}
\end{figure}

\begin{figure}[H]
\centering
\includegraphics[width=0.95\linewidth]{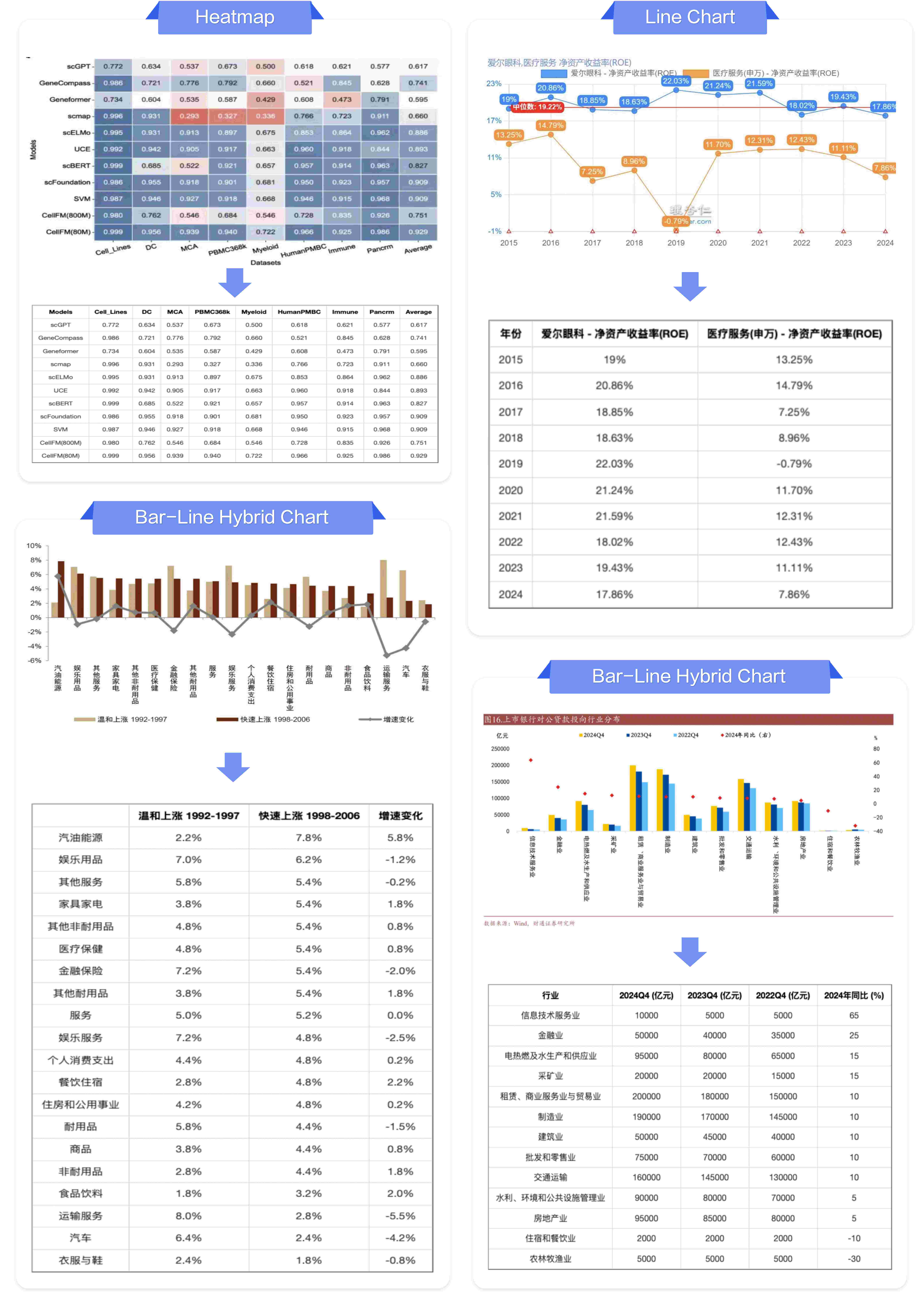} 

\caption{
    \centering
    The markdown output for various types of Charts.
}
\label{fig:chart_02}
\end{figure}

\begin{figure}[!htbp]
\centering
\includegraphics[width=0.92\linewidth]{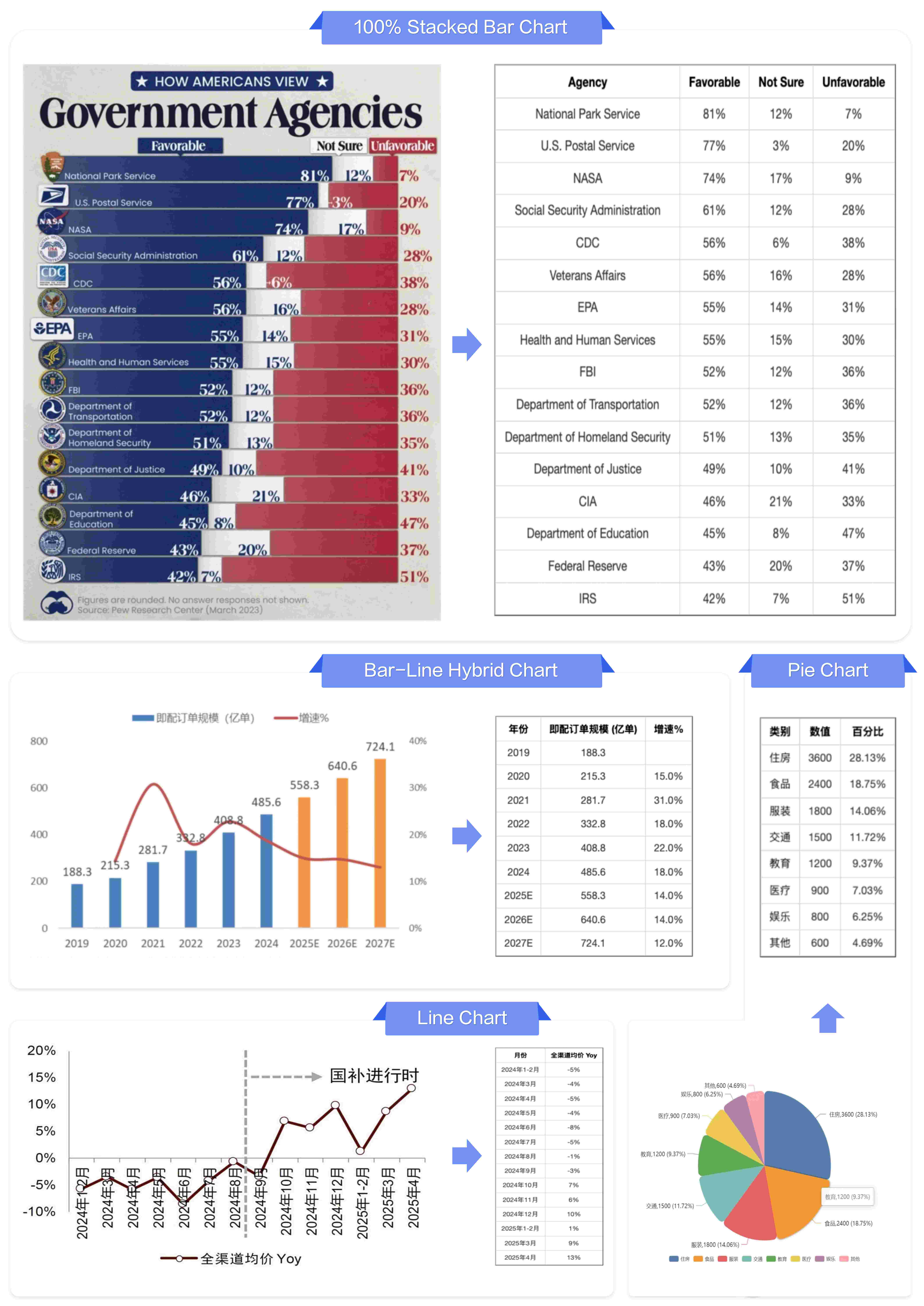} 

\caption{
    \centering
    The markdown output for various types of Charts.
}
\label{fig:chart_03}
\end{figure}

In section \ref{subsec:Chart Recognition}, PaddleOCR-VL demonstrates impressive chart recognition capabilities, a feature currently lacking in many expert OCR VLMs like MinerU2.5 \cite{MinerU2}, dots.ocr \cite{dotsocr} or MonkeyOCR \cite{li2025monkeyocr}. Figures \ref{fig:chart_01}-\ref{fig:chart_03} showcase our ability to parse various chart types, including pie charts, bar charts, line charts, bar-line hybrid charts and heatmap.

\clearpage 
\newpage
\subsection{Comprehensive Document Parsing}
\label{subsec:Comprehensive Document Parsing}

\clearpage 
\newpage

\clearpage 
\newpage

\clearpage 
\newpage

\newpage
\subsection{Layout Detection}
\label{subsec:Layout Detection}

\newpage
\subsection{Reading Order}
\label{subsec:Reading Order}

\newpage
\subsection{Text Recognition }
\label{subsec:Text Recognition}

\subsubsection{Multilingual Text Recognition}

\newpage

\subsubsection{Handwriting Text Recognition}

\subsubsection{Vertical Text Recognition}

\clearpage 
\newpage

\subsection{Table Recognition}
\label{subsec:Table Recognition}

\clearpage 
\newpage

\subsection{Formula Recognition}
\label{subsec:Formula Recognition}

\newpage
\subsection{Chart Recognition}
\label{subsec:Chart Recognition}

\newpage
\section{Compare with Others}
\label{subsec:Compare with Others}

\begin{figure}[H]
\centering
\includegraphics[width=0.92\linewidth]{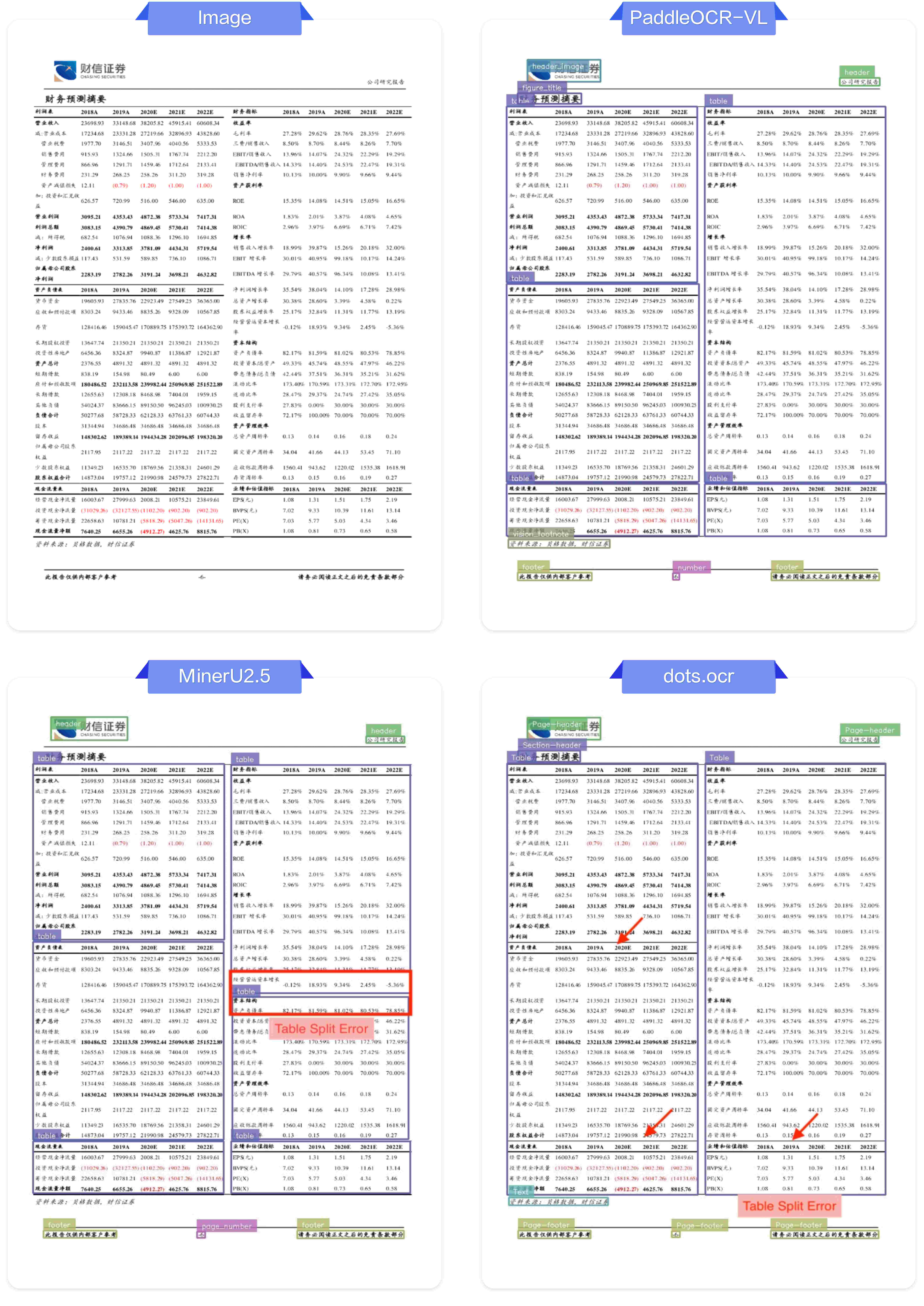} 

\caption{
    \centering
    Compare with others in Layout Detection.
}
\label{fig:cmp_layout_01}
\end{figure}

\begin{figure}[H]
\centering
\includegraphics[width=0.95\linewidth]{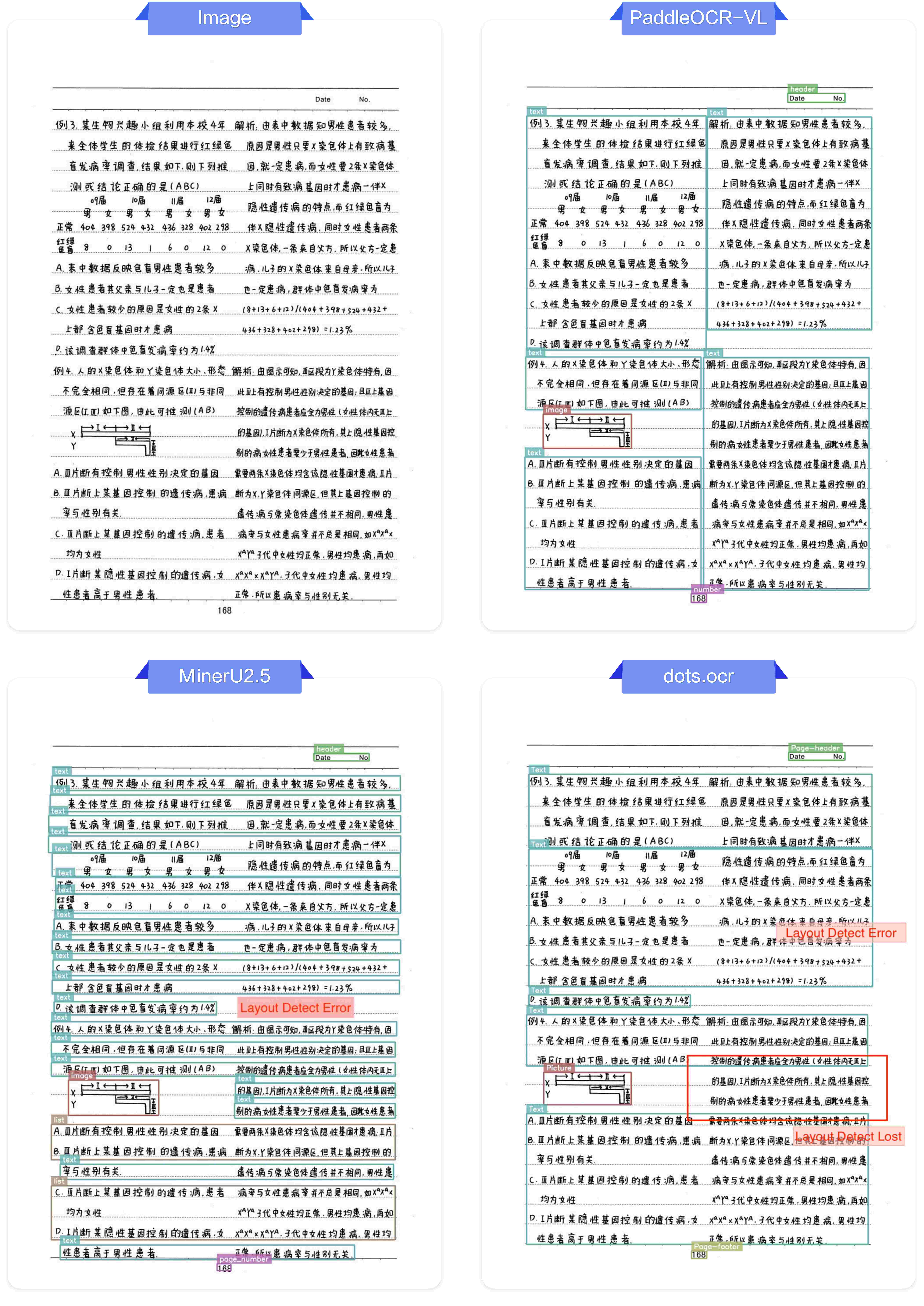} 

\caption{
    \centering
    Compare with others in Layout Detection.
}
\label{fig:cmp_layout_02}
\end{figure}

PaddleOCR-VL showcases superior performance in scenarios involving PDF pages with complex layout, consistently outperforming existing state-of-the-art (SOTA) models. This is evident from Figures \ref{fig:cmp_layout_01} and \ref{fig:cmp_layout_02}, which highlight its exceptional capability in handling pages with intricate layouts and unique elements, surpassing other solutions.

\begin{figure}[H]
\centering
\includegraphics[width=0.90\linewidth]{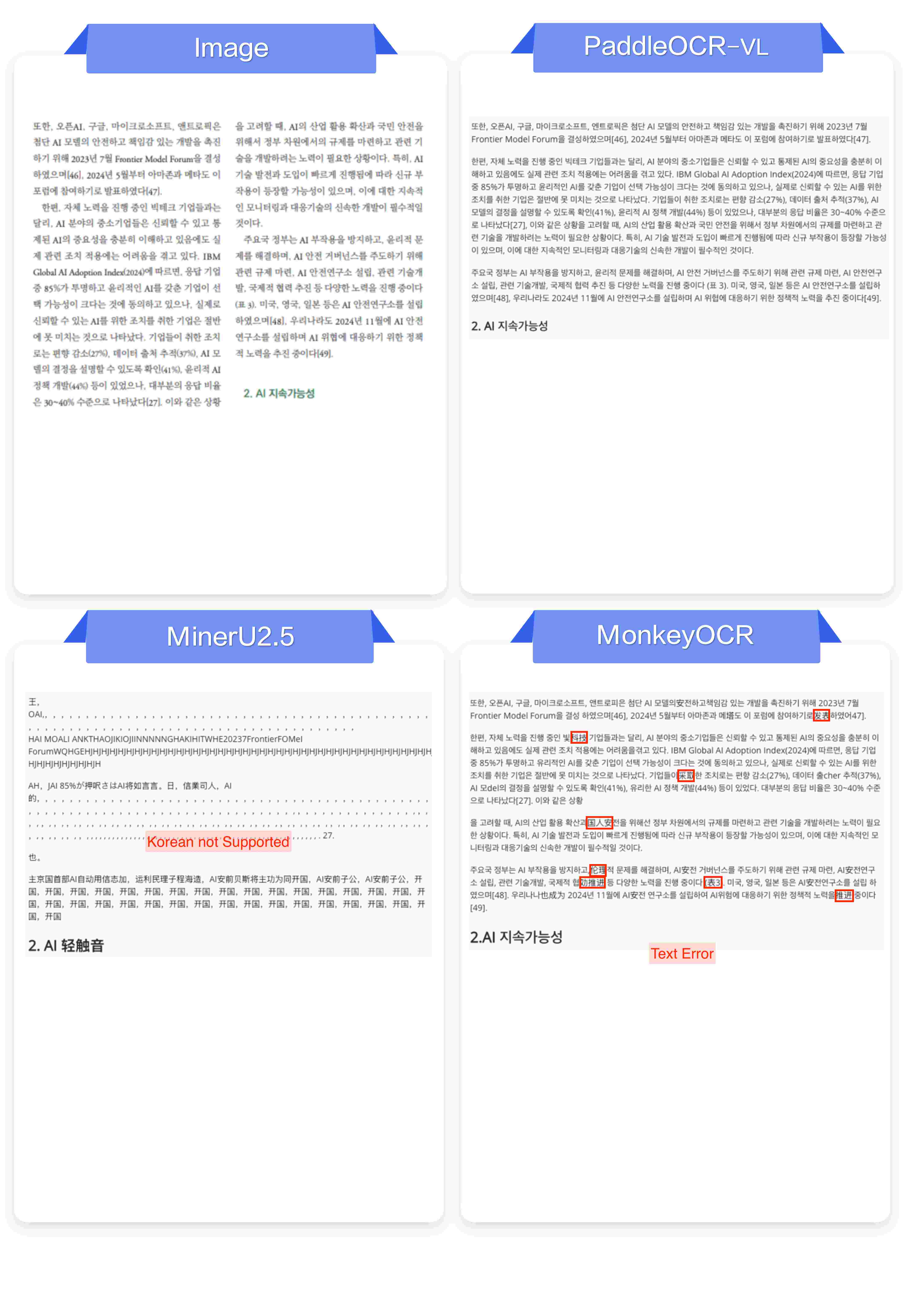} 

\caption{
    \centering
    Compare with others in Multilingual Text Recognition.
}
\label{fig:cmp_text_recognition_multilingual_01}
\end{figure}

\begin{figure}[H]
\centering
\includegraphics[width=0.95\linewidth]{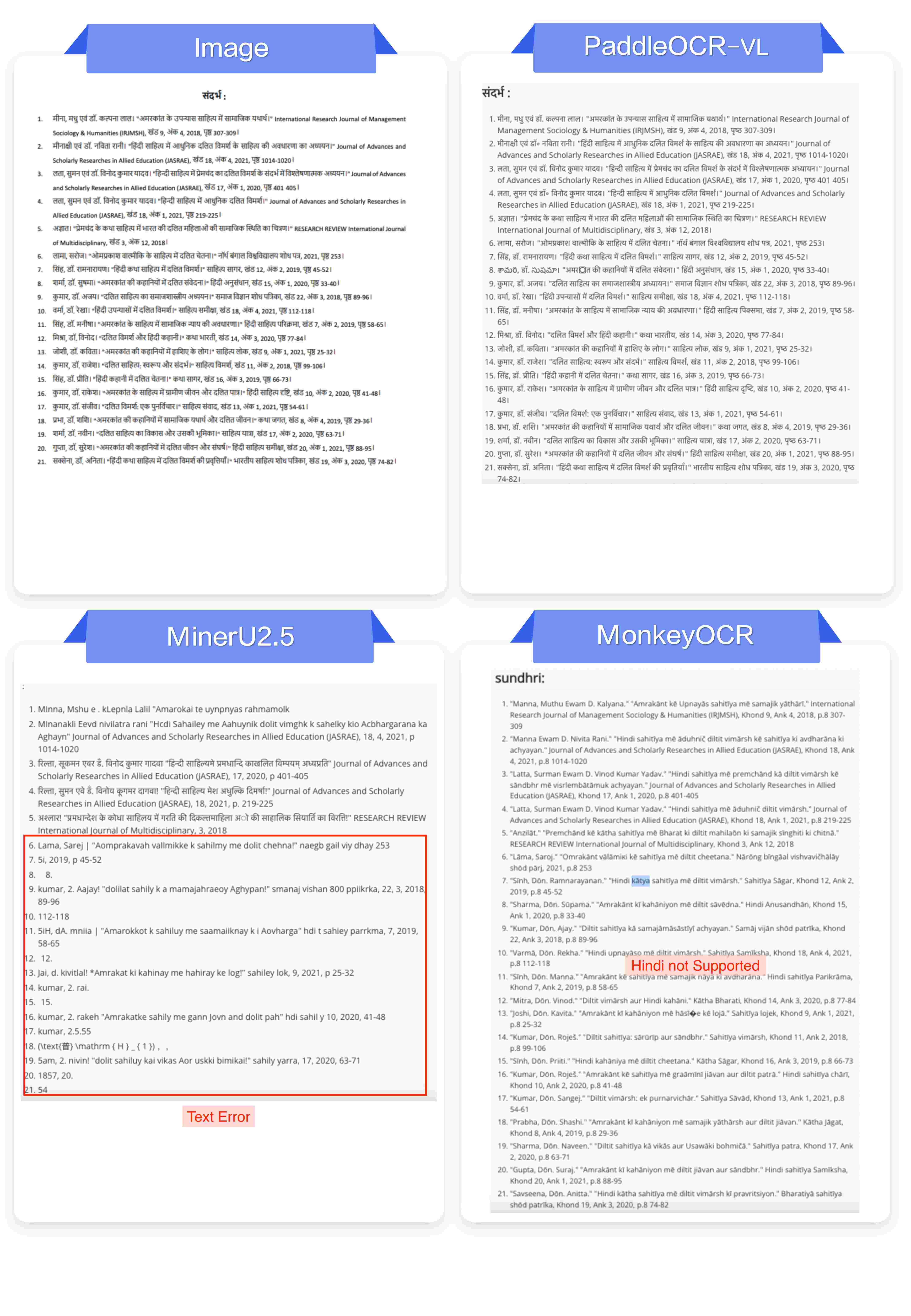} 

\caption{
    \centering
    Compare with others in Multilingual Text Recognition.
}
\label{fig:cmp_text_recognition_multilingual_02}
\end{figure}

\begin{figure}[H]
\centering
\includegraphics[width=0.95\linewidth]{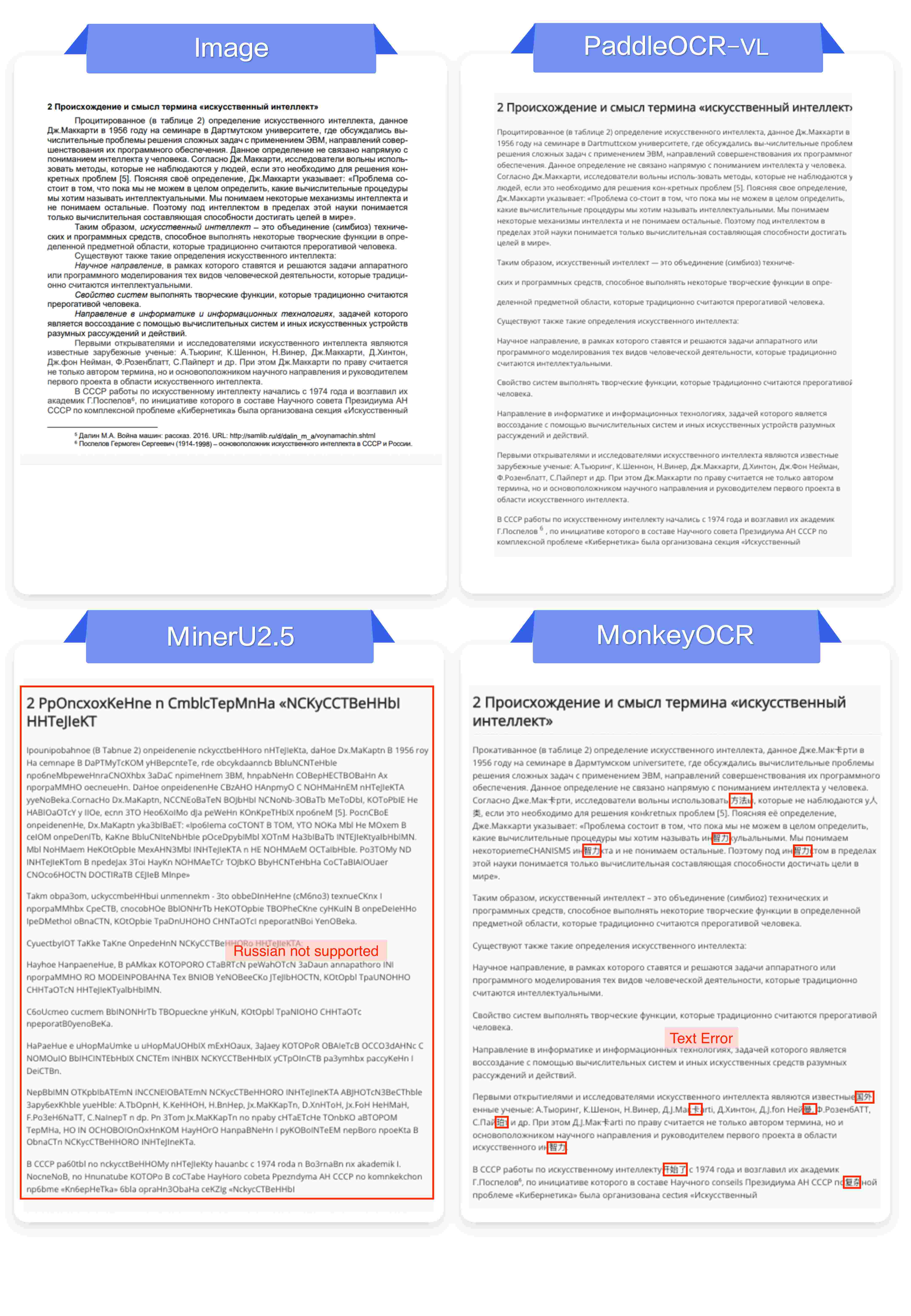} 

\caption{
    \centering
    Compare with others in Multilingual Text Recognition.
}
\label{fig:cmp_text_recognition_multilingual_03}
\end{figure}

\begin{figure}[H]
\centering
\includegraphics[width=0.92\linewidth]{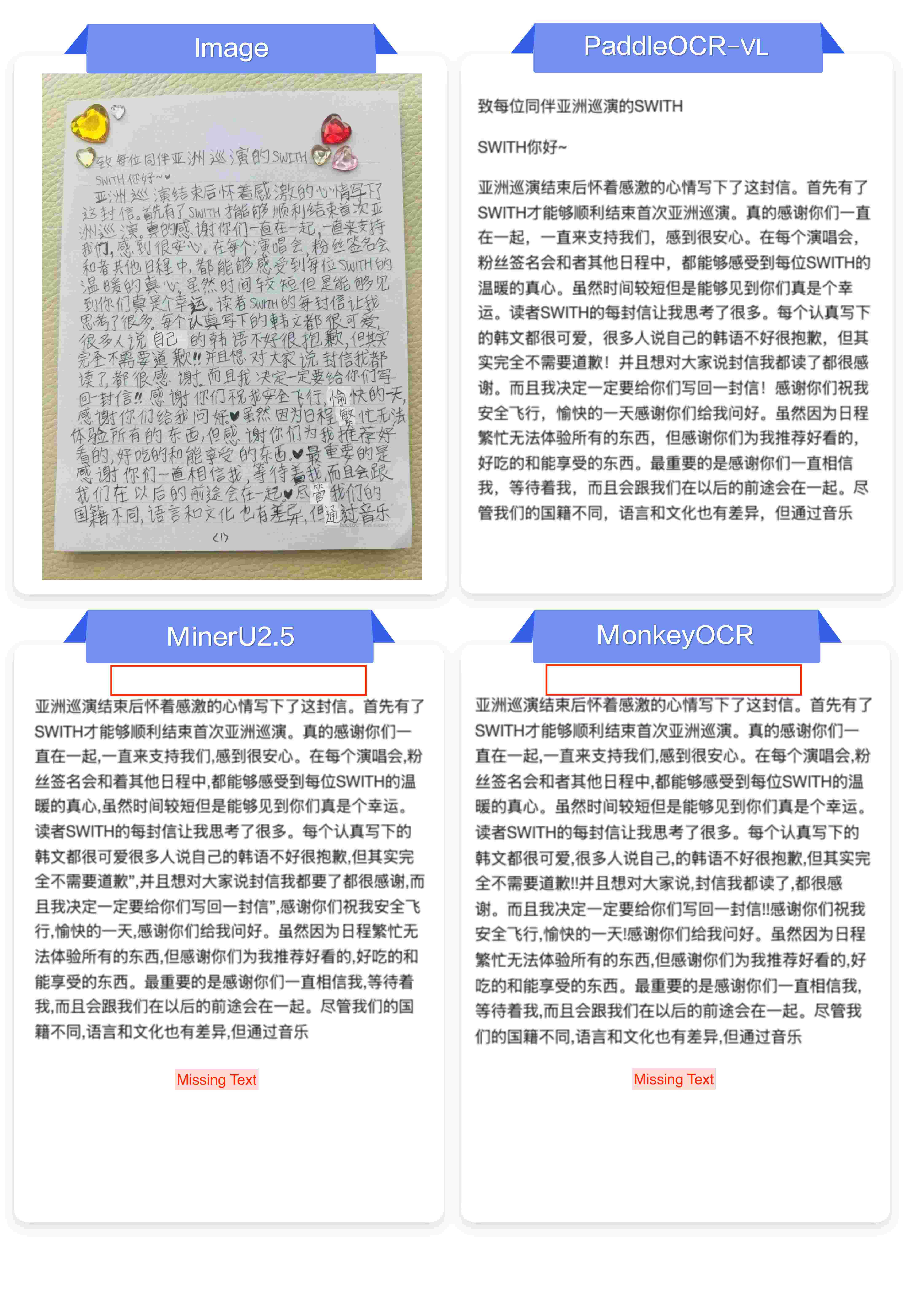} 

\caption{
    \centering
    Compare with others in Handwriting Text Recognition.
}
\label{fig:cmp_text_recognition_handwrite_01}
\end{figure}

\begin{figure}[H]
\centering
\includegraphics[width=0.95\linewidth]{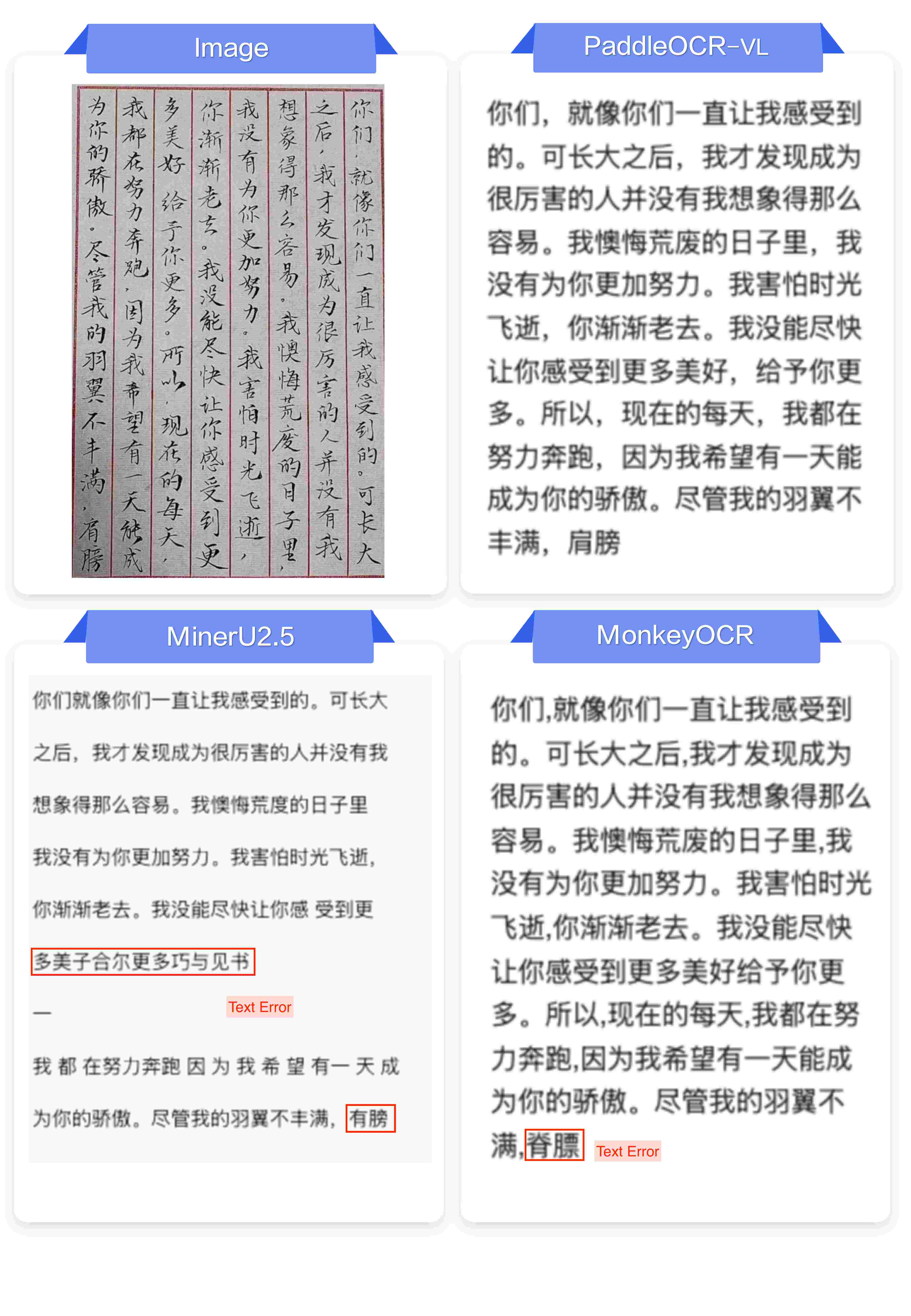} 

\caption{
    \centering
    Compare with others in Handwriting Text Recognition.
}
\label{fig:cmp_text_recognition_handwrite_02}
\end{figure}

\begin{figure}[H]
\centering
\includegraphics[width=0.92\linewidth]{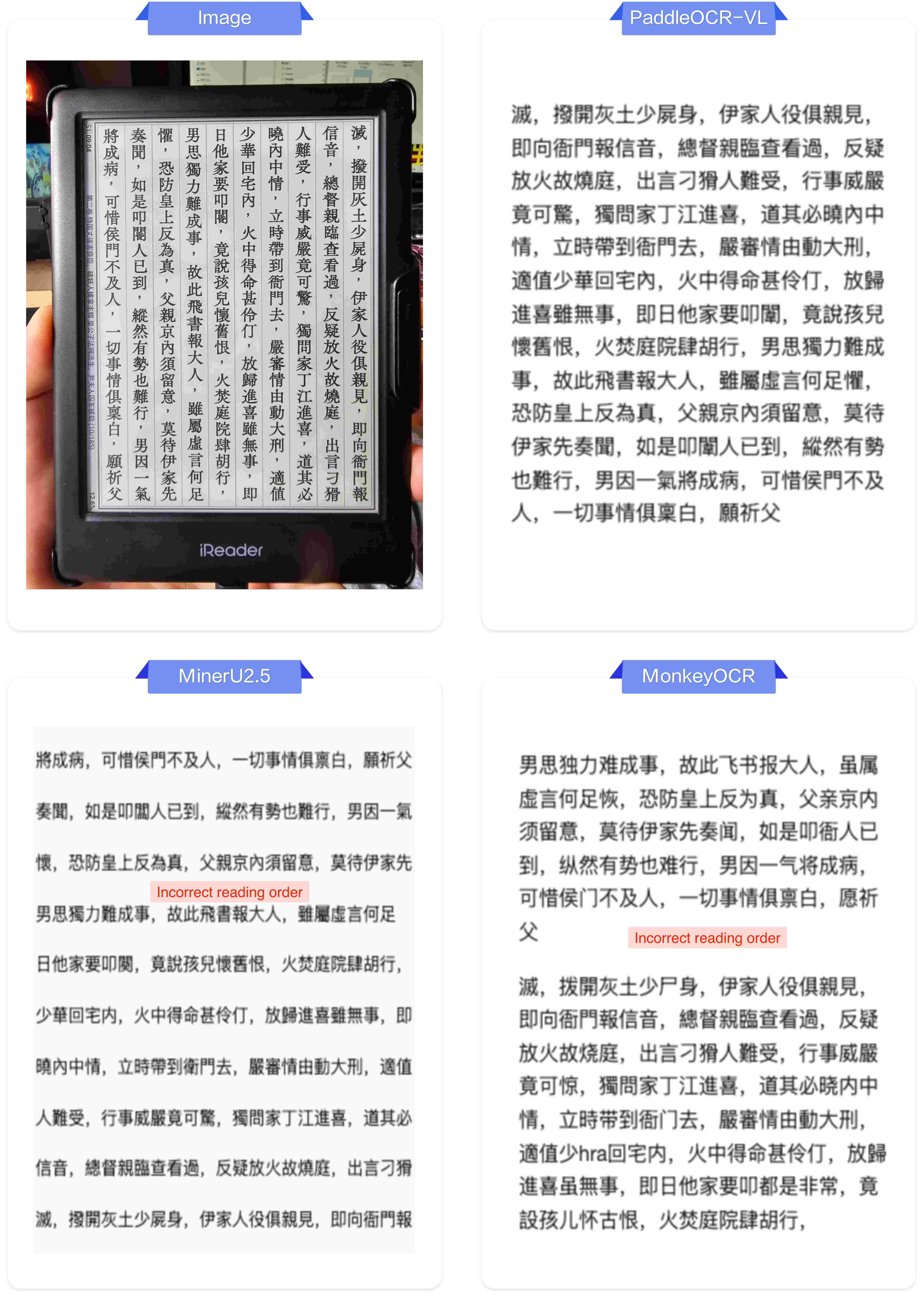} 

\caption{
    \centering
    Compare with others in Vertical Text Recognition.
}
\label{fig:cmp_text_recognition_vertical}
\end{figure}

Moreover, the model demonstrates exceptionally high recognition accuracy in several domains, including Multilingual Text Recognition, Handwriting Text Recognition, and Vertical Text Recognition. Figures \ref{fig:cmp_text_recognition_multilingual_01}- \ref{fig:cmp_text_recognition_vertical} illustrate how PaddleOCR-VL outperforms competitors such as MinerU2.5~\cite{niu2025mineru2} and MonkeyOCR~\cite{li2025monkeyocr}, which tend to misidentify languages like Russian and Hindi as English, overlook some handwritten characters, and struggle with vertical text recognition.

\begin{figure}[H]
\centering
\includegraphics[width=0.92\linewidth]{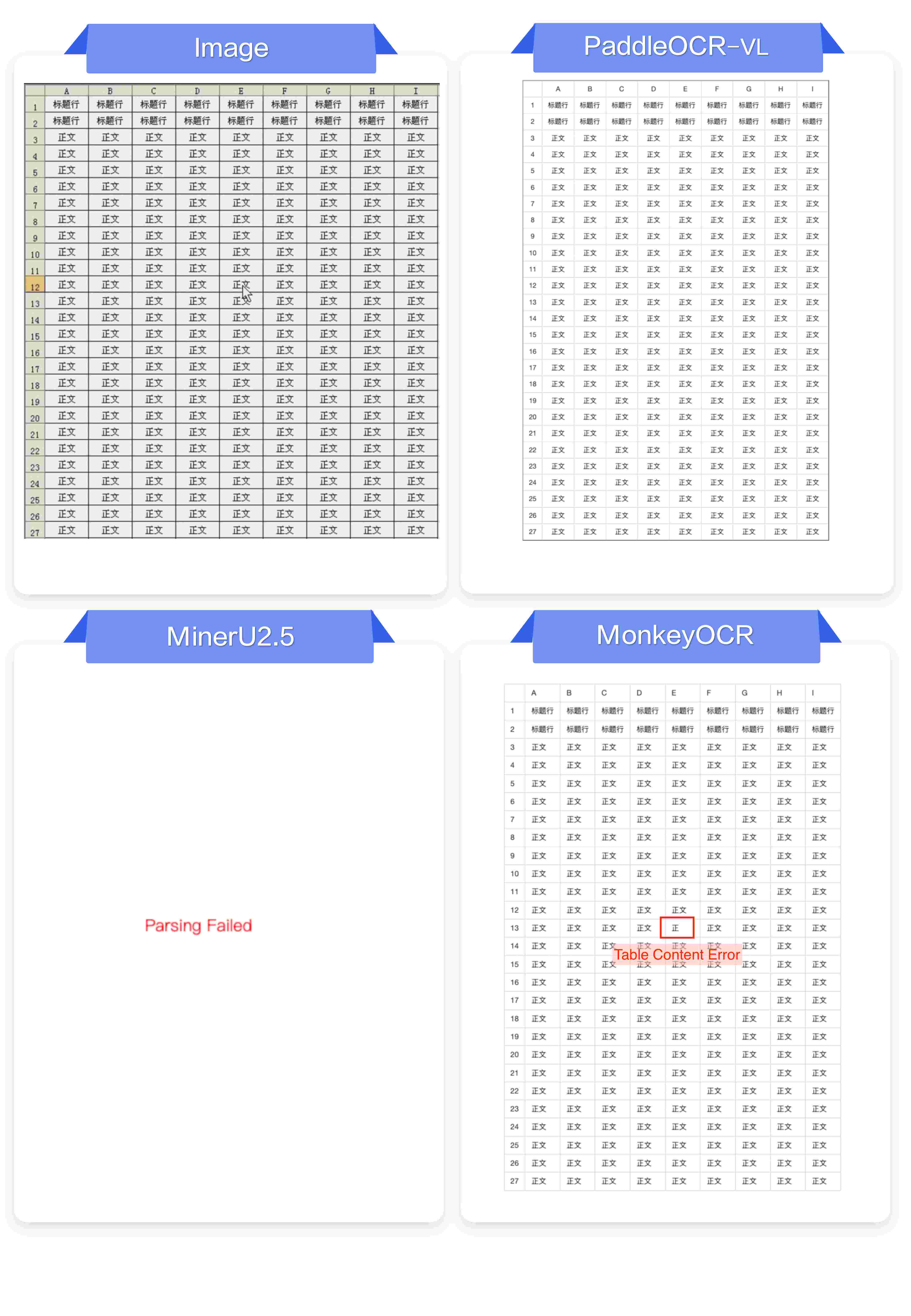} 

\caption{
    \centering
    Compare with others in Table Recognition.
}
\label{fig:cmp_table_01}
\end{figure}

\begin{figure}[H]
\centering
\includegraphics[width=0.95\linewidth]{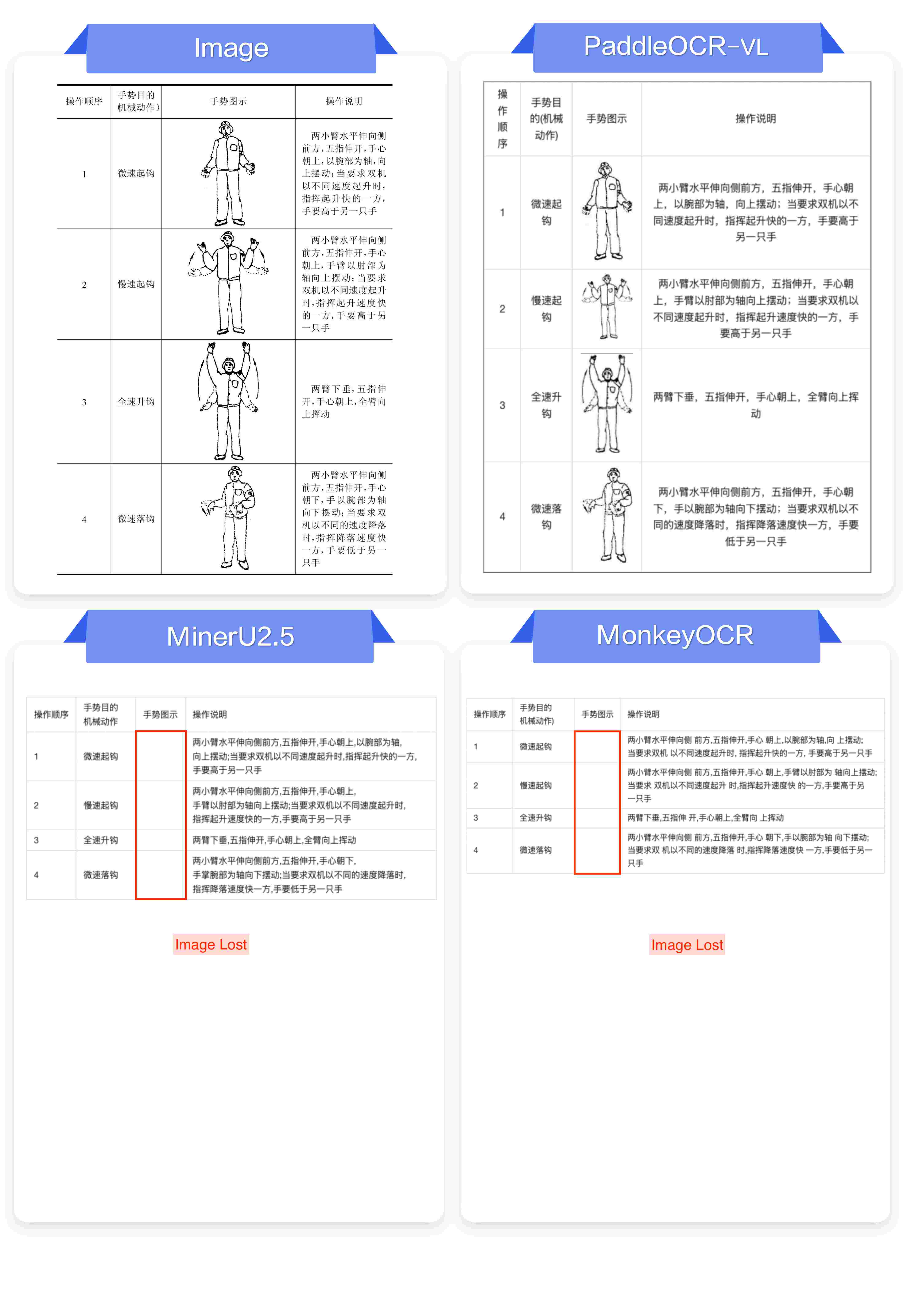} 

\caption{
    \centering
    Compare with others in Table Recognition.
}
\label{fig:cmp_table_02}
\end{figure}

In dealing with complex tables, PaddleOCR-VL's parsing accuracy stands out, as evidenced by Figures \ref{fig:cmp_table_01} and \ref{fig:cmp_table_02}. This is a domain where other models frequently encounter difficulties.

\begin{figure}[H]
\centering
\includegraphics[width=0.95\linewidth]{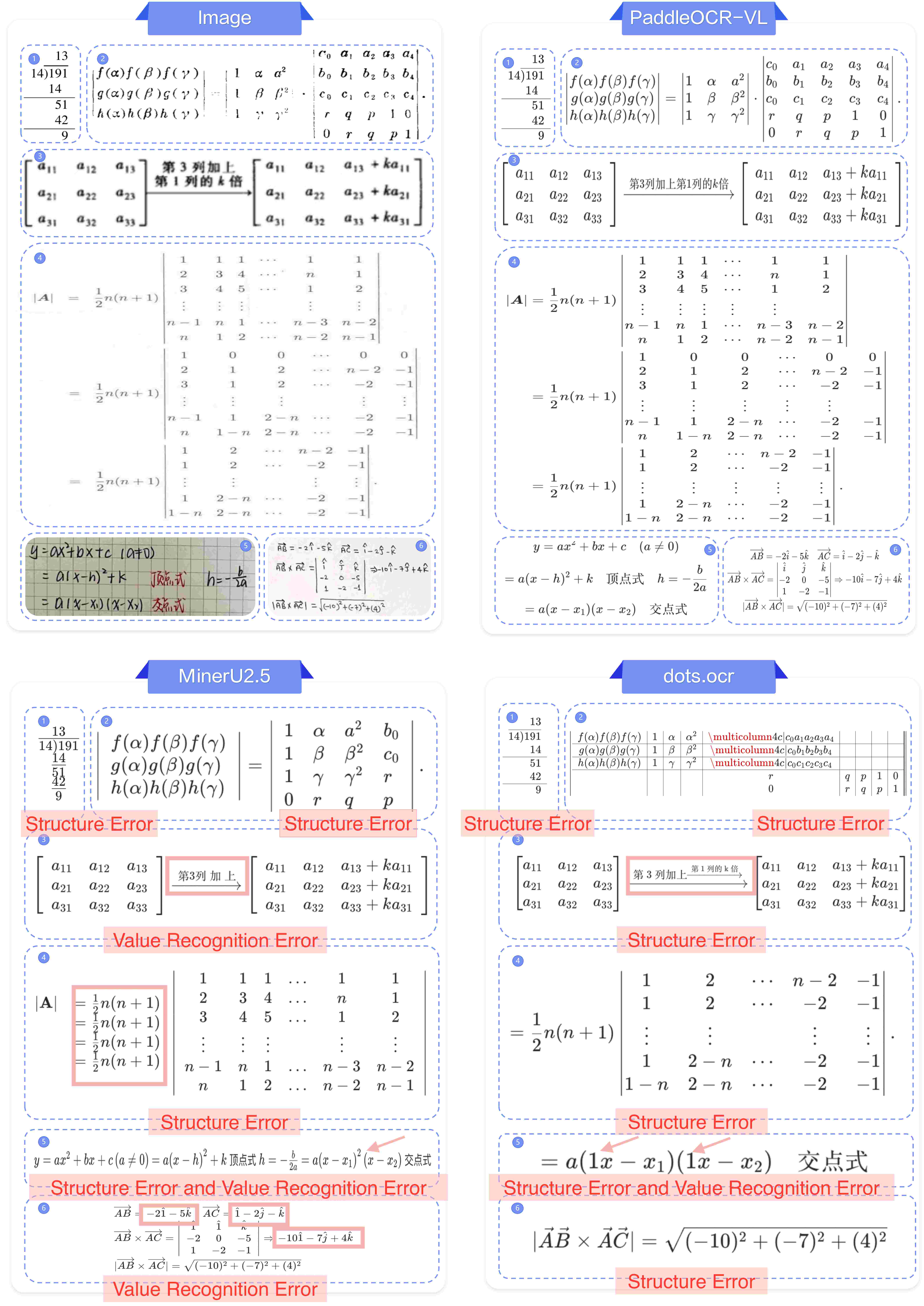} 

\caption{
    \centering
    Compare with others in Formula Recognition.
}
\label{fig:cmp_formula}
\end{figure}

Additionally, Figure \ref{fig:cmp_formula} demonstrates PaddleOCR-VL's proficiency in accurately parsing complex formulas. In contrast, other SOTA models often produce incorrect or flawed outputs when faced with challenging mathematical notations.

\begin{figure}[H]
\centering
\includegraphics[width=0.92\linewidth]{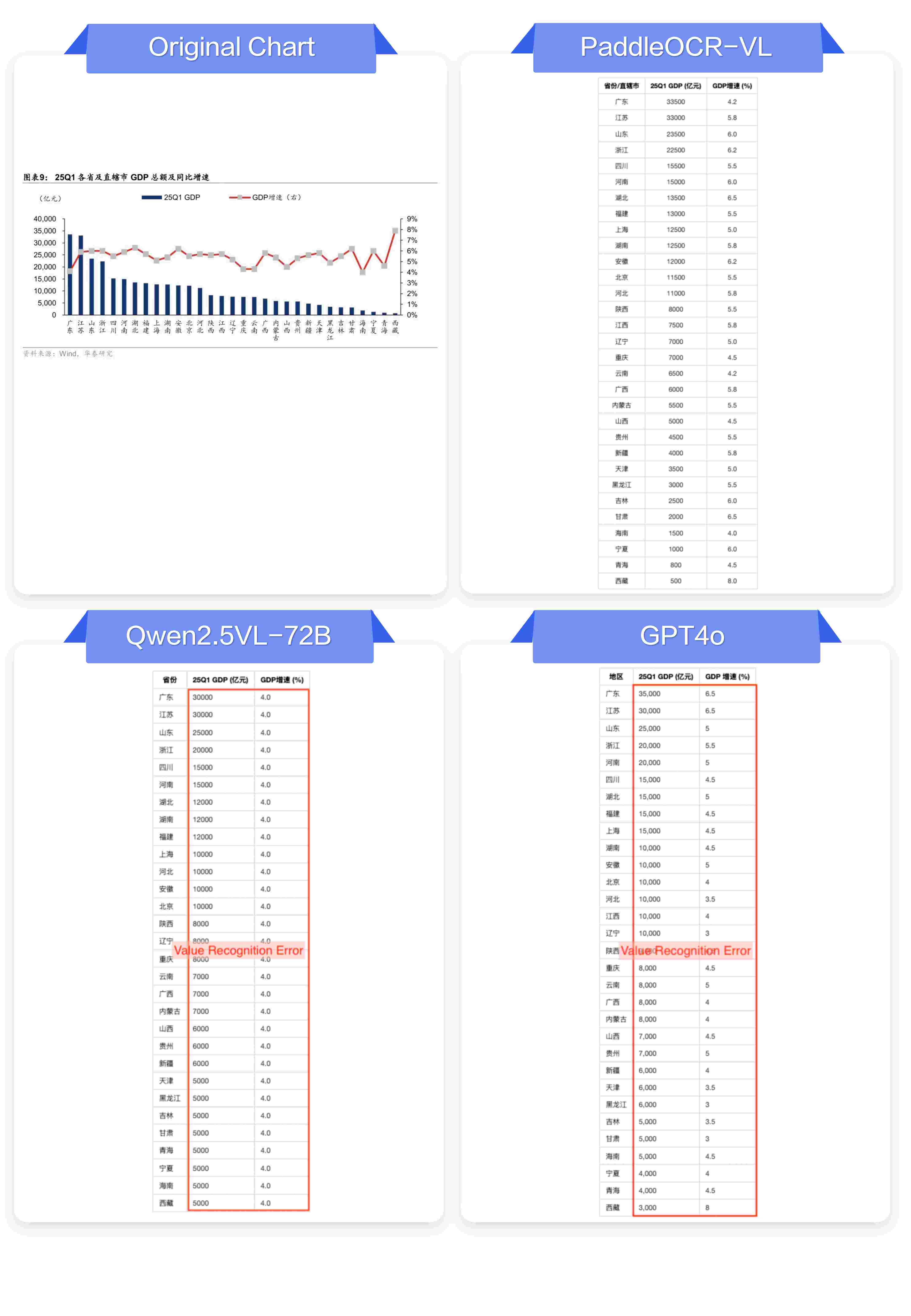} 

\caption{
    \centering
    Compare with others in Chart Recognition.
}
\label{fig:cmp_chart_01}
\end{figure}

\begin{figure}[H]
\centering
\includegraphics[width=0.95\linewidth]{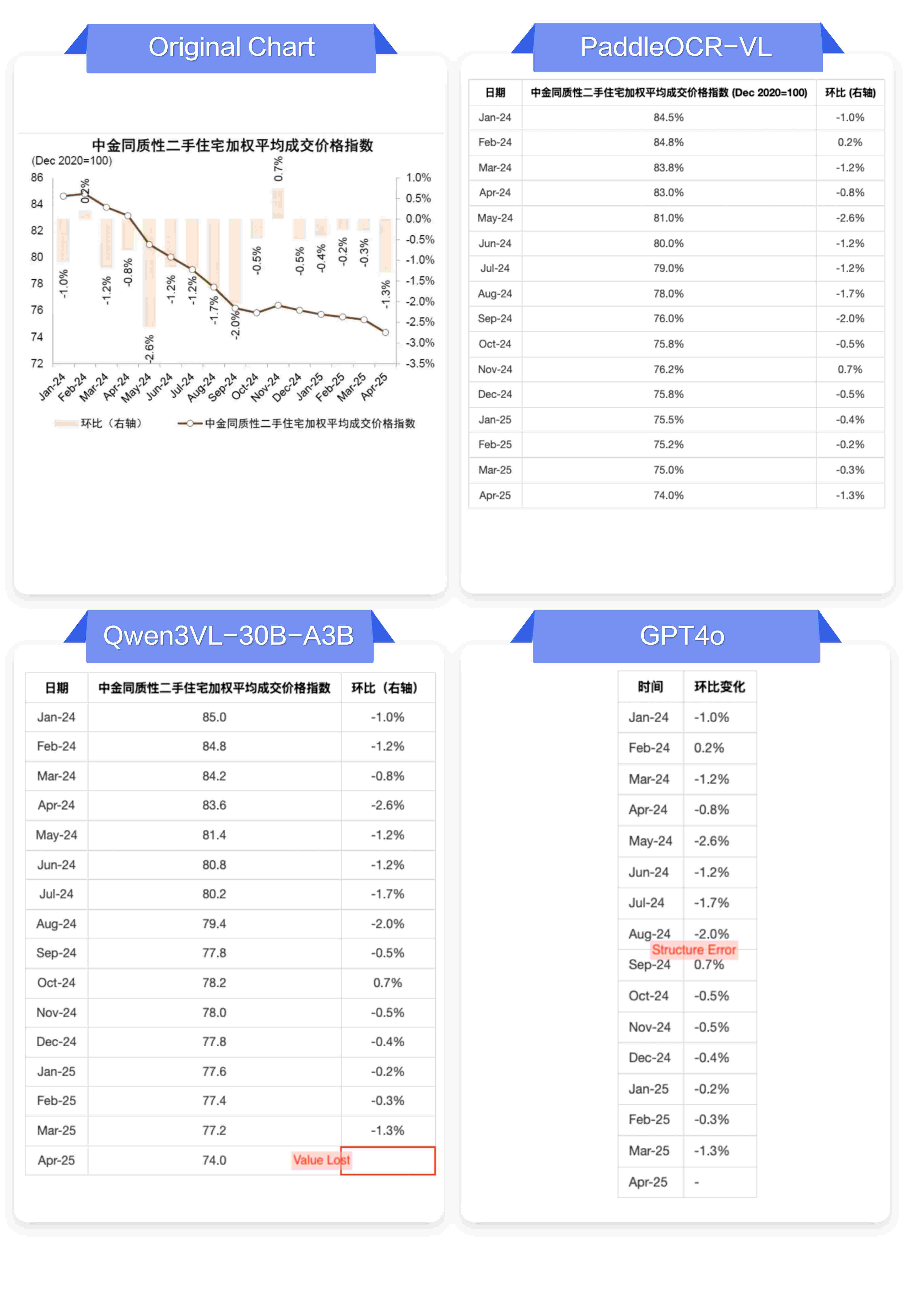} 

\caption{
    \centering
    Compare with others in Chart Recognition.
}
\label{fig:cmp_chart_02}
\end{figure}

Finally, as depicted in Figures \ref{fig:cmp_chart_01} and \ref{fig:cmp_chart_02}, PaddleOCR-VL also excels in Chart Recognition. It outperforms multi-modal large language models like Qwen2.5VL-72B~\cite{bai2025qwen2} and GPT-4o by accurately reconstructing the structure and content of charts.

\subsection{Layout Detection}
\label{subsubsec:layout detection}
\clearpage 
\newpage

\newpage
\subsection{Text Recognition}

\label{subsubsec:text_recognition}
\subsubsection{Multilingual Text Recognition}
\label{subsubsec:text_recognition_multilingual}

\subsubsection{Handwriting Text Recognition}
\label{subsubsec:text_recognition_handwriting}

\subsubsection{Vertical Text Recognition}
\label{subsubsec:text_recognition_vertical}

\newpage
\subsection{Table Recognition}
\label{subsubsec:table_recognition}

\newpage
\subsection{Formula Recognition}
\label{subsubsec:formula_recognition}

\newpage
\subsection{Chart Recognition}
\label{subsubsec:chart_recognition}

\end{document}